\RequirePackage{iftex}
\ifPDFTeX
  \pdfoutput=1
\fi
\ifdefined\asysarticleonly\else
\documentclass{article}
\fi

\usepackage[utf8]{inputenc}
\usepackage[T1]{fontenc}
\usepackage[
  letterpaper,
  textwidth=6.5in,
  textheight=9in,
  top=1in,
  headheight=14pt,
  headsep=25pt,
  footskip=30pt
]{geometry}
\PassOptionsToPackage{hyphens}{url}
\usepackage{xr-hyper}
\usepackage[breaklinks=true,hidelinks]{hyperref}
\hypersetup{
  pdftitle={Agentic Symbolic Search: Approximating Solutions of PDEs Beyond Hand-crafted Expressions, Meshes, and Neural Networks},
  pdfauthor={Zongmin Yu, Jun Zhai and Liu Yang},
  pdfsubject={Mathematical ansatz construction for dynamical systems},
  pdfkeywords={agentic symbolic search, dynamical systems, mathematical models, scientific machine learning}
}
\usepackage{url}
\usepackage{booktabs}
\usepackage{array}
\usepackage{tabularx}
\newcolumntype{Y}{>{\raggedright\arraybackslash}X}
\usepackage{amsfonts}
\usepackage{amsmath}
\usepackage{amssymb}
\usepackage{newtxtext}
\usepackage{newtxmath}
\usepackage{nicefrac}
\usepackage{microtype}
\usepackage{graphicx}
\usepackage[dvipsnames]{xcolor}
\usepackage{subcaption}
\usepackage{natbib}
\usepackage{listings}
\usepackage{float}
\usepackage{placeins}
\usepackage{needspace}
\usepackage{longtable}
\usepackage[ruled,vlined]{algorithm2e}
\usepackage[most]{tcolorbox}
\usepackage{fancyhdr}

\newcommand{\asysrunningtitle}{Agentic Symbolic Search}
\newcommand{\asysheaderright}{}

\fancypagestyle{plain}{%
  \fancyhf{}%
  \fancyfoot[C]{\thepage}%
}

\makeatletter
\renewcommand{\section}{%
  \@startsection{section}{1}{\z@}%
    {-2.0ex \@plus -0.5ex \@minus -0.2ex}%
    {1.5ex \@plus 0.3ex \@minus 0.2ex}%
    {\large\bfseries\raggedright}%
}
\renewcommand{\subsection}{%
  \@startsection{subsection}{2}{\z@}%
    {-1.8ex \@plus -0.5ex \@minus -0.2ex}%
    {0.8ex \@plus 0.2ex}%
    {\normalsize\bfseries\raggedright}%
}
\renewcommand{\subsubsection}{%
  \@startsection{subsubsection}{3}{\z@}%
    {-1.5ex \@plus -0.5ex \@minus -0.2ex}%
    {0.5ex \@plus 0.2ex}%
    {\normalsize\bfseries\raggedright}%
}
\renewcommand{\paragraph}{%
  \@startsection{paragraph}{4}{\z@}%
    {1.5ex \@plus 0.5ex \@minus 0.2ex}%
    {-1em}%
    {\normalsize\bfseries}%
}
\renewenvironment{abstract}
  {\begin{center}\large\bfseries Abstract\end{center}\begin{quote}}
  {\end{quote}}
\makeatother

\SetCommentSty{mycommfont}
\SetKwComment{Comment}{// }{}
\setcitestyle{authoryear,round,citesep={;},aysep={,},yysep={;}}

\definecolor{codebg}{HTML}{F7F7F5}
\definecolor{codeframe}{HTML}{E2E5E9}
\definecolor{codetext}{HTML}{2F343D}
\definecolor{codenum}{HTML}{A7ADB5}
\definecolor{codekeyword}{HTML}{6E6CCF}
\definecolor{codecomment}{HTML}{A6ADB8}
\definecolor{codestring}{HTML}{B38AD8}
\definecolor{codefunc}{HTML}{B46A7A}
\definecolor{codeconst}{HTML}{5E9C93}

\lstdefinestyle{asys_style}{
    backgroundcolor=\color{codebg},
    basicstyle=\ttfamily\footnotesize\color{codetext},
    commentstyle=\color{codecomment},
    keywordstyle=\bfseries\color{codekeyword},
    stringstyle=\color{codestring},
    identifierstyle=\color{codetext},
    numberstyle=\scriptsize\color{codenum},
    numbers=left,
    numbersep=8pt,
    xleftmargin=4pt,
    frame=single,
    rulecolor=\color{codeframe},
    framerule=0.4pt,
    framesep=5pt,
    breaklines=true,
    breakatwhitespace=false,
    showspaces=false,
    showstringspaces=false,
    showtabs=false,
    keepspaces=true,
    tabsize=4,
    captionpos=b,
    aboveskip=6pt,
    belowskip=6pt,
    emph={range,len,min,max,sum,zip,enumerate,print,abs,sorted,np,torch,jnp},
    emphstyle=\color{codefunc},
    emph={[2]True,False,None},
    emphstyle={[2]\color{codeconst}}
}

\ifdefined\asysarticleonly
\fi

\title{Agentic Symbolic Search: Approximating Solutions of PDEs Beyond Hand-crafted Expressions, Meshes, and Neural Networks}

\author{Zongmin Yu$^{1,2,*}$ \qquad Jun Zhai$^{1,*}$ \qquad Liu Yang$^{1,\dagger}$ \\[6pt]
  \normalsize $^{1}$National University of Singapore, Singapore \\
  \normalsize $^{2}$NUS (Suzhou) Research Institute, Suzhou, China}

\date{}

\begin{document}
\maketitle
{\renewcommand{\thefootnote}{}\footnotetext{$^{*}$These authors contributed equally to this work.\quad $^{\dagger}$Corresponding author: \texttt{yangliu@nus.edu.sg}}}

\begin{abstract}
Numerical methods and neural networks have transformed our understanding of nonlinear partial differential equations (PDEs), yet discovering mathematical representations that explain their dynamics remains largely an expert task.
We introduce Agentic Symbolic Search (ASYS), in which coding agents construct and refine explicit approximations from a known PDE and supplied mathematical knowledge.
ASYS couples evolutionary search over differentiable symbolic programs with gradient-based fitting, using the governing equation to evaluate candidates and guide their revision.
For a singular generalized Constantin--Lax--Majda profile, ASYS infers a missing decay term that sets how fast the field approaches its singular-time limit, and builds a basis that fits the profile to a median error of $3.90\times10^{-5}\%$.
For Fujita blow-up, separating diffusion from reaction and deriving a local response lowers the supplied formula's error from $55.5\%$ to $0.0879\%$.
For Allen--Cahn interfaces, a 14-parameter construction yields a mode-coupling law confirmed by independent simulations.
These studies show how agentic search develops mathematical starting points into accurate, analyzable approximations.

\end{abstract}
\section{Introduction} \label{sec:intro}

Analytical expressions have long clarified how physical systems evolve, from linear wave propagation to heat conduction~\citep{fourier1822}. For most nonlinear partial differential equations (PDEs), however, no closed-form solutions are known. Numerical methods and neural networks can approximate individual evolutions with high field accuracy, yet they store the dynamics in grid values or network weights, not in formulas that can be inspected. Mathematical understanding therefore still relies on structured approximations---explicit expressions designed to make nonlinear behavior accessible to analysis. The choice of mathematical form is decisive: it determines which physical mechanisms can be represented, how reliably parameters can be fitted and what analytical deductions the approximation can support.

An \emph{ansatz} is an informed guess about the structure of a solution, written as an expression that can be tested against the governing equation. Standard computational approaches fit parameters within a fixed representation, such as basis coefficients or network weights. Parameter optimization alone, however, cannot supply a physical feature that the representation excludes, such as an omitted boundary layer, asymptotic power or mode interaction. At the same time, enriching an expression with flexible terms often adds terms that nearly duplicate one another, which makes the fit ill-conditioned. Developing an effective approximation has therefore required human experts to inspect where an ansatz fails, deduce what terms the equation requires and revise both the symbolic representation and the numerical procedure used to fit it.

Large language model coding agents offer a way to carry out much of this process from a mathematical starting point. Operating over both mathematical text and executable code, agents can translate domain principles---such as scaling laws, symmetries and boundary behaviors---into concrete structural hypotheses implemented as differentiable programs. Candidate fields can then be differentiated, evaluated against the PDE residual and fitted by gradient-based optimization. By examining the symbolic formula alongside its optimization diagnostics, an agent can determine whether a remaining error comes from a missing term in the ansatz or from a poorly conditioned fit, and revise the program accordingly.

Existing automated PDE solvers address parts of this problem within restricted search spaces. Physics-informed neural networks (PINNs)~\citep{raissi2019physics}, neural operators~\citep{li2021fourier,lu2021deeponet} and neural self-similar blow-up solvers~\citep{wang2023self,wang2025unstable} achieve strong numerical accuracy---including for unstable singular profiles---but return network surrogates rather than explicit analytical expressions. Symbolic solvers seek closed-form or finite expressions directly from the governing equation: the finite expression method (FEX)~\citep{liang2025fex}, SSDE~\citep{ssde2025} and SymPlex~\citep{park2026symplex} search over expression trees, while the Symbolic Iterative Grammar Solver (SIGS)~\citep{oikonomou2026sigs} searches a latent grammar space, and \citet{bhatnagar2025equations} use fine-tuned language models to restrict FEX's operator library. Because these methods build candidates from prescribed operator sets or grammars, they cannot easily derive a coordinate, a response term or a parameterization specific to the problem.

A complementary line of work uses language models to search over executable programs. Evolutionary systems such as FunSearch~\citep{romera2024funsearch} and AlphaEvolve~\citep{alphaevolve2025} evolve code under quantitative evaluators, while LLM-SR~\citep{llmsr2025} and SR-Scientist~\citep{xia2025srscientist} apply program search to symbolic regression from observational data. In scientific machine learning, multi-agent frameworks such as ATHENA~\citep{toscano2025athena} and GRAFT-ATHENA~\citep{toscano2026graftathena} coordinate classical and neural solvers across end-to-end workflows, and AgenticSciML~\citep{jiang2026agenticsciml} uses collaborating agents to develop neural architectures and training strategies for operator and physics-informed learning. For PDE approximation, this creates an opportunity to deploy agentic program search directly at the level of mathematical modeling, iteratively revising both the analytical form of an ansatz and the numerical conditioning used to fit it.

Here we introduce Agentic Symbolic Search (ASYS), a framework that develops mathematical starting points into explicit, analyzable approximations of nonlinear PDEs. ASYS uses coding agents within an evolutionary loop to propose differentiable symbolic programs whose structure is scored by a fixed equation-residual evaluator while trainable parameters are fitted in an inner optimization loop. Candidate programs specify the analytical representation together with coordinate maps or fitting routines suited to the problem, and final constructions are assessed against independent numerical reference solutions. Rather than competing with meshes or neural networks on raw field error, ASYS aims at something different: compact explicit expressions whose terms and limits can be analyzed and checked.

We evaluate ASYS across singular blow-up and moving-interface problems, beginning with localized concentration. For a self-similar singular profile of the generalized Constantin--Lax--Majda (gCLM) equation, ASYS infers a secondary algebraic decay term from far-field nonlinear balance that the initial ansatz lacks, and in its best search builds a mixed basis together with an invertible change of coefficient coordinates that improves the conditioning of the fit. Across five independent searches, the median profile error is $3.90\times10^{-5}\%$ (Table~\ref{tab:accuracy-comparison}), and the inferred tail term gives the rate at which the physical field approaches its singular-time limit. In radial Fujita reaction--diffusion, ASYS separates diffusion from reaction and derives a local response from the equation, lowering relative space--time $L^2$ error from $55.5\%$ for the fitted starting ansatz to $0.0879\%$. Under the same fourteen parameters and fitting procedure, an earlier candidate's uniform response leaves the error at $3.99\%$ (Table~\ref{tab:accuracy-comparison}).

In two-dimensional interface motion, ASYS likewise derives explicit relations that grid values or network weights do not show. Starting from geometric curvature guidance for the Allen--Cahn equation, ASYS integrates a second-order curvature expansion to derive how an initial elliptical elongation generates a fourfold angular harmonic ($m = 4$) without a separately fitted amplitude. Built into a field with 14 fitted parameters, this gives an explicit mode-coupling law confirmed by independent PDE simulations, whereas a 12,801-parameter PINN baseline achieves lower field error ($0.122\%$ versus $0.798\%$; Table~\ref{tab:accuracy-comparison}) but does not show how the modes interact. Together with six further studies, these results show how agentic symbolic search can turn governing equations and domain principles into compact approximations that support mathematical analysis.

\section{Results}\label{sec:results}

\subsection{Overview of the Agentic Symbolic Search framework}
\label{sec:asys}

\begin{figure}[htbp]
\centering
\includegraphics[width=\linewidth]{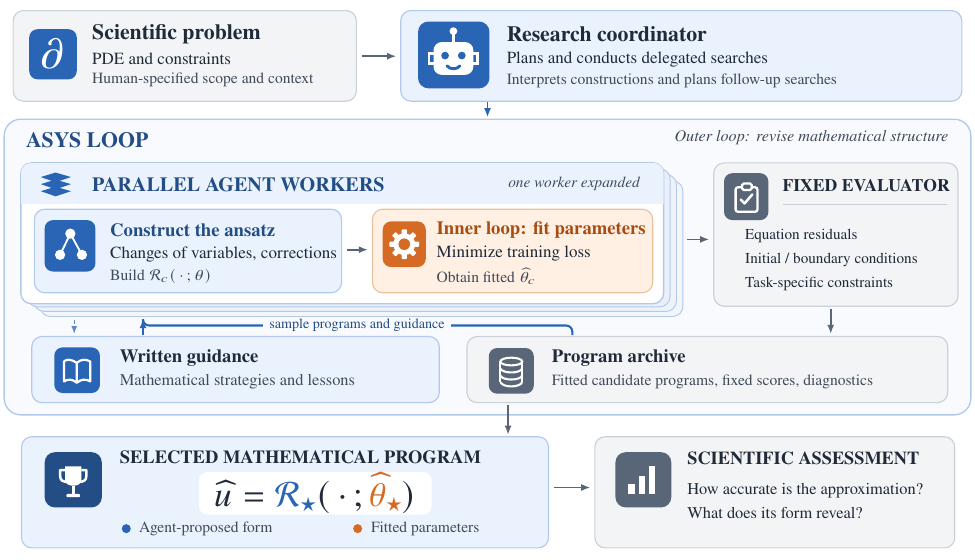}
\caption{\textbf{Agent-designed ans\"atze and coefficient fitting.}
ASYS alternates between revising the mathematical form of an ansatz and numerically fitting its coefficients. The training loss guides fitting, whereas the evaluator's fixed score guides structural search. Programs and written guidance carry experience into subsequent proposals.}
\label{fig:method-pipeline}
\end{figure}

ASYS takes as input a textual description of a known PDE, its initial or boundary conditions, any additional constraints such as symmetry or normalization, and the starting mathematics supplied with the task. It constructs an explicit mathematical approximation through two nested loops (Fig.~\ref{fig:method-pipeline}). The outer loop proposes and revises the mathematical form of the ansatz---its coordinate transformations, decomposition into branches and choice of trainable parameters---while the inner loop fits any trainable coefficients against the equation and prescribed constraints. The search is therefore discrete at the level of mathematical structure and continuous at the level of parameter values.

Each returned program evaluates the approximate field from explicit formulas, computing any response integrals numerically, with coefficients either fitted in the inner loop or prescribed directly by the construction. This is what we mean by an explicit approximation: its functional form is written out and can be read and analyzed term by term, and it holds no grid values or network weights, although fixed constants of its basis and some integrals are evaluated numerically.

The two loops use distinct objectives. The inner loop minimizes a scalar training loss, whereas the outer loop compares fitted constructions using a fixed evaluator. Both rely on the equation residual: the mismatch obtained by substituting the approximation into the PDE. Where the task permits, the agent also specifies the training loss and fitting procedure as part of its proposal, so that it can change how the fit is trained while the evaluation standard stays fixed.

ASYS runs the outer loop within Evolving Ensemble of Agents (EvE), which retains candidate programs and written guidance. In the reported experiments, workers ran in the Codex CLI harness with tools to edit and execute code, receiving earlier constructions and diagnostic feedback as context for subsequent proposals. In the reported studies, the researchers prepared each task with its starting mathematics and guidance, the workers developed the constructions, and a coordinator, itself an AI agent, inspected the returned constructions and decided when to continue or stop a search (Methods; Table~\ref{tab:supp-run-config} lists what each study was given). Coding sessions took 22.73 agent hours across the five gCLM searches, 2.26 hours for Fujita and 2.54 hours for Allen--Cahn, not counting coefficient fitting and evaluation.

\FloatBarrier

\subsection{Constructing approximations from the equations}
\label{sec:construction-overview}

We examined whether ASYS could develop mathematical starting points into accurate, explicit approximations that support further analysis. Each study tests a part of this task: choosing a useful mathematical form, improving it through the governing equation and working out what the resulting expression implies.

The gCLM study first tests how ASYS expands an inadequate family of profiles and makes the new family fit accurately, yielding a prediction for the rate at which the field approaches its singular-time limit. The Fujita study isolates the effect of one revision to a formula by comparing two reaction responses with the same parameter count and fitting procedure. The Allen--Cahn study tests whether the governing equation can determine interactions within an approximation: ASYS derives interactions among interface shape modes to give a compact description of a moving boundary. Together, these studies connect the construction of an approximation to both its accuracy and its mathematical usefulness.

During search, constructions were assessed against the governing equations and prescribed constraints. Independent numerical references supplied separate field-accuracy measurements afterward, supplemented in Allen--Cahn by a direct check of the derived shape dynamics against the exact curvature law. Table~\ref{tab:accuracy-comparison} reports fitted parameter counts and field errors for the three main problems, with symbolic and neural reference runs under each method's own setup, and Supplementary Table~\ref{tab:construction-overview} summarizes the mathematical outcomes across the wider set of studies introduced in Section~\ref{sec:further-constructions}.

\begin{table}[htbp]
\centering
\small
\caption{\textbf{Accuracy and fitted parameter counts.} Relative $L^2$ errors are reported in percent using a common target and readout within each study. For gCLM, search medians and the selected construction are reported separately.}
\label{tab:accuracy-comparison}
\renewcommand{\arraystretch}{1.12}
\begin{tabular*}{\linewidth}{@{\extracolsep{\fill}}lllrr@{}}
\toprule
Problem & Regime & Method or construction & Error (\%) & $n_\theta$ \\
\midrule
gCLM & Blow-up & SIGS adaptation, median of 2 searches & $0.0119$ & 6 \\
 & & FEX adaptation, median of 2 searches & $0.00851$ & 13 \\
 & & ASYS, median of 5 searches & $3.90\times10^{-5}$ & 11--19 \\
 & & ASYS & $2.47\times10^{-8}$ & 15 \\
\addlinespace
Fujita & Blow-up & PINN & $5.30$ & 12,739 \\
 & & ASYS with uniform response$^{\ast}$ & $3.99$ & 14 \\
 & & ASYS & $0.0879$ & 14 \\
\addlinespace
Allen--Cahn & Bounded & PINN & $0.122$ & 12,801 \\
 & & ASYS & $0.798$ & 14 \\
\bottomrule
\end{tabular*}
\par\smallskip
\begin{minipage}{\linewidth}\footnotesize\raggedright
$n_\theta$ counts fitted scalar parameters. For gCLM, the range spans the five returned programs (nine-hour limit per search), and the selected program also stores 495 normalization and basis constants that are not fitted; SIGS and FEX were supplied with the fractional coordinate from the completed ASYS search and run twice with a four-hour allocation on one CPU thread (Supplementary Table~\ref{tab:gclm-tail-capable}). Fujita errors use the radial-profile norm over the full trajectory; its PINN entry is the median of three fits on 5,760 collocation points with a one-hour allowance each, whereas the analytical response comparison uses a 300-second allowance (Supplementary Section~\ref{supp:fujita-neural-control}). $^{\ast}$Selected ASYS construction refitted after replacing the local response $F(s,c)$ with the time-only response $F(s,1)$ from iteration 4 (Table~\ref{tab:fujita-history}). Allen--Cahn errors use trapezoidal time weights across 262 saved times; its PINN entry is the median of three 20-minute fits under the same L-BFGS settings as the reported ASYS construction (Supplementary Section~\ref{supp:ac-current-neural-controls}), and the twelve-parameter extension is assessed in Supplementary Section~\ref{supp:ac-coupled-metrics}.
\end{minipage}
\end{table}

\FloatBarrier

\FloatBarrier

\subsection{gCLM: constructing a singular profile and its far-field decay}
\label{sec:gclm}

The generalized Constantin--Lax--Majda (gCLM) equation models the competition between transport and vortex stretching in fluid flow~\citep{constantin1985clm,degregorio1990,okamoto2008generalization}, providing a one-dimensional setting for studying finite-time singularity formation from smooth fields. In a self-similar singularity, the field concentrates while preserving its rescaled shape~\citep{barenblatt1996scaling}; an explicit expression connecting the smooth core to a slowly decaying far field makes this concentration law explicit and links the singular core to the surrounding flow. We considered
\begin{equation}
\omega_t+a u\omega_x-u_x\omega=0,\qquad u_x=H\omega,\qquad a=1/4,
\label{eq:gclm_pde}
\end{equation}
where $H$ is the real-line Hilbert transform. For singularity time $T$, the similarity reduction $\omega(x,t)=-\lambda(T-t)^{-1}f(y)$, with $y=x(T-t)^{-\beta}$ and $\lambda=(1+\beta)/(1-a)$, leaves the profile $f$ and scaling exponent $\beta>0$ as unknowns. We sought the known odd profile that is positive for $y>0$ and normalized by $f'(0)=(-Hf)(0)=1$~\citep{huang2024gclm,lushnikov2021gclm}.

The supplied starting ansatz, $h_0(x)=x(1+x^2)^{-(1+\alpha)/2}\exp\{c[(1+x^2)^{-1}-1]\}$ in the scaled variable $x=Ay$, has the leading decay $y^{-\alpha}$, with $\alpha=1/\beta$, but its corrections proceed in steps of two powers ($y^{-\alpha-2}, y^{-\alpha-4}, \ldots$). Balancing the equation's nonlinear far-field terms~\citep{bender1999asymptotic}, ASYS inferred a $y^{-2\alpha}$ correction absent from this family. At the fitted $\alpha\simeq1.4609$, this $y^{-2.92}$ term lies between the leading decay $y^{-1.46}$ and the starting ansatz's first correction $y^{-3.46}$, so fitting coefficients alone cannot supply it. ASYS therefore introduced a coordinate $v$ decaying as $y^{-\alpha}$ so that a correction linear in $v$ provides the missing power---the term that subsequently gives the rate at which the field approaches its singular-time limit.

Representing the full profile accurately also required control over higher-order decay and intermediate radii. ASYS enriched a polynomial in $v$ with additional fractional powers, giving independent coefficients for further tail terms, and added localized rational functions for more shape freedom between the center and far field. Combined into $Q(v)=\sum_{j=1}^{14}c_j\phi_j(v)$, these fourteen basis functions reshape the profile through an exponential factor, with weights $c_j$ and exponent $\beta$ fitted against the profile equation. With the positive scale $A$ fixed by the normalization $(-Hf)(0)=1$, the profile takes the form
\begin{equation}
\begin{aligned}
 r&=\sqrt{1+A^2y^2},\qquad v=r^{-\alpha},\\
 f(y)&=\frac{y}{r}\,v\exp\!\bigl[(1-v)^2Q(v)\bigr].
\end{aligned}
\label{eq:h3_ansatz}
\end{equation}
Here $v=1$ at the center and $v\to0$ in the far field. The factor $y/r$ enforces odd symmetry, $v$ sets the leading decay, the exponential preserves positivity and $(1-v)^2$ maintains $f'(0)=1$. Supplementary Section~\ref{supp:h3-construction} gives the complete basis and fitted expression.

Fitting this richer family accurately required a further conditioning construction. Because several basis functions had nearly identical shapes, changes in their coefficients could cancel one another and impede fitting~\citep{higham2002accuracy,golub2013matrix}. ASYS subtracted the fractional powers' overlap with the polynomial part and applied an invertible Gram-coordinate transformation that mixes and rescales the coefficient directions using the basis functions' measured overlap. This change of coordinates kept the same family of possible profiles while making its coefficients easier to fit. Table~\ref{tab:gclm-history} shows how the search developed this construction beyond the initial tail correction.

\begin{table}[htbp]
\centering\small
\caption{Selected construction steps in the search that produced the reported gCLM profile. Values are equation-residual mean squared errors (MSEs) observed during search rather than errors against the numerical reference profile. All fits used Adam with cosine learning-rate decay under a 20,000-update or 1,200-second allowance.}
\label{tab:gclm-history}
\begin{tabular}{rll}
\toprule
ASYS iteration & Change & Equation-residual MSE\\
\midrule
0 & Supplied starting ansatz & $1.10\times10^{-5}$\\
1 & Polynomial in the fractional coordinate $v$ & $2.83\times10^{-11}$\\
2 & Mixed-tail terms added & $5.49\times10^{-11}$\\
3 & Redundant radial scale fixed; Jacobi polynomial basis & $2.98\times10^{-14}$\\
4 & Fractional powers' overlap with the polynomial part removed; Gram coordinates & $1.56\times10^{-15}$\\
5 & Independent fourth-order tail term & $2.68\times10^{-18}$\\
6 & Further mixed-tail terms & $7.93\times10^{-18}$\\
10 & Selected construction, with a pair of rational terms & $8.72\times10^{-20}$\\
\bottomrule
\end{tabular}
\end{table}

\begin{figure}[htbp]
\centering
\includegraphics[width=\linewidth]{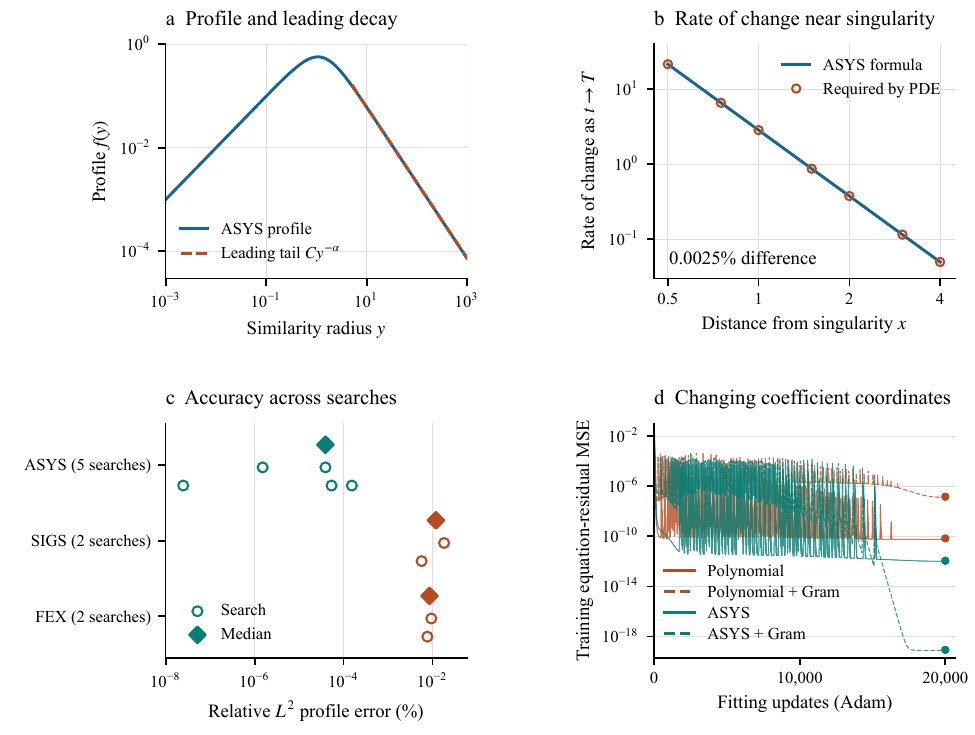}
\caption{\textbf{Constructing a profile with accessible far-field dynamics.}
\textbf{a,} The selected ASYS profile and its leading tail $Cy^{-\alpha}$ on logarithmic axes.
\textbf{b,} Rate of change of the field at each fixed position $x>0$ as the singular time $T$ approaches. The rate $\lambda D x^{-2\alpha}$ read from the ASYS formula (line) agrees within $0.0025\%$ with the rate required by the PDE's far-field balance for the leading tail (circles), both evaluated from the fitted profile.
\textbf{c,} Profile errors from five ASYS searches and two searches each for the SIGS and FEX adaptations. Open circles show individual outcomes; diamonds mark each method's median.
\textbf{d,} Training equation-residual MSE for the polynomial and ASYS families, each with and without the Gram transformation. All four fits fix $\beta$, use fourteen shape coefficients and share the zero initialization, sampling rules and learning-rate schedule. Both ASYS fits retain the subtraction of polynomial overlap. The transformation improves the ASYS fit late in training but worsens the polynomial fit; Supplementary Table~\ref{tab:gclm-matched-conditioning} gives the field and tail errors.}
\label{fig:gclm}
\end{figure}

To separate the role of the function family from that of its fitting coordinates, we compared the polynomial and ASYS families with and without the same Gram coordinates, holding $\beta$ fixed and fitting fourteen shape coefficients from the same zero initialization for 20,000 Adam updates (Fig.~\ref{fig:gclm}d). The transformation reduced ASYS's profile error by a factor of about $21{,}600$---reaching $2.23\times10^{-8}\%$, about $32{,}200$ times smaller than the better polynomial fit---but increased the polynomial's error by a factor of about $327$. These opposite responses show that the basis functions and their fitting coordinates mattered together under the common optimizer schedule; nearly orthogonal basis functions alone did not ensure an accurate fit.

Across five independent ASYS searches from the same starting ansatz (Supplementary Fig.~\ref{fig:supp-gclm-search}), the median relative $L^2$ profile error was $3.90\times10^{-5}\%$, and the selected program reached $2.47\times10^{-8}\%$. For reference, adaptations of SIGS and FEX supplied with the fractional coordinate reached median profile errors of $0.0119\%$ and $0.00851\%$ over two four-hour searches each (Table~\ref{tab:accuracy-comparison}; Fig.~\ref{fig:gclm}c). Without evolutionary iteration, three independent coding agents---each working in a single session and given two hours with the same model, initial task and guidance---also recovered the fractional tail and stopped on their own within twelve minutes, at a median profile error of $7.83\times10^{-4}\%$ (Supplementary Table~\ref{tab:single-session-controls}). None developed the mixed basis or Gram coordinates reached in later iterations of the selected ASYS search, where further workers built on the inferred tail term to construct the basis and coefficient coordinates that fit the profile accurately.

The fractional correction connects the similarity profile to its evolution in physical time. In $f(y)=Cy^{-\alpha}+Dy^{-2\alpha}+\cdots$, both amplitudes follow from the value and derivative of $Q$ at $v=0$:
\begin{equation}
C=e^{Q(0)}A^{-\alpha-1},\qquad
D=e^{Q(0)}[Q'(0)-2Q(0)]A^{-2\alpha-1},
\label{eq:tail_ratio_analytic}
\end{equation}
where $Q'$ denotes differentiation with respect to $v$. As $t\to T^-$, any fixed position $x>0$ samples ever farther into the profile's tail, giving the constructed similarity field $\widehat\omega$ the limiting behavior
\begin{equation}
\widehat\omega(x,T^-)=-\lambda Cx^{-\alpha},\qquad
\lim_{t\to T^-}\partial_t\widehat\omega(x,t)=\lambda D x^{-2\alpha}.
\label{eq:gclm-tail-payoff}
\end{equation}
Thus $C$ determines the limiting field and $D$ its rate of approach, supplying a dynamical prediction that the starting ansatz cannot express. Across fixed positions $x>0$ away from the singularity (Fig.~\ref{fig:gclm}b), the approach rate extracted from the ASYS formula agrees within $0.0025\%$ with the rate required by the PDE's far-field balance for the fitted leading tail; Supplementary Section~\ref{supp:tail-applicability} derives these limits and the equation's relation among $C$, $D$ and $\alpha$.

\FloatBarrier

\subsection{Fujita: combining diffusion with a local reaction response}
\label{sec:fujita-transient}

The next study isolates a single revision made during search, in which ASYS replaced the reaction response of its own earlier candidate. We compare the two responses within the same ansatz while holding the parameter count and equation-based fitting procedure fixed. The task follows a radially symmetric field in which diffusion spreads the profile while cubic reaction concentrates it into an increasingly narrow, rapidly growing core~\citep{quittner2019superlinear}. The Fujita equation~\citep{fujita1966blowing} for a radial field in five space dimensions is
\begin{equation}
 u_t=u_{rr}+\frac4r u_r+u^3,\qquad
 u(r,0)=5\cos(\pi r/10),\qquad 0\leq r\leq5,
\label{eq:fujita-transient}
\end{equation}
with symmetry at the origin and $u(5,t)=0$ at the outer wall. The time window ends at $0.999\,T_\ast$, where the blow-up time $T_\ast=0.020406$ was estimated beforehand from a numerical solution and supplied with the task. ASYS began from a supplied concentrating cosine, $u_{\mathrm{start}}=5\cos(\pi r/10)\sqrt{(T_s+br^2)/(T_s-t+br^2)}$, whose two fitted parameters $T_s$ and $b$ control its time scale and radial width. Although this starting ansatz satisfies the initial and boundary conditions, it grows at every interior radius, whereas the PDE requires initial decay near the outer wall; an accurate representation must therefore capture central amplification and outer decay together.

For the reaction-only equation $u_t=u^3$, an initial amplitude $a$ evolves as $u(t)=a/\sqrt{1-2a^2t}$. ASYS used the factorization $u=v/\sqrt q$ to combine this square-root blow-up factor with diffusive spreading, placing heat evolution (with a boundary correction) in the numerator $v$ and reaction-driven amplification in the positive denominator $q$. Early candidates adjusted the profile width through a time-only logarithmic response; ASYS subsequently derived a spatially varying response $F$ from the PDE so that the correction tracks the differing local reaction rates across the profile. With $c=\cos(\pi r/10)$, the denominator separates central amplification $C(t)$ from a radial width factor $W(r,t)$:
\begin{equation}
 q=c^2C(t)+(1-c^2)W(r,t),\qquad
 W(r,t)=\frac{e^{E(r,t)}}{1+\gamma F(t/T,c)},
\label{eq:fujita-factorization}
\end{equation}
where $E$ describes the evolving width, $\gamma$ sets the response strength and $T$ is a fitted time scale. ASYS constrained the core and width factors to match the first three initial interior time derivatives required by the PDE. The reaction-only solution shows why a time-only response falls short: a point with initial amplitude $5c$ would blow up at $t=1/[2(5c)^2]$, so the reaction runs fastest at the center and ever more slowly toward the wall. Approximating the gradient term $-3q_r^2/(2q)$ in the equation that $q$ satisfies when $v$ solves the heat equation (Supplementary Section~\ref{supp:fujita-transient}), and substituting the leading reaction denominator $q_0=1-2(5c)^2t$, yields a contribution proportional to $t^2/q_0$ whose time integral introduces a logarithm.

Subtracting the order-$t^3$ part already fixed by the prescribed initial derivatives and introducing the scaled time $s=t/T$ gives the fourth-order local response in $W$, written in terms of the local reaction variable $sc^2$:
\begin{equation}
 F(s,c)=\int_0^s\frac{\tau^3c^4}{1-\tau c^2}\,d\tau
 =\frac{-\log(1-sc^2)-sc^2-(sc^2)^2/2-(sc^2)^3/3}{c^4}.
\label{eq:fujita-local-response}
\end{equation}
This function satisfies $F=s^4c^4/4+O(s^5)$ and vanishes smoothly at the outer wall; Supplementary Section~\ref{supp:fujita-transient} gives the complete core and width construction.

Table~\ref{tab:fujita-history} follows the search; its errors are those of the candidates as the search fitted them, and the controlled comparison of the two responses comes after it. Introducing the heat--reaction factorization in the first iteration reduced trajectory error from $55.5\%$ for the fitted starting ansatz to $2.38\%$. Across iterations 2--5, enforcing higher initial time derivatives and wall conditions improved the equation score by up to a factor of five, yet trajectory error remained between $3.1\%$ and $3.5\%$ while candidates retained a time-only response. Replacing that uniform term with the local response $F(s,c)$ in the sixth iteration lowered trajectory error from $3.47\%$ to $0.0879\%$.

\begin{table}[htbp]
\centering\small
\caption{Selected candidates from the Fujita search, each evaluated with the coefficients fitted during search. The equation score is the evaluator's geometric-mean normalized residual (lower is better). Field errors are relative $L^2$ errors in percent against the numerical reference, measured after the search and never returned to it. The sixth-iteration candidate descends directly from the fourth; the fifth-iteration candidate is a branch that was not continued.}
\label{tab:fujita-history}
\begin{tabular}{rlrrr}
\toprule
Iteration & Change & Equation score & Trajectory (\%) & Final time (\%)\\
\midrule
0 & Supplied concentrating cosine & $1.19\times10^{-1}$ & $55.5$ & $2382$\\
1 & Heat/reaction factorization $u=v/\sqrt q$ & $7.43\times10^{-4}$ & $2.38$ & $16.3$\\
2 & Second initial time derivative from the PDE & $4.88\times10^{-4}$ & $3.10$ & $20.6$\\
3 & Third initial time derivative & $3.10\times10^{-4}$ & $3.42$ & $22.4$\\
4 & Wall responses; time-only response $F(s,1)$ & $1.42\times10^{-4}$ & $3.47$ & $22.7$\\
5 & Core curvature modes & $1.84\times10^{-4}$ & $3.46$ & $22.5$\\
6 & Local response $F(s,c)$ & $1.37\times10^{-5}$ & $0.0879$ & $0.702$\\
\bottomrule
\end{tabular}
\end{table}

\begin{figure}[htbp]
\centering
\includegraphics[width=\linewidth]{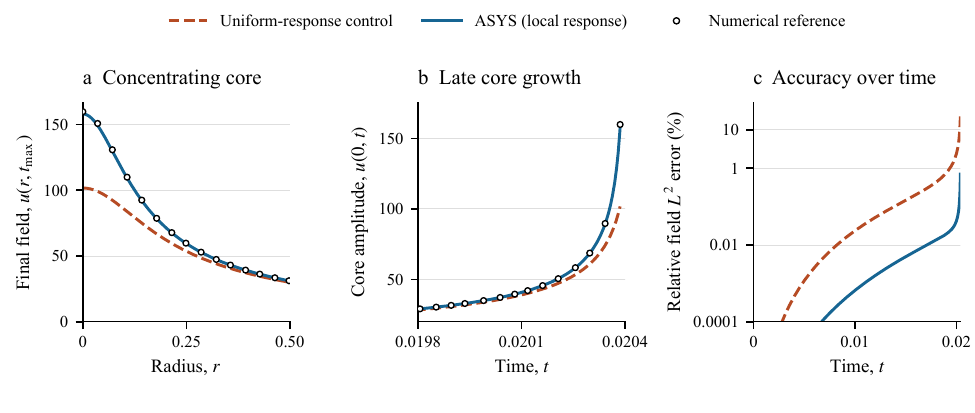}
\caption{\textbf{Local and uniform reaction responses after matched fitting.} The uniform-response control reinstates the time-only response $F(s,1)$ that ASYS replaced with $F(s,c)$. Both forms are fitted with fourteen scalars under identical initialization, loss, optimizer and fitting allowance. \textbf{a,} Final radial profile, enlarged around the core ($0\leq r\leq0.5$). \textbf{b,} Growth at the origin over the final portion of the trajectory; open circles mark the numerical reference. \textbf{c,} Relative radial-profile $L^2(dr)$ error over time across the full domain $0\leq r\leq5$, plotted in percent on a logarithmic axis.}
\label{fig:fujita-local-response}
\end{figure}

Because the fourth- and sixth-iteration candidates differed in both their response function and its initial strength, we fitted $F(s,c)$ and $F(s,1)$ on equal terms within the developed fourteen-parameter construction, starting from identical initial coefficients under the same equation-based procedure. The local response achieved $0.0879\%$ relative space--time $L^2$ error, compared with $3.99\%$ for the uniform response (Table~\ref{tab:accuracy-comparison}; Fig.~\ref{fig:fujita-local-response}). With the parameter count, initialization and fitting procedure held fixed, the improvement came from the equation-derived response in two ways: at the coefficients of the local fit it lowers trajectory error directly by a factor of $3.32$, and during fitting it gives a reaction term that the fit keeps, whereas the strength of the uniform response is driven to nearly zero (Supplementary Table~\ref{tab:fujita-frozen-responses}).

For reference, a 12,739-parameter physics-informed neural correction of the two-parameter starting ansatz, trained on 5,760 collocation points, reached a median trajectory error of $5.30\%$ across three fits with a one-hour allowance each (Supplementary Section~\ref{supp:fujita-neural-control}). Three single-session coding agents, given the same starting task and a two-hour allowance, stopped on their own within twelve minutes; the lowest trajectory error among them was $0.222\%$, with a lower final-time error than the ASYS construction (Supplementary Table~\ref{tab:single-session-controls}).

\FloatBarrier

\subsection{Allen--Cahn: curvature couples interface shape modes}
\label{sec:allen_cahn}

The Allen--Cahn equation models phase-boundary evolution in materials, such as domain ordering and coarsening in alloys~\citep{allen1979microscopic}. Its scalar field $u$ takes values near $+1$ and $-1$ in the two phases, separated by a diffuse transition layer of thickness $\epsilon$. Because the moving interface determines how the domains shrink and deform, its position and shape provide natural coordinates for an analytical approximation. With spatial Laplacian $\Delta=\partial_{xx}+\partial_{yy}$, the equation reads
\begin{equation}
u_t=\epsilon^2\Delta u+u-u^3,
\qquad (x,y)\in[-1,1)^2,\quad t\in[0,8],\quad \epsilon=0.04.
\label{eq:allen_cahn}
\end{equation}
On the periodic square, two overlapping diffuse disks form an asymmetric peanut-shaped domain whose waist fills in as the boundary retreats. Mean-curvature motion provides a leading-order geometric description~\citep{allen1979microscopic,rubinstein1989fast}, but two independently shrinking circles cannot capture this changing shape.

\begin{figure}[htbp]
\centering
\includegraphics[width=\linewidth]{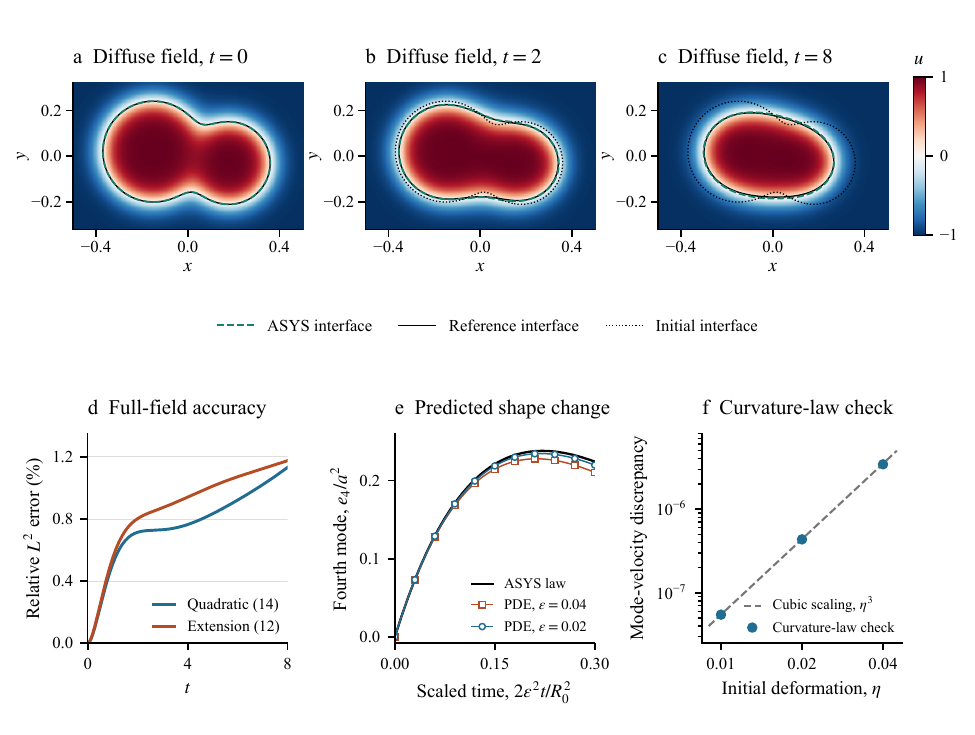}
\caption{\textbf{Curvature couples the components of an evolving shape.}
\textbf{a--c,} The complete fourteen-parameter field as the peanut rounds off and contracts. Dashed teal, solid black and dotted black contours mark the ASYS, reference and initial interfaces. The display is cropped around the evolving phase.
\textbf{d,} Relative field errors on the full periodic square across all 262 saved times. Trapezoidal time weighting gives trajectory errors of $0.798\%$ for the quadratic construction and $0.916\%$ for its twelve-parameter extension.
\textbf{e,} Generated fourth harmonic in independent PDE evolutions compared with the ASYS law~\eqref{eq:ac-fourth-mode}. Both simulations start with $a=0.08$, $b=0$ and area radius $R_0=0.28$. The horizontal axis shows scaled physical time, $1-s=2\epsilon^2t/R_0^2$, with $c=\epsilon^2$ in the prediction; the vertical axis normalizes the measured fourth-mode amplitude by $a^2$. A model with independent mode decay leaves this harmonic identically zero.
\textbf{f,} Discrepancies between quadratic mode velocities and the exact polar curvature law for initial amplitudes $\eta(1,0.2,-0.3,0.1,0.1)$ in modes two through six. The dashed line indicates cubic scaling anchored at the smallest deformation. This panel tests the curvature expansion, whereas panel d measures diffuse-field accuracy.}
\label{fig:ac-mode-coupling}
\end{figure}

Starting from the static initial field and supplied curvature-motion guidance (Supplementary Section~\ref{supp:workspace-example}), ASYS constructed an approximation that tracks the peanut as its neck smooths out into a contracting oval (Fig.~\ref{fig:ac-mode-coupling}a--c), using a hyperbolic-tangent profile across the narrow transition layer~\citep{fife1977travelling}. To connect the initial two-lobed region to an evolving polar curve, ASYS corrected the normal distance to prevent layer stretching, introduced a localized neck displacement that starts quadratically in time---preserving both the initial field and its initial velocity---and added a short-lived transient matching the initial PDE evolution.

The polar interface representation separates overall shrinkage from changes in shape, writing the radius at angle $\vartheta$ for a nearly circular boundary as
\begin{equation}
r(\vartheta,t)=\frac{R(t)\left[1+\sum_{m=2}^{6}e_m(t)\cos(m\vartheta)\right]}
{\sqrt{1+\frac12\sum_{m=2}^{6}e_m(t)^2}},
\qquad R(t)^2=R_0^2-2ct,
\label{eq:ac-polar-interface}
\end{equation}
where $c$ is the curvature-speed coefficient and the denominator normalizes the enclosed area to $\pi R(t)^2$. The harmonics $e_m(t)$ represent angular departures from a circle, from elliptical elongation ($m=2$) to finer higher-mode deformations. Of the fourteen fitted parameters, seven belong to this curve ($R_0$, $c$ and the five initial amplitudes $e_2(0),\ldots,e_6(0)$), three set its center and orientation, and four set how fast the initial peanut blends into the polar curve, the neck displacement and the decay of the initial transient (Supplementary Table~\ref{tab:ac-curvature-coefficients}).

Whereas an earlier candidate let each harmonic decay independently, ASYS expanded the polar curvature law to second order in the mode amplitudes and analytically integrated the resulting couplings, allowing one shape component to drive another. For initial elongation $a=e_2(0)$ and fourth-harmonic amplitude $b=e_4(0)$, the quadratic forcing from the second mode gives
\begin{equation}
e_4(t)=b s^7+\frac{11}{20}a^2(s^2-s^7),
\qquad s=\frac{R(t)^2}{R_0^2}.
\label{eq:ac-fourth-mode}
\end{equation}
Uncoupled linear decay retains only $b s^7$; the quadratic term generates a fourth harmonic even when $b=0$, with its strength $11/20$ fixed by curvature rather than by an extra fitted parameter. The full quadratic construction similarly couples sum and difference harmonics through $m=6$, so parameter fitting selects the overall geometry while the derived couplings determine how the modes evolve together.

A separate set of PDE simulations, started from purely elongated interfaces ($b=0$) rather than from the peanut, confirmed this predicted fourth harmonic. Doubling the initial elongation from $a=0.04$ to $0.08$ approximately quadrupled the generated $m=4$ amplitude, matching the $a^2$ scaling of Eq.~\eqref{eq:ac-fourth-mode}. With the curvature speed fixed at $c=\epsilon^2$, the predicted amplitude agreed with the PDE evolution to within $2\%$ at $s=0.7$ for $\epsilon=0.02$ (Fig.~\ref{fig:ac-mode-coupling}e), while the comparison at $\epsilon=0.04$ shows the diffuse interface approaching the sharp-interface curvature law as $\epsilon$ decreases (Supplementary Table~\ref{tab:ac-generated-harmonic}).

Direct comparison against the exact polar curvature law further verified the quadratic truncation: across successively halved small deformations, the fourteen-parameter construction's mode-velocity discrepancy dropped from $3.44\times10^{-6}$ to $4.35\times10^{-7}$ and $5.48\times10^{-8}$, a reduction by a factor of about eight per halving, consistent with a cubic remainder (Fig.~\ref{fig:ac-mode-coupling}f).

Evaluated over the full evolving peanut---including the initial transient and neck displacement---the diffuse field achieved $0.798\%$ relative $L^2$ trajectory error and $1.13\%$ error at $t=8$ against the numerical reference (Fig.~\ref{fig:ac-mode-coupling}d). A twelve-parameter extension from the same search, which generates modes five and six from lower harmonics instead of fitting their amplitudes and includes selected cubic feedback, attained a slightly smaller equation residual but a larger trajectory error of $0.916\%$ (Supplementary Section~\ref{supp:ac-mode-coupling} and Supplementary Table~\ref{tab:ac-curvature-coefficients}). A 12,801-parameter physics-informed neural field fitted under the same L-BFGS settings achieved lower field error, with a median trajectory error of $0.122\%$ over three 20-minute fits (Supplementary Section~\ref{supp:ac-current-neural-controls}), whereas the fourteen-parameter construction provides an explicit law for how the interface modes interact.

\FloatBarrier

\subsection{Further constructions}
\label{sec:further-constructions}

Six further studies apply the same process to a wider range of problems; Supplementary Table~\ref{tab:construction-overview} lists the problem, the construction and its outcome for each. Two build exactly conserved quantities or directly readable values into the formula. For Keller--Segel chemotaxis, a positive Gaussian--Laguerre mixture preserves total mass and the second moment exactly while giving the enclosed mass in closed form, with $0.0058\%$ density error for the illustrated construction (Methods Section~\ref{methods:ks}). For the stationary similarity profile of the Fujita equation, a compactified expression joins a regular core to an algebraic tail and agrees with the reference at its numerical resolution, with origin and tail amplitudes read directly from the formula. Two porous-medium profiles show structural repairs of the kind seen in gCLM. A waiting-time profile, in which the edge of the fluid stays at rest before it starts to move, retains a logarithmic term that smooth fits cannot represent. A Graveleau profile, which describes a shrinking hole, removes a constant far-field offset through a fractional coordinate taken from the equation's balance. The last two concern wave interaction and inverse inference. For a periodic nonlinear Schr\"odinger wave, ASYS constructed an explicit field for the radiation at the boundary and a perturbative correction for its interaction with the wave. In a Hamilton--Jacobi inverse problem, an exponential time basis suggested by the equation reduced the error of the predicted future field from $3.36\%$ to $1.06\%$.

\FloatBarrier

\section{Discussion}\label{sec:discussion}

ASYS provides a way to develop incomplete mathematical descriptions of PDEs into explicit approximations that support analysis. Supplied guidance and the language model's mathematical knowledge provide the starting ideas, while coding agents construct revisions to both the symbolic expression and its fitting procedure and test each one against the equation. Across the reported problems, this loop produces compact approximations alongside analytical predictions---such as the approach rate near a singularity and a mode-coupling law---that are absent from the starting descriptions.

In this workflow, the governing equation serves both to test a candidate formula and to guide its revision. The equation residual can point to a missing mechanism, whose functional form is derived symbolically while remaining degrees of freedom are calibrated numerically. As the gCLM conditioning comparison shows, constructing workable coefficient coordinates is itself part of the modeling problem. Symbolic deduction and numerical optimization therefore need not have a fixed boundary: the agent can develop the representation and the means of fitting it together.

Several limits define the scope of the present results. Our evidence comes from selected, mathematically informed problems, and how reliably agentic search performs across a wider range of problems, over repeated searches or with less guidance remains open. When field accuracy is the only aim, a large neural network can be more accurate: in Allen--Cahn, the 12,801-parameter PINN baseline achieves lower field error, whereas the fourteen-parameter construction supplies an explicit mode-interaction law.

The experiments also show a limit of searching by the residual: the sampled equation residual does not always rank candidates in the same order as the reference error. In Fujita, parameter fitting lowered the residual while raising field error, and the twelve-parameter Allen--Cahn extension achieved a smaller residual but larger trajectory error than the fourteen-parameter field. Because ASYS returns explicit formulas rather than network weights, candidates can be inspected beyond their residuals---for example, in one gCLM search the sampled profile error was $5.35\times10^{-5}\%$, yet the tail coefficient read from the formula fell about $10.6\%$ short of the equation's balance (Supplementary Table~\ref{tab:gclm-tail}). At the same time, numerical and asymptotic checks do not provide rigorous solution-error bounds.

For more complex dynamics, explicit constructions of singular cores or moving interfaces could be coupled to spectral~\citep{trefethen2000spectral} or neural~\citep{karniadakis2021pinns} representations of the surrounding field, with agents constructing their interaction and using the combined expression to answer a scientific question. Combining these expressions with computer-assisted verification could further connect agent-constructed approximations to rigorous error bounds~\citep{nakao2019verification,chen2023euler}.

\section{Methods}\label{sec:method}

\subsection{Ansatz construction and fitting}\label{subsec:dsp}

For candidate $c$, let $\mathcal{R}_c$ denote the ansatz, $\mathcal{L}_c$ its scalar training loss and $\mathcal{A}_c$ the fitting procedure. Given initial coefficients $\theta_{c,0}$ and a fitting allowance $B_c$, the fitted field and its evaluation are
\begin{equation}
\widehat{\theta}_c=\mathcal{A}_c(\mathcal{L}_c,\theta_{c,0},B_c),
\qquad
\widehat{u}_c=\mathcal{R}_c(\,\cdot\,,\widehat{\theta}_c),
\qquad
\mathbf{s}_c=\mathcal{S}_{\mathcal{P}}(\widehat{u}_c),
\label{eq:construction-and-evaluation}
\end{equation}
where $\mathcal{S}_{\mathcal{P}}$ is the evaluator for the scientific task $\mathcal{P}$. The evaluator is specified separately from the candidate and held fixed within the search stage.

An ansatz that satisfies the initial condition analytically can devote its training loss to the evolution equation while the evaluator still checks both, and coordinate transformations or basis changes can improve the conditioning of the fit. Table~\ref{tab:method-fitting} specifies the fitting procedure used in each study.

\subsection{Evolutionary search with EvE}\label{subsec:iterative-refinement}

EvE~\citep{eve2026} maintains separate populations for candidate programs and written guidance, together with evaluation feedback. Guidance records mathematical strategies, fitting advice and lessons from previous attempts.

Each iteration assigns a guidance entry to each worker and supplies examples sampled from the program population. A sampled program initializes the editable workspace. Each worker reads the scientific task, the sampled examples and their diagnostics, works out a useful change, edits the program and tests the resulting construction. It returns one candidate for formal evaluation and may also revise the guidance to record reusable findings. Evaluation executes the candidate's declared construction and fitting procedure, or reloads its saved state, before computing the task score. Completed candidates and their feedback become available to later iterations. An iteration comprises a batch of worker attempts, each of which may include several local edits and numerical tests.

Program examples are sampled using EvE's rank-based rule. For the configured score components $d$, candidate $c$ receives weight
\begin{equation}
w(c)=\sum_d \alpha_d\exp\!\left[-r_d(c)/T_d\right],
\label{eq:supp-selection}
\end{equation}
where $r_d(c)$ is its rank, with zero assigned to the best score, $\alpha_d$ is the component weight and $T_d>0$ controls the preference for high-ranked candidates. Examples are drawn without replacement using these weights. This allows programs with different strengths to contribute to subsequent constructions. The archive retains admitted candidates, including those later surpassed in score.

Guidance is sampled by its rating. When an iteration produces multiple scored candidates, EvE compares their task scores pairwise and assigns Elo credit to the guidance used by each worker. A revised guide receives the rating associated with that worker's result and can guide later attempts. ASYS retains EvE's sampling and credit mechanisms, while each scientific task definition specifies the admissible ansatz space, fitting interface and evaluator.

\subsection{Research coordination}\label{subsec:coordination}

Researchers choose how much of an investigation to specify: they may supply only a scientific objective and resource budget, or also define the construction tasks, starting programs, mathematical context and search plan. The coordinator works within this specification and handles the decisions delegated to it. It can examine the returned programs alongside their numerical behavior, using mathematical reasoning and targeted calculations to identify promising structures, diagnose failures and develop follow-up questions.

In the experiments reported here, the human researcher prepared the task READMEs and configurations with AI assistance and reviewed the setup before execution. Coordinators were instructed to pursue scientifically significant constructions and managed search continuation and assessment within that setup. Solver agents within the ASYS searches developed the mathematical constructions.

An investigation can therefore span several searches. The coordinator can carry programs and guidance into a new task, compare alternative fitting procedures, or request a focused correction after identifying a structural defect. The coordinator can also organize assessments of field accuracy, asymptotic behavior, conservation and evaluation cost. Those assessments inform subsequent research, while each search stage uses its specified evaluator to compare candidates.

\subsection{Scientific evaluation and comparisons}\label{subsec:scoring}

For the time-dependent PDE tasks, evaluation measures the governing equation and its initial and boundary conditions. Given $u_t=F[u]$, initial field $u_0$ and boundary operator $\mathcal{B}$, the pointwise residuals are
\begin{equation}
R_{\rm phys}=\partial_t\widehat{u}-F[\widehat{u}],
\qquad
R_{\rm ic}=\widehat{u}(x,0)-u_0(x),
\qquad
R_{\rm bc}=\mathcal{B}[\widehat{u}].
\end{equation}
The boundary operator measures the prescribed value or flux, or the mismatch between paired periodic boundaries. An additional initial-time check measures whether the approximation begins to evolve in the direction prescribed by the equation. For the Allen--Cahn implementation, it uses
\begin{equation}
\ell_{\rm comp}
=\frac{1}{N_0}\sum_{i=1}^{N_0}
\left|
\frac{\widehat{u}(x_i,h)-\widehat{u}(x_i,0)}{h}
-F[u_0](x_i)
\right|^2,
\label{eq:initial-tangent}
\end{equation}
with $h=0.008$ in the curvature-mode search. Spatial derivatives in $F[u_0]$ are computed by automatic differentiation. This secant check exposes a time-independent approximation that matches $u_0$ but has the wrong initial evolution.

Allen--Cahn averages squared strong-form residuals over fixed evaluation points. The four components are reported separately, with signs chosen so that larger scores are better. The sampling and aggregation rules are given in Section~\ref{supp:score}.

Because the real-line gCLM task seeks a stationary self-similar profile and its scaling exponent, without prescribed time-dependent initial data or periodic boundaries, its evaluator scores the similarity equation together with the origin conditions $f(0)=0$, $f'(0)=1$ and $g(0)=1$, where $g=-Hf$. The squared equation imbalance is divided by the sum of squared equation terms, measuring relative cancellation across the profile's widely varying scales. Averaging over logarithmic radius gives comparable weight to successive spatial scales; normalization fixes the profile's otherwise arbitrary amplitude and spatial scale. These terms form one scalar search score, also used by the common profile assessor for the SIGS/FEX comparisons.

Numerical PDE references provide a separate measure of field accuracy. For samples indexed by $i$ with quadrature weights $q_i$, the relative error is
\begin{equation}
E_{L^2}
=\left(
\frac{\sum_i q_i|\widehat{u}_i-u_{{\rm ref},i}|^2}
{\sum_i q_i|u_{{\rm ref},i}|^2}
\right)^{1/2}.
\label{eq:method-field-error}
\end{equation}
Reported relative field and profile errors are $100E_{L^2}$ in percent. The field and profile errors in Supplementary Table~\ref{tab:construction-overview} use each study's stated sampling domain and weights. The gCLM metric samples logarithmic radii.

\subsection{Statistics and reproducibility}\label{subsec:reproducibility}

Repeated searches and repeated coefficient fits are reported separately. Tables identify the number and type of repeats and give the individual outcomes underlying the reported medians. Search stopping is given in Table~\ref{tab:supp-run-config} and gCLM candidate selection in Section~\ref{supp:shared}; fitting seeds, sampling and state-selection rules are specified with each comparison in Section~\ref{method:comparisons}. The comparisons are descriptive; no significance tests or confidence intervals are used.

In the forward-solution studies, the searches scored the governing equations and constraints. The coordinator used reference comparisons for scientific assessment; reference values and accuracy scores remained outside the solver agents' fitting objectives and solver-facing search guidance. The exception is the transient Fujita time scale: a numerically estimated blow-up time set the evaluation horizon, the starting ansatz and the core sampling (Section~\ref{supp:fujita-setup}). The supplementary Hamilton--Jacobi inverse task additionally fitted supplied observations (Section~\ref{methods:hj}).

Comparisons evaluated starting ans\"atze, ASYS outputs and alternative methods on common targets within each study. Fixed-form fitting controls tested whether additional coefficient optimization explained a gain. Neural controls and the SIGS/FEX adaptations compared different ansatz families. Fitting allowances, parameter counts and repeated fits are specified with those comparisons. The reported search and fitting budgets varied across methods. Sections~\ref{supp:eval}, \ref{method:comparisons} and~\ref{methods:additional} give the reference solutions and study-specific protocols.

\subsection{Experimental implementation}\label{subsec:computational-cost}

The searches in Table~\ref{tab:supp-run-config} and the single-session controls ran in the Codex CLI agent harness using the \texttt{gpt-6.1-sol} model at the extra-high reasoning setting. Codex provided the agent harness for tool use within each session; EvE organized the search across sessions. Numerical construction, fitting and scoring used NUS high-performance computing resources. Table~\ref{tab:method-fitting} summarizes the fitting procedures for the field constructions. Section~\ref{methods:additional} specifies the additional profile and inference tasks.

Allen--Cahn constructed geometric fields and derived interactions between their shape modes; the curvature calculation was assessed separately from the full PDE residual. Reference-field comparisons used each program's frozen coefficients and quadrature data. Search settings, continuation histories and stopping decisions are recorded in Section~\ref{supp:shared}.

\begin{table}[htbp]
\centering
\small
\caption{Numerical fitting and evaluation for the three main studies and the supplementary Keller--Segel study. Allowances apply to individual fits. Further constructions in Supplementary Table~\ref{tab:construction-overview} are described in the corresponding supplementary sections.}
\label{tab:method-fitting}
\begin{tabularx}{\linewidth}{@{}>{\raggedright\arraybackslash}p{0.18\linewidth}>{\raggedright\arraybackslash}X>{\raggedright\arraybackslash}X@{}}
\toprule
Study & Coefficient fitting & Evaluation target \\
\midrule
gCLM profile & Adam with cosine schedule, at most 20,000 updates; 20-minute search fits and 60-minute matched-conditioning controls & Similarity-equation residual and origin normalization \\
Fujita evolution & Fourteen-scalar construction fitted by L-BFGS with a 300-second allowance & Radial equation, initial field and boundary constraints \\
Allen--Cahn interface modes & Twelve--fourteen scalars; agent-chosen loss and optimizer within an evaluator-owned loop, with a 300-second allowance & Equation, initial value, initial secant and periodic boundary; separate curvature-law check \\
Keller--Segel & Damped Gauss--Newton on the equation residual, fitting mixture and shape coefficients; 300-second allowance & Radial equation and initial and boundary constraints \\
\bottomrule
\end{tabularx}
\end{table}

\FloatBarrier

\subsection{Search configuration and evaluation}\label{supp:shared}

Table~\ref{tab:supp-run-config} summarizes how the field investigations were set up and carried out. The researcher supplied the equations, constraints, starting workspaces and fixed evaluators, with AI assistance and human review. The table distinguishes the study-specific mathematical inputs from the subsequent ASYS contributions; each iteration denotes a batch of worker attempts (Section~\ref{subsec:iterative-refinement}), and Section~\ref{methods:additional} specifies the additional profile tasks.

\begin{table}[htbp]
\centering\small
\caption{Inputs and execution of the ASYS investigations. Agent hours are summed elapsed times of completed coding sessions, including tool use within those sessions; they exclude separately run coefficient fitting and evaluation. Input/output tokens are in millions; input includes cached tokens. Hours and token counts cover the listed stages and exclude coordinator work and subsequent scientific assessments. Individual fitting allowances are given in Table~\ref{tab:method-fitting}.}
\label{tab:supp-run-config}
\setlength{\tabcolsep}{3pt}
\renewcommand{\arraystretch}{1.13}
\begin{tabularx}{\linewidth}{@{}>{\raggedright\arraybackslash}p{.09\linewidth}YYYYY@{}}
\toprule
Study & Supplied starting mathematics & ASYS constructed & Reuse across the investigation & Agent time and tokens & Stopping decision \\
\midrule
gCLM & Two-parameter power-envelope starter; compactification and Hilbert-transform identities & Fractional tail corrections, analytic bases and fitting coordinates & Each search's own programs and public evaluations; initial guide retained & Five searches, 81 iterations; 22.73 h; $124.19/2.190$ M tokens & Each search reached its nine-hour wall-time limit \\
\addlinespace
Fujita & Two-parameter concentrating cosine; initial-time incompatibility identified in the task & Heat--reaction representation with a spatially dependent reaction response & Programs and guidance from the same search & Six iterations; 2.26 h; $12.57/0.189$ M tokens & Coordinator stopped after two reviewer agents, each reading the result independently, judged it sufficient \\
\addlinespace
NLS & Breather, correction templates and boundary-radiation equation & Explicit radiation, nonlinear response hierarchy and defect correction & Continuation imported three programs and two guides; fixed-control follow-up reused the response bank & Four plus two search iterations, one fixed-control iteration; 5.13 h; $32.88/0.503$ M tokens & Coordinator paused the continuation for assessment, then ended the fixed-control follow-up after review \\
\addlinespace
Allen--Cahn & Static initial field; signed-distance, curvature and peanut-to-oval guidance & Moving-interface field and derived interactions between shape modes & Programs and guidance from the same search & Six iterations; 2.54 h; $17.51/0.192$ M tokens & Coordinator stopped after two reviewer agents, each reading the result independently, judged it sufficient \\
\addlinespace
Keller--Segel & Neural starter; Gaussian initial density, mass and moment identities & Positive Gaussian mixture, then a constrained radial-shape extension & Shape continuation imported nine programs and fitted states & Gaussian refinement and shape search: four iterations each; 3.57 h; $21.39/0.386$ M tokens & Population frozen for numerical density comparisons \\
\bottomrule
\end{tabularx}
\end{table}

The gCLM, Fujita and Allen--Cahn searches had 40-iteration ceilings. gCLM sampled one program and one guidance entry per worker, with a 60-minute worker-invocation limit.

For each gCLM search, we retained the candidate with the best equation-residual score among those passing the evaluator's checks. If two candidates had identical scores, we kept the one from the earlier iteration. The selected candidates came from iterations 16, 16, 10, 19 and 7 in Runs 1--5, respectively. The coordinator stopped the Fujita search after its sixth iteration (Table~\ref{tab:supp-run-config}); the reported construction is the candidate returned in that iteration, a direct descendant of the fourth-iteration candidate with the uniform response.

Each iteration used one worker in gCLM and two in Fujita, Allen--Cahn and Keller--Segel. The five gCLM searches ran concurrently for about nine hours each, summing to $45.02$ hours of wall time across agent sessions, coefficient fitting and evaluation ($22.16$ CPU hours, including unfinished final work). The Keller--Segel hours and token counts cover the Gaussian-refinement and shape-extension stages; its initial neural-to-mixture construction precedes those stages.

\subsubsection{Candidate construction interfaces}\label{supp:workspace}

Workers receive the equation, constraints, editable program files, sampled examples and guidance. The gCLM task fixes $a=0.25$ and seeks an odd, positive-on-$y>0$, decaying profile with $f'(0)=g(0)=1$. Workers edit \texttt{model.py} while a protected fitter supplies the Adam schedule and equation-based loss, allowing at most 20,000 updates and 1,200 seconds within a 1,500-second numerical evaluation allowance. The supplied context includes the similarity equation, reflection symmetry, Hilbert-transform identities, coordinate compactification and a leading power envelope. The fractional correction basis and its coefficient coordinates were constructed by the agents.

Allen--Cahn permits task-specific fitting losses and optimizers within an evaluator-owned loop, with a 300-second fitting allowance. Keller--Segel likewise allows task-specific fitting losses. Separate scientific assessments use frozen fitted states.

\subsubsection{Evaluation components}\label{supp:score}

The gCLM evaluator ranks the normalized similarity residual and origin penalties in Eq.~\eqref{supp:eq:assessor-score}.

The Allen--Cahn curvature-mode search averages four signed MSE components. Its fixed evaluator uses 1,024 uniformly sampled space--time points for the equation, 1,024 initial spatial points and 128 boundary base points, with a fixed random seed. Boundary checks compare paired values and first derivatives on opposite sides of both spatial axes. The initial-time component compares the secant over $h=0.008$ with the PDE right-hand side of the initial field. Spatial derivatives and the equation residual use automatic differentiation. The Keller--Segel and additional profile evaluators are specified with their study descriptions.

\subsubsection{Sampling and retention}\label{supp:selection}

Equation~\eqref{eq:supp-selection} gives the program-example sampler. gCLM uses the physics score with unit weight and temperature. Allen--Cahn gives unit weight and temperature to the physics, initial-value, initial-tangent and boundary components. Guidance uses rank-based sampling of its Elo rating. Admitted programs remain available after later programs surpass their scores.

\FloatBarrier
\subsection{Problem setups and supplied starting ans\"atze}\label{supp:setups}
Table~\ref{tab:supp-setups} collects the equations, domains and constraints for the three main studies; the Keller--Segel setup is given in Section~\ref{methods:ks}.

For transient Fujita, the final evaluation time is $t_f=0.020385594$.

\begin{table}[htbp]
\centering
\caption{Equations, domains and constraints for the three main studies; the Keller--Segel setup is given in Section~\ref{methods:ks}.}
\label{tab:supp-setups}
\begin{tabular}{@{}p{0.14\linewidth}p{0.34\linewidth}p{0.20\linewidth}p{0.24\linewidth}@{}}
\toprule
Case & Equation & Domain & Conditions \\
\midrule
gCLM & Similarity equation~\eqref{supp:eq:profile-ode} with $a=0.25$ and unknown $\beta>0$ & $y\in\mathbb{R}$ & Odd, decaying profile with $f(0)=0$, $f'(0)=1$ and $g(0)=1$ \\
Transient Fujita & $u_t=u_{rr}+4u_r/r+u^3$ & $r\in[0,5]$, $t\in[0,t_f]$ & $u(r,0)=5\cos(\pi r/10)$; $u_r(0,t)=u(5,t)=0$ \\

Allen--Cahn & $u_t = \epsilon^2\Delta u + u - u^3$, $\epsilon=0.04$ & $(x,y)\in[-1,1)^2$, $t\in[0,8]$ & Smooth union of two tanh disks; periodic value and derivative \\
\bottomrule
\end{tabular}
\end{table}

\subsubsection{gCLM: nonlocal equation and evaluation}\label{supp:gclm}

The generalized Constantin--Lax--Majda (gCLM) equation models vortex stretching and advection in one dimension~\citep{constantin1985clm,degregorio1990,okamoto2008generalization}. Self-similar blow-up profiles $\omega(x, t) = -\lambda(T-t)^{-1} f(x (T-t)^{-\beta})$ satisfy the nonlocal profile equation
\begin{equation}
f + \beta y f' + a \lambda V f' - \lambda g f = 0,
\label{supp:eq:profile-ode}
\end{equation}
where $\beta > 0$ is the self-similarity exponent, $a$ is the advection parameter, and $\lambda = (1+\beta)/(1-a)$ normalizes the spatial velocity gradient at the origin. We define the asymptotic spatial decay exponent as $\alpha := 1/\beta$. The velocity gradient $u_x = H\omega$ is determined by the Hilbert transform on the real line:
\begin{equation}
H f(y) = \frac{1}{\pi} \, \mathrm{P.V.} \int_{-\infty}^\infty \frac{f(t)}{y - t} \, dt.
\label{supp:eq:hilbert-def}
\end{equation}
Exploiting the odd symmetry $f(-y) = -f(y)$ yields the half-line velocity gradient $g(y)$ and velocity potential $V(y)$:
\begin{align}
g(y) &= -\frac{2}{\pi} \, \mathrm{P.V.} \int_0^\infty \frac{t f(t)}{y^2 - t^2} \, dt, \qquad y > 0, \label{supp:eq:g-halfline} \\
V(y) &= \frac{1}{\pi} \int_0^\infty f(t) \log \left| \frac{y + t}{y - t} \right| \, dt. \label{supp:eq:V-integral}
\end{align}

Each candidate program implements an ansatz whose continuous parameters are fitted by minimizing the scale-invariant equation residual
\begin{equation}
\mathcal{R}(y) = \frac{\left(\sum_{i=1}^4 T_i(y)\right)^2}{\sum_{i=1}^4 T_i(y)^2},
\label{supp:eq:residual-def}
\end{equation}
with $T_1 = f$, $T_2 = \beta y f'$, $T_3 = a \lambda V f'$, and $T_4 = -\lambda g f$.

Search candidates were scored on a logarithmic grid independent of their fitting points. To locate the profile's core, the evaluator surveyed $\ell=\log y\in[-E,E]$ for $E\in\{18,36,72,144,288,576\}$. It used the $0.001$ and $0.999$ quantiles of the positive normalization measure $(2/\pi)f(e^\ell)\,d\ell$, whose total is $g(0)=1$. A refined evaluation grid was constructed from 257 uniform logarithmic nodes across this interval, extended by two log units on each side, supplemented by transition sites and mass quantiles, followed by midpoint refinement. On this grid, the assessor evaluated the logarithmic trapezoidal mean of the pointwise residual $\mathcal{R}(y)$ alongside unit-weighted boundary and gauge penalties:
\begin{equation}
\mathcal{S}_{\text{assessor}} = \operatorname{mean}_{\log y}[\mathcal{R}(y)] + f(0)^2 + (f'(0) - 1)^2 + (g(0) - 1)^2.
\label{supp:eq:assessor-score}
\end{equation}
Candidates that passed evaluation were ranked by $-\mathcal{S}_{\text{assessor}}$. Each search retained the highest-scoring candidate, with the earlier iteration breaking an exact tie. Reference errors were measured after this selection.

\subsubsection{Transient Fujita: equation and supplied starting ansatz}\label{supp:fujita-setup}

The transient Fujita task follows a concentrating field in five radial dimensions:
\begin{equation}
 u_t=u_{rr}+\frac4r u_r+u^3,\qquad
 u(r,0)=5\cos(\pi r/10),\qquad u_r(0,t)=u(5,t)=0.
\end{equation}
The evaluation interval is $0\leq r\leq5$, $0\leq t\leq0.020385594$. Its endpoint is $0.999\,T_\ast$, where $T_\ast=0.020406$ is a blow-up time estimated from a preliminary numerical solution of this problem. The same value initializes the starting ansatz's time scale and locates the core samples of the diagnostic and fitting loss, where it is written $T_c$ (Section~\ref{method:fujita-comparison}); the neural and single-session controls received it as well. Candidates could fit their own time scale, which in the reported construction remained within $5\times10^{-10}$ of $T_\ast$. This initial-value problem is distinct from the stationary profile in Supplementary Section~\ref{supp:fujita-stationary}.

The supplied starting ansatz was the concentrating cosine
\begin{equation}
 u_{\mathrm{start}}(r,t)=5\cos(\pi r/10)
 \sqrt{\frac{T_s+b r^2}{T_s-t+b r^2}},\qquad
 T_s>0.020385594,\quad b>0.
\end{equation}
Its two trainable parameters, the time scale $T_s$ and radial width $b$, were initialized at $0.020406$ and $0.01$. Although $u_{\mathrm{start}}$ satisfies the initial and boundary conditions, $\partial_t u_{\mathrm{start}}>0$ throughout $0\leq r<5$, whereas the PDE's initial time derivative is negative near the outer wall; the task instructions noted this incompatibility, and the search developed a representation combining central amplification with outer diffusive decay.

The fixed Fujita evaluator used the normalized residual
\begin{equation}
 \rho(r,t)=\frac{u_t-D-u^3}{\max(|u_t|+|D|+|u|^3,10^{-12})},\qquad D=u_{rr}+4u_r/r,\quad D(0,t)=5u_{rr}(0,t).
 \label{eq:fujita-normalized-residual}
\end{equation}
Its equation component was the geometric mean
\begin{equation}
\ell_{\rm phys}=\exp\left\{\frac1{N_f}\sum_{i=1}^{N_f}\log\max(|\rho(r_i,t_i)|,10^{-12})\right\}
\end{equation}
on $N_f=2,048$ Latin-hypercube points over the full space--time domain. The evaluation samples were fixed across candidates, using random seed 920614 independently of the fitting samples.

Initial-value error was the MSE on 512 radii, comprising both endpoints and 510 Latin-hypercube samples. The initial-tangent component was the MSE between the automatically differentiated $u_t(r,0)$ and $\Delta_5u_0+u_0^3$ on the same radii with $r=5$ excluded. The boundary component averaged the MSEs of $u_r(0,t)$ and $u(5,t)$ at 256 common times, comprising both time endpoints and 254 Latin-hypercube samples. The evaluator returned the negative of each component and their equally weighted mean as its scalar score.

A separate core diagnostic measured the mean of $\rho^2$ on 2,048 points with $\delta=T_c-t$ logarithmically sampled in $[10^{-3}T_c,10^{-1}T_c]$ and $0\le r\le10\sqrt\delta$, together with 128 origin points drawn from the same logarithmic time distribution, where $T_c=0.020406$. These samples used seed 920811. This diagnostic did not enter the scalar score. The coefficient-fitting loss is specified in Section~\ref{method:fujita-comparison}.

\subsubsection{Allen--Cahn: initial field and phase velocity}\label{supp:allen_cahn}
The two-dimensional Allen--Cahn task evolves the phase field on the periodic square $[-1, 1)^2$ over $t \in [0, 8]$:
\begin{equation}
\partial_t u = \epsilon^2 \Delta u + u - u^3, \qquad (x, y) \in [-1, 1)^2, \quad t \in [0, 8],
\label{supp:eq:allen_cahn}
\end{equation}
with interface thickness parameter $\epsilon = 0.04$ and spatial periodic boundary conditions $u(x+2, y, t) = u(x, y+2, t) = u(x, y, t)$.

The initial phase field $u(x, y, 0) = u_0(x, y)$ is prescribed as the smooth algebraic union of two circular domain inclusions:
\begin{equation}
u_0(x, y) = \frac{1}{2}\Bigl( p_1(x, y) + p_2(x, y) + 1 - p_1(x, y)p_2(x, y) \Bigr),
\label{supp:eq:ac_initial}
\end{equation}
where for each inclusion $i \in \{1, 2\}$, the localized phase profile is
\begin{equation}
p_i(x, y) = \tanh\left(\frac{R_i - r_i(x, y)}{\sqrt{2}\,\epsilon}\right), \qquad r_i(x, y) = \sqrt{(x - c_{x,i})^2 + (y - c_{y,i})^2 + 10^{-12}},
\label{supp:eq:ac_disks}
\end{equation}
with geometric centers $(c_{x,1}, c_{y,1}) = (-0.15, 0.02)$ and $(c_{x,2}, c_{y,2}) = (0.18, -0.03)$, disk radii $R_1 = 0.22$ and $R_2 = 0.18$ and a $10^{-12}$ floor on the squared radius. The scalar field transitions smoothly from $u \approx +1$ in the domain interior to $u \approx -1$ in the exterior background.

Evaluating the governing differential operator at $t = 0$ dictates the initial time derivative at strictly interior points,
\begin{equation}
\left.\partial_t u\right|_{t=0} = \epsilon^2 \Delta u_0(x, y) + u_0(x, y) - u_0(x, y)^3 \neq 0.
\label{supp:eq:ac_jet}
\end{equation}
For a phase representation $u=\tanh H$, the corresponding initial phase velocity is
\begin{equation}
\partial_tH(X,0)=\epsilon^2\Delta H_0+
\tanh H_0\bigl(1-2\epsilon^2|\nabla H_0|^2\bigr),
\qquad H_0=\operatorname{arctanh}u_0.
\label{eq:ac-phase-jet}
\end{equation}
The starting program supplied to the search was the static field $u_{\text{seed}}(x, y, t) = u_0(x, y) + t \cdot 0$; because $\partial_t u_{\text{seed}} \equiv 0$, it does not match the initial velocity required by the spatial Laplacian and cubic reaction at $t = 0$.

\FloatBarrier
\subsection{gCLM fitting coordinates}
\label{methods:gclm-coordinates}

During fitting, the selected gCLM program combines the projected powers in Supplementary Eq.~\eqref{supp:eq:F-proj} into mixed-tail modes, continuing their limits when exponents coincide. It then applies a fixed Cholesky transformation of the coefficients. The Gram matrix is averaged over $\alpha\in\{1,5/4,3/2,7/4,2\}$ using the unfitted basis geometry. All fourteen shape coordinates start at zero and $\beta$ at one; the scale $A$ is determined by normalization. The implementation stores 495 fixed normalization and basis scalars as representation constants, distinct from the fifteen fitted parameters; they follow from the basis definition and the averaged Gram matrix alone and are not adjusted during fitting. The collected coefficients in Supplementary Table~\ref{tab:gclm-h3-state} already incorporate the mode combinations and coordinate transformation.

\subsection{Allen--Cahn fitting and curvature checks}
\label{methods:ac-protocol}

The fourteen-parameter quadratic field and twelve-parameter extension in Supplementary Section~\ref{supp:ac-mode-coupling} were fitted against the equation and prescribed constraints, then evaluated with their coefficients held fixed. Their fitting loss sums the PDE, initial-value, initial-secant and periodic-boundary MSEs. A further PDE term, weighted by $0.35$, scales the spatial coordinates of its sample points by $0.55$ to concentrate them around the peanut. A common factor of $10^3$ improves numerical scaling without changing the relative weights. Each fit uses 4,096 uniform residual points and 512 initial points; the boundary counts are 64 per coordinate direction for the quadratic construction and 128 for the extension. L-BFGS uses learning rate $0.5$, strong-Wolfe line search and history size 20, retaining the smallest observed fitting loss.

The main field error uses all 262 saved reference times on the $384^2$ periodic grid, with uniform spatial weights and trapezoidal time weights. These times are $\{0\}\cup\{2^{-10},\ldots,2^{-6}\}\cup\{k/32:k=1,\ldots,256\}$. For spatial error energy $E_i$ and reference energy $R_i$ at time $t_i$, the reported percentage is
\begin{equation}
100\sqrt{\frac{\sum_iw_iE_i}{\sum_iw_iR_i}},
\qquad \sum_iw_i=8,
\end{equation}
where $w_i$ are the trapezoidal weights. The norm covers the full periodic square, including the saturated background. The four displayed frames at $t=0,2,4,8$ come from the finer $512^2$ reference; they do not define the trajectory norm.

We assess the quadratic curvature expansion separately from the diffuse field. At $t=0$, set the amplitudes of modes two through six to $\eta(1,0.2,-0.3,0.1,0.1)$, with $R_0=0.28$, $c=0.04^2$ and $\eta\in\{0.04,0.02,0.01\}$. We compare the radial velocity obtained by differentiating Eq.~\eqref{eq:ac-polar-interface} with the exact curvature velocity in Supplementary Eq.~\eqref{eq:ac-polar-curvature}. On 4,096 equally spaced angles, we project their difference onto cosine modes two through six and take the Euclidean norm of the five coefficients. Halving $\eta$ tests the order of the remainder after quadratic truncation. Separate algebraic checks verify seven selected coupling coefficients and three integrated forcing formulas.

\paragraph{Generated fourth harmonic in the PDE.}
We initialized the interface at $r(\vartheta,0)=R_0[1+a\cos(2\vartheta)]/\sqrt{1+a^2/2}$, with $R_0=0.28$, $a\in\{0,0.04,0.08\}$ and $\epsilon\in\{0.02,0.04\}$. The initial field was $\tanh[d/(\sqrt2\epsilon)]$, where $d$ is the signed Euclidean distance to that curve, positive inside. An independent Fourier ETDRK4 solver evolved the PDE on the periodic $512^2$ grid with time steps at most $1/256$. We saved 61 times through $s=1-2\epsilon^2t/R_0^2=0.7$: $t=29.4$ for $\epsilon=0.02$ and $7.35$ for $\epsilon=0.04$.

At each time, quintic interpolation and ray-root searches extracted the full zero contour at 2,048 angles. Writing $r_0=\langle r\rangle$ and $r_m=2\langle r\cos(m\vartheta)\rangle$, we measured $e_m=r_m/r_0$ and compared $e_4$ with Eq.~\eqref{eq:ac-fourth-mode}, with $b=0$ and $c=\epsilon^2$ fixed. For $a=0.08$, $\epsilon=0.02$, separate $768^2$ and half-time-step runs checked convergence. Circular initial interfaces measured spurious fourth harmonics. We also integrated the exact polar curvature equation using 128 angular Fourier nodes and fourth-order Runge--Kutta steps of $10^{-4}$ in $\tau=\epsilon^2t/R_0^2$, refined to 256 nodes and steps of $2.5\times10^{-5}$. Supplementary Table~\ref{tab:ac-generated-harmonic} reports the generated amplitudes.

\FloatBarrier
\subsection{Comparison procedures}\label{method:comparisons}

\subsubsection{Single-session coding controls}\label{method:single-session}

For gCLM and Fujita, we ran three independent Codex CLI sessions per task with the same \texttt{gpt-6.1-sol} model and extra-high reasoning setting used by the ASYS workers. Each received the scientific problem, initial mathematical guidance, unfitted starter and its equation-based diagnostics from the released implementation. As in ASYS, neural surrogates and numerical PDE solvers were forbidden as candidate representations; numerical coefficient fitting remained part of the task. Previous ASYS constructions, revised guidance and population-search instructions were excluded. Filesystem and network sandboxes prevented access to other runs, the manuscript and numerical references. Each session had a two-hour wall-time allowance, and all six agents chose to finish after 6.1--11.8 minutes (excluding subsequent coefficient fitting and evaluation) without follow-up prompting. Individual ASYS workers had time limits of 30 minutes for Fujita and 60 minutes for gCLM. All returned control programs satisfied the representation restrictions and were evaluated without repair or replacement.

Fitting used one CPU thread per run. The gCLM programs used the protected Adam schedule in Section~\ref{supp:workspace}, with seed 2, at most 20,000 updates and a 1,200-second allowance. The learning rate, collocation and quadrature settings matched the final fits in Section~\ref{method:gclm-comparisons}; the last completed update supplied the fitted state. Candidate initialization could not solve the profile equation or fit coefficients. Fujita retained its original 300-second allowance and lowest-training-loss state selection. Its interface allowed each candidate to choose the loss, sampling configuration and optimizer, so these were independently constructed fitting recipes. Numerical references were used only after the fitted states were fixed. The comparisons use the original reported ASYS fields without refitting.

\subsubsection{gCLM: SIGS and FEX adaptations}\label{method:gclm-comparisons}

The SIGS and FEX adaptations were evaluated on the same $a=1/4$ equation, normalization conditions and reference banks as ASYS. Two-slot SIGS and depth-two FEX searches each used seeds 0 and 1 and received the coordinate $v=(1+A^2y^2)^{-1/(2\beta)}$ of Eq.~\eqref{eq:h3_ansatz}, informed by the completed ASYS profile search. Their corrections, $E(v)-E(1)$ for SIGS and $E(1-2v)-E(-1)$ for FEX, admit the secondary fractional tail through a linear term in $v$. Both fitted $\beta$, initially set to one. Each search had a four-hour allocation on one CPU thread, with up to two hours for preliminary search; the four distinct structures with the lowest eligible preliminary fitting losses were retained for final fitting.

For SIGS~\citep{oikonomou2026sigs}, the correction was a sum of two one-variable expressions, each initialized with amplitude $0.1$. Numerical literals were fitted coefficients, except for integer exponents and mathematical constants. Stored expressions came from the released SIGS library; multivariate and constant expressions, and those failing interval-arithmetic checks for finiteness on $v\in[0,1]$, were excluded. Proposals balanced choices across up to five $k$-means clusters of the library's latent vectors. One broad round and two focused rounds each drew 60 expression pairs; focused rounds reclustered the groups containing the best stored proposal so far. Covariance matrix adaptation evolution strategy (CMA-ES) then searched the concatenated latent vectors, starting from that proposal with initial step size $0.5$, population 20 and at most 100 generations. The released pretrained GrammarVAE decoded each latent proposal into expressions; its library and decoder were unchanged.

The FEX adaptation~\citep{liang2025fex} represented the correction by a depth-two expression tree. Binary nodes used addition, subtraction or multiplication; unary nodes applied $a\,g(\cdot)+b$, with $g$ chosen from $0$, $1$, identity, square, cube, fourth power, $\exp$, $\sin$ and $\cos$. Coefficients $(a,b)$ were initialized independently from a zero-mean normal distribution with standard deviation $0.1$. A controller with one hidden layer of 50 ReLU units proposed operator choices in batches of ten, using sampling temperature 5 and 10\% uniform exploration. Policy-gradient updates used Adam at learning rate $0.002$, for at most 500 batches. The reward was $1/(1+10^6L)$, where $L$ was the validated preliminary fitting loss; only positive reward differences above the batch median contributed to the controller update.

Preliminary coefficient fits used 20 Adam updates at learning rate $0.001$, followed by 20 L-BFGS updates at rate $0.5$, with strong-Wolfe line search and history size 20. They sampled 97 logarithmic residual sites and 128 quadrature nodes, refreshing the candidate-dependent support every ten updates, within a 30-second allowance. Final fits continued the retained search states for 20,000 Adam updates with a cosine learning-rate schedule starting at $0.01$, 193 residual sites and 256 quadrature nodes, refreshing support every 100 updates. Each final fit had a 1,200-second allowance. The common equation score selected one of the four qualified final fits before reference errors were measured.

\subsubsection{gCLM: matched coefficient coordinates}\label{method:gclm-conditioning}

The polynomial used $Q(v)=\sum_{j=0}^{13}c_jT_j(2v-1)$, with $T_j$ the Chebyshev polynomials, in the same profile envelope and normalization as Supplementary Eq.~\eqref{supp:eq:h3-def}. Its identity coordinates were the coefficients $c_j$.

To test the role of coefficient coordinates, we compared identity and Gram coordinates for both the degree-13 polynomial and the ASYS family. All four fits fixed $\beta$ at the selected ASYS program's fitted value, $0.6845136212067134$, and optimized fourteen shape coefficients from zero, so their initial fields and losses coincided. Both families already admitted the secondary fractional tail. Both ASYS fits retained the projected fractional powers in Supplementary Eq.~\eqref{supp:eq:F-proj}; they differed only in the final coefficient map. Each fit completed 20,000 Adam updates with the same cosine learning-rate schedule (initial rate $0.01$), 193 residual points, 256 quadrature nodes and the same adaptive support-selection rule, refreshed every 100 updates. Each had a 3,600-second allowance on one CPU thread. The reported states are the last completed updates, without selection by reference error.

For each family, the Gram transformation used the overlap of the log-profile basis functions on 768 Gauss--Legendre nodes in log radius, averaged over $\alpha\in\{1,1.25,1.5,1.75,2\}$. With $G=LL^\top$, the invertible coefficient map was $L^{-\top}$. This map changes coordinates within each family; it does not orthogonalize the PDE residual or its derivatives with respect to the fitted coefficients.

Tail coefficients were extracted analytically at $v=0$ using 80-digit arithmetic. The complete profile and tail results are reported in Supplementary Table~\ref{tab:gclm-matched-conditioning}.

\subsubsection{Fujita: local and uniform responses}\label{method:fujita-comparison}

The construction in Supplementary Section~\ref{supp:fujita-complete-formula} has fourteen fitted scalars: the time scale $T$, three core coefficients $a_j$, nine width coefficients $b_{ij}$ and $\gamma$. During fitting, each $a_j$ and $b_{ij}$ was parameterized as one half of a hyperbolic tangent; $T$ was constrained above the final time and $\gamma>0$. The matched comparison in Fig.~\ref{fig:fujita-local-response} replaced only the local width response $F(s,c)$ with the fourth-iteration candidate's time-only response $F(s,1)$. Table~\ref{tab:fujita-history} reports selected candidates as the search evaluated them, each with its own 300-second fit and the fixed evaluator's equation score; the coordinator measured their field errors afterwards in the same norms, without returning them to the solver agents. Both fits started from $T=0.020406$, $\gamma=3(\pi/10)^2/25$ and zero shape coefficients, and used identical losses, collocation points and L-BFGS settings.

With $\rho$ defined in Eq.~\eqref{eq:fujita-normalized-residual}, the fitting objective was the mean of $\rho^2$ on 512 Latin-hypercube points over the full space--time domain, plus $0.2$ times its mean on 208 core points. Of these, 192 sampled the remaining time $\delta=T_c-t$ logarithmically over $[10^{-3}T_c,10^{-1}T_c]$ and a radius in $[0,10\sqrt\delta]$, where $T_c=0.020406$; the remaining 16 lay at the origin with the same logarithmic time distribution. The fixed full-domain and core samples used seeds 107014 and 107015, respectively.

Both fits used float64 arithmetic on one CPU thread and a 300-second allowance. L-BFGS used learning rate $0.5$, strong-Wolfe line search, up to 20 internal iterations per step, \texttt{max\_eval}=25 and gradient/change tolerances $10^{-9}$ and $10^{-12}$. Fitting was limited to 1,000 optimizer calls and stopped after 90 seconds without a lower loss. The lowest-loss evaluated closure state was returned, including line-search trial states. Numerical references were used afterward to measure field accuracy.

To separate the response change from coefficient fitting, we evaluated each response at the common initial coefficients and at each fitted coefficient vector, without further optimization. All six fields used the same reference and error norms as the fitted comparison. Their equation objectives were evaluated on the original fitting points and on the independent sample used for the neural control: 8,192 full-domain points and 8,704 core points, including 512 at the origin, with seeds 2026110596 and 2026110597. The residual normalization and weights $1$ and $0.2$ were unchanged.

\subsubsection{Fujita: neural correction and sampling density}\label{method:fujita-neural}

We fitted a physics-informed neural correction~\citep{raissi2019physics} of the same concentrating ansatz supplied to ASYS, $u_{\rm start}$ in Section~\ref{supp:fujita-setup}. It represented the field as
\[
u_N(r,t)=u_{\rm start}(r,t;T_s,b)\exp\!\left[\frac{t}{t_f}N(z,\eta)\right],
\qquad z=(r/5)^2,\qquad
\eta=\frac{-\log(1-t/T_c)}{-\log(1-t_f/T_c)},
\]
with $t_f=0.020385594$ and the fixed supplied clock $T_c=0.020406$. The network had four hidden layers of 64 tanh units and one output, giving 12,737 neural parameters. The two parameters of the supplied ansatz used $T_s=t_f+T_c\operatorname{softplus}(\theta_T)$ and $b=e^{\theta_b}$, initialized at $T_s=0.020406$ and $b=0.01$. Thus the complete model had 12,739 fitted scalars. Hidden weights and biases were initialized uniformly on $[-1/\sqrt{n_{\rm in}},1/\sqrt{n_{\rm in}}]$, where $n_{\rm in}$ is the layer's input width. The output weights and bias were zero, so every fit began from $u_{\rm start}$. The exponential correction preserves the initial field, outer wall value and origin symmetry.

The network used the normalized equation objective and the same 512 full-domain and 208 core points specified in Section~\ref{method:fujita-comparison}. These samples were fixed and shared across three network initializations.

Each fit used float64 arithmetic and one CPU thread, with a 3,600-second allowance. Adam used a cosine learning rate from $10^{-3}$ to $10^{-4}$ during the first 2,400 seconds, followed by L-BFGS with learning rate $0.5$, strong-Wolfe line search, one internal iteration, \texttt{max\_eval}=25, history size 20 and gradient/change tolerances $10^{-9}$ and $10^{-12}$. Training ended at 3,585 seconds or after 50 consecutive unchanged L-BFGS steps. The lowest training loss among the initial field and completed steps selected the returned state; numerical references did not enter fitting or selection.

Field evaluation used the same radial $dr\,dt$ norm and numerical reference as the reported Fujita construction. The neural fitting allowance differs from the 300-second local-versus-uniform response comparison. We also evaluated the residual objective on 8,192 independent uniform points and 8,704 independent core points, including 512 at the origin.

To test the effect of sampling density, we appended seven independently generated blocks to the original 720 points, giving 4,096 uniform and 1,664 core points, including 128 at the origin. This paired comparison retained the three initial model states, objective weights, architecture, optimizer, stopping rules and one-hour allowance. Supplementary Table~\ref{tab:fujita-neural-fits} reports every fit at both sampling densities; Table~\ref{tab:accuracy-comparison} uses the denser setting.

\subsubsection{Allen--Cahn: physics-informed neural field}\label{method:ac-neural}

We fitted a physics-informed neural field to the same periodic initial-value problem using the loss and L-BFGS settings of the reported ASYS construction. Three independent fits varied the initial network weights and collocation samples. The comparison uses the original ASYS field with its reported coefficients.

The neural control used four hidden layers of 64 tanh units and a scalar output, giving 12,801 trainable parameters. Hidden weights used Xavier-normal initialization with zero biases, and output weights were initialized to zero; network inputs were $(x,y,t/4-1)$. The network corrected an analytical field built from the initial two-disk union and the initial velocity required by the PDE. To define that field, let $G(X,t)$ be the phase of the algebraic union after replacing each disk radius $R_j$ and radial distance $\rho_j$ by $\sqrt{R_j^2+q(t)}$ and $\sqrt{\rho_j^2+q(t)}$, respectively, where $q(t)=2\epsilon^2(1-e^{-t})$ and $\rho_j^2=|X-c_j|^2+10^{-12}$. Let $V_0$ be the initial phase velocity in Eq.~\eqref{eq:ac-phase-jet}, applied to $H_0=G(X,0)$, and $D(X)=1+\epsilon^2\sum_j\rho_j^{-2}$. With the same outer window $W$ as the curvature construction in Supplementary Section~\ref{supp:ac-mode-coupling}, the neural field was
\[
\begin{aligned}
H_N(X,t)&=G(X,t)+te^{-tD(X)}\bigl(V_0(X)-G_t(X,0)\bigr)
             +\left(\frac{t}{8}\right)^2N(X,t/4-1),\\
u_N(X,t)&=-1+W(x)W(y)\bigl[1+\tanh H_N(X,t)\bigr].
\end{aligned}
\]
The network therefore received initial-data, initial-velocity, radial-smoothing and boundary priors. It did not receive the polar shape law, mode interactions, fitted geometric parameters or learned neck displacement. The two representations retained their own internal core regularizations.

All fits used float64 arithmetic, one CPU thread and a 1,200-second total allowance including setup and saving. Their loss was the curvature construction's loss described in Section~\ref{methods:ac-protocol}: PDE, initial-value, initial-secant and periodic-boundary MSEs, plus a focused PDE term of weight $0.35$, all multiplied by $10^3$. Each repeat used 4,096 uniform residual points, copies with both spatial coordinates multiplied by $0.55$, 512 initial points and 64 paired boundary samples per direction. The initial secant used $h=0.008$. All batches remained fixed during fitting. Training ended at 1,185 seconds to reserve time for saving; guards also capped each fit at 100,000 updates or 20,000 loss-and-gradient evaluations.

L-BFGS used learning rate $0.5$, strong-Wolfe line search, one internal iteration, \texttt{max\_eval}=25, history size 20 and gradient/change tolerances $10^{-9}$ and $10^{-12}$. The lowest training loss among the initial field and completed optimizer steps selected the returned state. Interrupted steps were discarded, and numerical failures were retained without replacement runs. Numerical reference fields did not enter fitting or selection.

Evaluation used the full $384^2$ periodic grid and all 262 saved times, with the space--time norm in Section~\ref{methods:ac-protocol}. The reference used the task's $10^{-12}$ squared-radius regularization. We also evaluated the reference-defined transition band $|u_{\rm ref}|\le0.9$, using the same quadrature for its error and reference energies. Interface discrepancies were measured by the bidirectional RMS and maximum nearest-point distances between linearly interpolated zero crossings of grid edges, with periodic distances.

Supplementary Table~\ref{tab:ac-current-neural-controls} reports all three fits, and Supplementary Fig.~\ref{fig:ac-neural-controls} compares their field and interface errors with the fourteen-parameter ASYS construction.

\FloatBarrier
\subsection{Additional profile and inference studies}
\label{methods:additional}

\subsubsection{Conditional waiting-time profile}
\label{methods:waiting}

For the two-dimensional porous-medium equation $\rho_t=\Delta_r(\rho^2)$, pressure $p=2\rho$ satisfies $p_t=p(p_{rr}+p_r/r)+p_r^2$. Near a stationary positive inner front, a regular single-scale waiting reduction gives a planar profile equation~\citep{lacey1982waiting,aronson1983waiting}. With nonzero cubic normalization fixed to one, the supplied problem was
\begin{equation}
 FF''+(F')^2-\frac73xF'+\frac{11}3F=0,\quad 0\leq x\leq 0.01,
 \qquad F(0)=F'(0)=0,\quad F''(0)=\frac13,\quad F'''(0)=6.
\label{eq:supp_profile_ode}
\end{equation}
The input included $P_7=x^2/6+x^3+(45/2)x^4+945x^5+67095x^6+9542610x^7$, the order-eight resonance, and its leading term $Kx^8\log x$, $K=-1241866080$. The constructive task was to extend this local information into an explicit approximation on the interval.

The eighteen coefficients of the returned expression (Supplementary Section~\ref{supp:waiting-profile}) were fitted by equation-only least squares across 401 uniform and 101 logarithmic points during the worker's 54 local evaluations, and formal evaluation retained that state without further updates. A separate assessment used 257 logarithmic points in $[10^{-5},0.01]$ and normalized the ODE defect by the sum of the absolute values of its four terms. We report the RMS and maximum of this normalized defect; these sampled equation checks do not estimate profile error.

\subsubsection{Stationary Fujita profile}
\label{methods:fujita-stationary}

The radial cubic-reaction equation $u_t=u_{rr}+4u_r/r+u^3$ admits the similarity reduction $u(r,t)=(T-t)^{-1/2}\Phi(\eta)$, with $\eta=r/\sqrt{T-t}$ and singularity time $T$. We sought a positive, nonconstant profile, stationary in rescaled coordinates, with unknown origin amplitude:
\begin{equation}
 \begin{gathered}
 \Phi''+\frac4\eta\Phi'+\Phi^3-\frac12\Phi-\frac12\eta\Phi'=0,\qquad \Phi'(0)=0,\\
 5\Phi''(0)+\Phi(0)^3-\frac12\Phi(0)=0,\qquad 350\Phi'(350)+\Phi(350)=0.
 \end{gathered}
\label{eq:fujita-profile}
\end{equation}
The final condition represents the leading algebraic tail at a finite endpoint and enters the fitting loss as a penalty. The origin amplitude is fitted together with the shape; the origin regularity relation then determines its curvature.

The construction in Supplementary Section~\ref{supp:fujita-stationary} has 38 fitted scalars: an amplitude parameter and 37 Chebyshev coefficients. Initialization used $A=5$ and zero polynomial coefficients, without fitted profile samples. ASYS derived differentiated Chebyshev recurrences and a factored ODE residual for coefficient fitting. Strong-Wolfe L-BFGS took 150 outer steps in 6.919 seconds under a 300-second allowance. With $R$ denoting the ODE residual, the loss was
\[
10^{10}\left\{\operatorname{mean}[(1+\eta^2)R^2]
+\operatorname{mean}[(\eta\Phi'+\Phi)^2]_{\eta=350}\right\},
\]
using 512 training points in each of four equally weighted strata: the core, logarithmic tail, near-origin region and graded outer region. Scrambled Sobol coordinates $s\in[0,1]$ were mapped to
\[
12s,\qquad 12(350/12)^s,\qquad s^2,\qquad
350-\frac{\sinh[s\operatorname{arsinh}(350^2)]}{350},
\]
respectively, with radii clipped below at $10^{-6}$. Assessment differentiated the frozen field independently.

The reference branch was obtained from the collocation seed $2/\sqrt{1+\eta^2}$ by increasing the outer radius from $20$ to $700$, with an algebraic expansion through $\eta^{-9}$ at the outer boundary. Independent inward/outward shooting, matched at $\eta=3$, checked the branch. Relative profile $L^2$ error used 14,001 nodes in $[0,350]$ with trapezoidal $d\eta$ weights. The interior ODE residual was also evaluated on 11,001 independent automatic-differentiation points, with the endpoint defect reported separately. The reference retained higher tail terms rather than enforcing the zero leading-tail condition at $350$.

\subsubsection{Keller--Segel density construction}
\label{methods:ks}

The radial Keller--Segel task sought an explicit density $u(r,t)$ and its enclosed mass $M(r,t)$ satisfying
\begin{equation}
u_t=\frac{(r u_r+uM)_r}{r},\qquad
M(r,t)=\int_0^r s u(s,t)\,ds,\qquad
u(r,0)=9.5e^{-r^2},
\label{eq:ks-ivp}
\end{equation}
on $0\leq r\leq30$ and $0\leq t\leq T=0.86652$, with origin symmetry and remote decay. The total radial mass is $4.75$, using the convention without $2\pi$. Earlier ASYS searches developed a positive Gaussian-mixture representation with analytic enclosed mass. The reported continuation inherited those programs and fitted states, and allowed changes to the representation, loss and optimizer. The two revisions compared here began from the same fitted mixture of thirteen Gaussians with time-dependent weights and widths. The shape extension in Supplementary Section~\ref{supp:overview-ks} retained its temporal representation and initialized thirteen additional shape coordinates at zero, recovering that Gaussian field.

Fitting used four-point Gauss--Legendre quadrature in each of 144 radial cells, giving 576 radial nodes. The cell faces were zero followed by 144 logarithmically spaced values from $10^{-5}$ to $30$; quadrature weights included the radial measure $r\,dr$ and were normalized by the total radial area. At each of 64 times $t_j=T[1-\cos(\pi j/63)]/2$, we computed the radial mean squared PDE residual, then averaged these values with equal time weights. The loss also included the MSEs of $u(30,t)$ and $u_r(30,t)$ at 128 equally spaced times, and a $10^{-10}$ mean-square penalty on the temporal coefficients. Initial constraints and origin symmetry were built into the expression.

We compare the fitted Gaussian mixture with two ASYS revisions. For residual reweighting, let $E_j$ be the radial mean of $u_t^2+D^2+C^2$ at $t_j$, where $D$ and $C$ are the diffusion and attraction terms, evaluated once from the inherited field. With $\overline E=\frac1{64}\sum_j E_j$, the fixed time weights were
\[
\widetilde W_j=\operatorname{clip}\!\left[\sqrt{\frac{\overline E}{E_j}},1,32\right],\qquad
W_j=\frac{\widetilde W_j}{\frac1{64}\sum_k\widetilde W_k}.
\]
These weights multiply the radial residual MSE before the time average. The $L_4$ revision instead retained equal time weights and changed the component shapes, adding a penalty $10^{-10}\operatorname{mean}(\theta_j^2)$. Both revisions retained damped Gauss--Newton and the 300-second fitting allowance; shape-coordinate steps were capped at $0.01$. A further control continued fitting the unchanged Gaussian mixture under the same allowance.

The fixed evaluator assessed the continuous PDE independently of the fitting loss. It used 1,280 radial cells with a face at the origin and geometrically spaced positive faces from $10^{-5}$ to $30$. At each cell center, automatic differentiation supplied $u_t$, $u_r$ and $u_{rr}$; four-point Gauss--Legendre quadrature over full and partial cells supplied the enclosed mass. Squared equation residuals were weighted by each cell's radial area $(r_{i+1/2}^2-r_{i-1/2}^2)/2$ and divided by the total radial area. The equation loss was the arithmetic mean over 36 times: 21 equally spaced points in $[0,T/2]$, followed by 15 points whose successive time intervals decrease geometrically with ratio $0.89257$ and together span $[T/2,T]$.

Initial-value and initial-tangent losses were unweighted MSEs on a separate 256-cell radial grid with the same geometric-face rule. The tangent target was the PDE right-hand side evaluated on the prescribed initial density. At 256 equally spaced times, the boundary loss summed the MSEs of $u_r(0,t)$, $u(30,t)$ and $u_r(30,t)$. The four losses were negated and ranked separately during search.

Density accuracy was measured by
\begin{equation}
\mathcal E_{\rm KS}=
\left[
\frac{\int_0^T\int_0^{30}(u-u_{\rm ref})^2r\,dr\,dt}
{\int_0^T\int_0^{30}u_{\rm ref}^2r\,dr\,dt}
\right]^{1/2},
\label{eq:ks-field-error}
\end{equation}
with trapezoidal quadrature over 77 times and 48,001 radial nodes. The reference solved $M_t=M_{rr}+(M-1)M_r/r$ on 192,000 cells using centered second-order differences and BDF, with relative/absolute tolerances $2\times10^{-9}/2\times10^{-11}$. The illustrated $L_4$ construction was selected for its lowest density error in the subsequent reference assessment; the fixed equation-based evaluator had instead ranked a higher-degree $L_4/L_8$ variant first.

\subsubsection{Graveleau pressure-profile task}
\label{methods:graveleau}

For radial porous-medium flow, a Graveleau similarity profile describes a shrinking hole in the fluid-filled region. Let $\eta\geq1$ be radius divided by the hole radius, $\beta$ its contraction exponent and $p=2-1/\beta\in(0,1)$. The normalized pressure profile obeys
\begin{equation}
 \Psi(\Psi''+\Psi'/\eta)+(\Psi')^2-\eta\Psi'+p\Psi=0,
 \qquad \Psi(1)=0,\quad\Psi'(1)=1.
\label{eq:graveleau-profile}
\end{equation}
The task examined whether the approximation family could represent the far-field balance required by this equation. The coordinate revision in Supplementary Section~\ref{supp:graveleau-tail-repair} retained the existing nine-coefficient shape polynomial within a 24-scalar pressure construction.

\subsubsection{Hamilton--Jacobi coefficient inference}
\label{methods:hj}

The inverse task fitted a field and four independent coefficients in
\begin{equation}
 u_t=d_xu_{xx}+d_yu_{yy}+b_xu_x^2+b_yu_y^2,
 \qquad (x,y)\in[0,2\pi)^2,
\label{eq:hj-inverse}
\end{equation}
with periodic boundaries. The prescribed initial field was the fixed Fourier sum
\begin{equation}
u_0(x,y)=\frac{1}{s_0}\sum_{0<|k|\leq5}
\frac{a_k\sin(k_xx+k_yy+\varphi_k)}{|k_x|+|k_y|+1},
\qquad s_0=3.2939726507728646,
\label{eq:hj-initial}
\end{equation}
over 80 integer wavevectors, with amplitudes $a_k$ and phases $\varphi_k$ supplied as fixed task data. The reference coefficients were $(d_x,d_y,b_x,b_y)=(0.115,0.085,-0.43,-0.61)$. Fitting treated them independently, with diffusion bounds $(0.01,0.3)$, nonlinear bounds $(-2,1)$ and starting vector $(0.06,0.20,-0.3,-0.7)$.

The observations sampled 205 points per frame from a $64\times64$ spatial grid at times $t_n=n/98$. They comprised 5,945 training values at frames $1$--$29$ and 1,025 validation values at frames $30$--$34$, with Gaussian noise at 3\% of the clean training-field RMS. Fitting used the training observations and equation constraints through $H=34/98$; validation observations were used for checkpoint selection. Clean fields and coefficient truth were withheld from the solver agents.

The fitting loss summed four equally weighted components: normalized equation, initial-value, periodic-boundary and training-observation MSEs. The equation residual was scaled by $H/s_u$, value mismatches by $1/s_u$ and periodic normal-derivative mismatches by $2\pi/s_u$, with $s_u=0.2729217507132037$. The boundary component averaged the four value and derivative MSEs across the two periodic seams. Each update drew fresh batches of 256 uniform space--time points, 128 uniform initial points, 64 uniform boundary points reused for both coordinate directions, and 256 training observations with replacement.

Both time-basis fits used 20,000 Adam updates, with the learning rate following a cosine decay from $0.003$ to $0.00006$. They used float64 arithmetic, one CPU thread and model seed 1101, under a 6,600-second time allowance. The training and validation sampling seeds were 1102 and 101102, respectively.

The evaluation score was the negative sum of the same four components, with validation observations replacing training observations. It was evaluated on four fixed validation batches and all validation observations at step zero, every 500 updates and the final step. The earliest checkpoint attaining the highest score was retained.

Holding the 1,257-function spatial Fourier basis, 7,542 field amplitudes and fitting settings fixed, the comparison replaced polynomial functions of physical time with the exponential time coordinate in Supplementary Section~\ref{supp:hj-inverse}. Future-field error evaluated each frozen expression directly at the fifteen later times $t=35/98,\ldots,49/98$. Relative $L^2$ norms equally weighted the $64\times64$ spatial grid and the assessed times. Coefficient error was the mean of the four relative absolute errors. The comparison used one observation-noise realization and one fitting seed.

\subsubsection{NLS periodic-wave study}
\label{methods:nls}

This subsection specifies the supplementary NLS study (Supplementary Section~\ref{sec:nls}). Its equation is $i\psi_t+\tfrac12\psi_{xx}+|\psi|^2\psi=0$ on $x\in[-L,L)$, $t\in[0,\pi/2]$, with $\psi(x,0)=2\operatorname{sech}x$ and periodic evolution in weak form.

\paragraph{Breather and boundary radiation.}\label{supp:nls}
On the periodic domain $[-L, L)$ over $t \in [0, T_{\max}]$ with $T_{\max} = \pi/2$, the focusing cubic nonlinear Schr\"odinger (NLS) equation reads
\begin{equation}
i \partial_t \psi + \frac{1}{2} \partial_{xx} \psi + |\psi|^2 \psi = 0.
\end{equation}
On the unbounded line $\mathbb{R}$, exact multisoliton solutions to the focusing cubic NLS equation are given by the inverse scattering transform~\citep{zakharov1972soliton,sulem1999nls}. In particular, the initial condition $\psi(x, 0) = 2 \operatorname{sech} x$ evolves as the second-order Satsuma--Yajima breather \citep{satsuma1974},
\begin{equation}
\psi_b(x, t) = \frac{4 e^{it/2} \bigl(\cosh 3x + 3 e^{4it} \cosh x\bigr)}{\cosh 4x + 4 \cosh 2x + 3 \cos 4t}.
\end{equation}
The task supplied this breather, its spatial derivative and unfitted finite-box correction templates. Restricting $\psi_b$ to $[-L, L]$ introduces first-derivative discontinuities at the periodic boundary $x = \pm L$. To enforce the initial datum exactly on $[-L, L]$, the localized core superimposes breather dynamics onto the static profile:
\begin{equation}
C(x, t) = 2 \operatorname{sech} x + \psi_b(x, t) - \psi_b(x, 0),
\end{equation}
satisfying $C(x, 0) = 2 \operatorname{sech} x$ identically.

The supplied guidance specified the zero-initial-value radiation equation $iB_n'-\omega_nB_n=(-1)^n\partial_x C(L,t)/(2L)$. ASYS derived the boundary source's temporal Fourier coefficients and constructed the finite radiation field $B(x,t)$. Resolved over $65$ spatial cosine modes ($n = 0, \dots, 64$) and 8 temporal boundary harmonics ($m = -3, \dots, 4$) with frequencies $\nu_m = 1/2 + 4m$, this field is
\begin{equation}
B(x, t) = \sum_{n=0}^{64} \mu_n \cos\left(\frac{n\pi x}{L}\right) \frac{(-1)^n}{2L} \sum_{m=-3}^{4} a_m(L) H_{\omega_n, \nu_m}(t),
\end{equation}
where $\mu_0 = 1$, $\mu_{n>0} = 2$, and dispersion frequencies are $\omega_n = \frac{1}{2} (n\pi / L)^2$. The temporal Fourier coefficients $a_m(L)$ of the boundary derivative $\partial_x \psi_b(L, t)$ are $a_m(L)=T_{\max}^{-1}\int_0^{T_{\max}}\partial_x\psi_b(L,t)e^{-i\nu_m t}\,dt$, evaluated analytically in the program. The propagator $H_{\omega, \nu}(t)$ solves $i H' - \omega H = e^{i\nu t}$ with $H(0) = 0$:
\begin{equation}
H_{\omega, \nu}(t) = -i t e^{i(\nu - \omega)t/2} \frac{\sin((\omega + \nu)t/2)}{(\omega + \nu)t/2}.
\end{equation}
Every finite cosine correction has zero derivative at $x=\pm L$. The returned field therefore retains the breather's derivative jump, $\widehat\psi_x(L,t)-\widehat\psi_x(-L,t)=2\partial_x\psi_b(L,t)$. We interpret the periodic evolution in weak form; the Fourier-moment residual in Section~\ref{methods:nls-diagnostics} includes this boundary seam.

The nonlinear difference $N(C+B)-N(C)$, with $N(z)=|z|^2z$, supplies the forcing for the interaction corrections developed in Supplementary Section~\ref{sec:nls}.

\paragraph{Search and evaluation.}
The NLS searches used two workers per iteration; the prescribed control used one. The NLS continuation sampled up to four programs and four guidance entries, with a 120-minute invocation limit.

The first NLS search produced eight revised guides. Its width-transfer continuation retained the two imported guides and assessed all four newly returned programs. The predictor $P$ and defect correction $E$ formed an eight-coordinate fitted family. We subsequently requested evaluations of $C+B$, $C+B+P$ and $C+B+P+E$ with prescribed amplitudes, retaining the same response bank and numerical settings. These evaluations supply the reported NLS results.

NLS permits changes to the representation, training loss and optimizer. Its 150-second fitting envelope reserves 30 seconds for finalization, leaving a 120-second optimization target. Search candidates could fit coordinates shared across widths; the reported $P+E$ field fixes those coordinates without optimization. The eight-coordinate program evaluates its loss using its own Gauss--Legendre quadrature. NLS computes equation, initial-value, initial-tangent and periodic-value components for each fitting width and both evaluation offsets. Its integrated weak residual uses 256 spatial points, 41 time intervals, 33 Fourier modes and eight-point temporal quadrature. Each component uses its worst value across widths and offsets; the scalar score averages the four signed components. The initial-time check compares weak modal secants with the equation's modal derivative. Program sampling gives unit weight and temperature to the physics, initial-value, initial-tangent and boundary components; initial-value errors within one machine epsilon in relative RMS are treated as tied, and raw errors are retained. The evaluation pools squared errors and reference energies across six domain widths.

\paragraph{Response construction and assessment.}\label{methods:nls-construction}

The NLS field uses the prescribed predictor $P$ and defect correction $E$ defined in Supplementary Section~\ref{supp:nls-formula}. The reported field uses fixed correction amplitudes. To evaluate their response integrals, we project each source onto cosine modes $0,\ldots,24$ using 256 midpoint spatial nodes. A degree-320 Chebyshev interpolant supplies the source values at 320 Gauss--Legendre nodes within each time integral. The quadrature then gives each modal response as a fixed weighted sum of source values. The radiation field retains modes $0,\ldots,64$.

We store the responses after dividing by $u^2$, where $u=t/T_{\max}$ and $T_{\max}=\pi/2$. Multiplication by $u^2$ on evaluation preserves their zero initial value and time derivative. Analytic limits supply the scaled values at $u=0$; these limits are derived in Supplementary Section~\ref{supp:nls-formula}. Each response direction is stored as 321 Chebyshev coefficients for each of its 25 spatial modes at each domain width.

The stored representation contains eight directions: two chains of four, starting from $P$ and $E$ and applying the linearized response about $C+B+P$ to advance each chain. We orthonormalize these directions by modified Gram--Schmidt: for each direction, we subtract its components along the preceding directions twice, then normalize. The real inner product acts on the physical responses, after restoring the $u^2$ factor, and averages over $L\in\{4,5,6\}$. Within each width, Clenshaw--Curtis weights integrate over time and cosine multiplicities $\mu_0=1$, $\mu_{n>0}=2$ weight the spatial modes. The contribution from width $L$ is scaled by $(Le^L)^2$.

For each orthonormal response $v_j$, we form the linearized equation residual about $\psi_0=C+B+P$. Its normalized cosine coefficients are
\[
\mathcal R_{Ln}[v_j]=Le^L\left[i\partial_t(v_j)_n-\frac12\left(\frac{n\pi}{L}\right)^2(v_j)_n
+\Pi_n\!\left(2|\psi_0|^2v_j+\psi_0^2\overline{v_j}\right)\right].
\]
Here $\Pi_n f$ is the spatial mean of $f\cos(n\pi x/L)$ on the same 256 midpoint nodes; the time derivative differentiates $u^2$ times the stored polynomial. The residual Gram matrix uses 384 Gauss--Legendre nodes on $[0,T_{\max}]$, with weights $q_i$ normalized to sum to one:
\[
G_{jk}=\operatorname{Re}\sum_{L\in\{4,5,6\}}\sum_{i=1}^{384}\sum_{n=0}^{24}
\frac{q_i\mu_n}{3\sum_{m=0}^{24}\mu_m}\,
\overline{\mathcal R_{Ln}[v_j](t_i)}\mathcal R_{Ln}[v_k](t_i).
\]
We add $10^{-10}\operatorname{tr}(G)/8$ to each diagonal entry and use the inverse transpose of its Cholesky factor as the fixed coordinate transformation $M$. The field evaluation in Supplementary Eq.~\eqref{supp:eq:nls_modal_eval} combines this map with the stored orthonormal directions. Coordinates are fixed without optimization to represent $P+E$ in the reported field and $P$ in the predictor comparison; both corrections have unit physical amplitude, though their coordinates in the conditioned bank need not equal one.

\paragraph{Field, mass and weak-residual diagnostics.}\label{methods:nls-diagnostics}

We measure whole-field discrepancy by $\mathcal E=\|\psi-\psi_{\rm ref}\|/\|\psi_{\rm ref}\|$ on a common readout grid, using uniform spatial means and trapezoidal time integration. Pooling these squared errors and reference energies directly across $L\in\{3.5,4,4.5,5,6,7\}$ gives the pooled NLS percentage in Supplementary Section~\ref{sec:nls}. This pooled comparison and the $C+B$ versus $C+B+P$ comparison use $1024\times161$ space--time points at each width. The finer predictor and defect comparison uses $4096\times321$ points for $L\le4.5$ and $2048\times321$ for wider domains.

The sampled relative mass drift is
\begin{equation}
\Delta M=\max_t\frac{|M(t)-M(0)|}{M(0)},
\qquad M(t)=2L\operatorname{mean}_x|\psi(x,t)|^2.
\end{equation}
The maximum is taken over the saved times. To assess quadrature sensitivity, we compared 2,048 and 4,096 spatial points at the same 321 times for $L=3.5$, and used 4,096 spatial points with nested uniform grids of 321, 641 and 1,281 times for the two narrowest boxes. Neither correction uses a mass-conservation penalty.

The fixed search diagnostic uses 33 Fourier modes, 41 staggered time intervals, 256 spatial points and eight-point Gauss--Legendre quadrature within each interval:
\begin{equation}
R_{nj}=\frac{i[m_n(b_j)-m_n(a_j)]}{h}
+\frac1h\int_{a_j}^{b_j}\left[-\tfrac12k_n^2m_n(t)+c_n(t)\right]dt,
\qquad J_\delta=\frac1{33\cdot41}\sum_{n,j}|R_{nj}|^2.
\label{eq:nls-weak-diagnostic}
\end{equation}
Here $k_n=n\pi/L$, and $m_n$ and $c_n$ are the Fourier moments of $\psi$ and $|\psi|^2\psi$. With $h=T_{\max}/42$, the intervals have endpoints $a_j=(j+1/4+\delta/2)h$ and $b_j=a_j+h$ for $j=0,\ldots,40$. The spatial nodes are $L[2(\ell+\delta)/256-1]$, with $\delta=0$ or $1/2$. The term $-k_n^2m_n/2$ represents half the periodic distributional second derivative, including the boundary seam, in each retained mode. This modal equation check is separate from the field-error and mass readouts.

\FloatBarrier
\subsection{Numerical reference solutions}\label{supp:eval}\label{supp:references}

Constructed fields were compared with separately generated numerical references. Reference values and accuracy scores were used for scientific assessment outside the solver agents' fitting objectives and search guidance, apart from the Fujita blow-up time estimate described in Section~\ref{supp:fujita-setup}. Each study used the sampling rule stated below and summarized in Supplementary Table~\ref{tab:supp-references}; the reported norms retain these different physical domains.

\paragraph{gCLM.}
Candidate profiles $\hat{f}$ were benchmarked against a numerical reference $f_{\mathrm{ref}}$ computed at $a = 0.25$ from the fixed-point integral formulation of \citet{huang2024gclm} on 4{,}097 logarithmic nodes over $w = \log x \in [-18, 18]$, using quintic splines converged to $10^{-11}$ ($\beta_{\mathrm{ref}} = 0.6845136212507712$). Profile errors were evaluated on 257 logarithmically spaced radii $y_k \in [10^{-3}, 10^3]$ using the unweighted discrete relative $L^2$ norm:
\begin{equation}
E_{\log} = \left( \frac{\sum_{k=1}^{257} \left| \hat{f}(y_k) - f_{\mathrm{ref}}(y_k) \right|^2}{\sum_{k=1}^{257} \left| f_{\mathrm{ref}}(y_k) \right|^2} \right)^{1/2},
\label{supp:eq:primary-metric}
\end{equation}
alongside three auxiliary metrics: the physical-radius trapezoidal relative $L^2$ error ($E_y$), the peak-normalized maximum error ($E_\infty$), and the similarity exponent error $|\Delta\beta| = |\hat{\beta} - \beta_{\mathrm{ref}}|$.

Across coarse, medium, fine and extended-domain reference banks, discrepancies relative to the fine grid remained below $1.2 \times 10^{-10}\%$, while the selected ASYS profile error stayed near $2.47 \times 10^{-8}\%$.

\paragraph{Transient Fujita.}\label{methods:fujita-reference}
The reference used radial finite volumes on 8,192 uniform shells, with the radial flux weighted by $r^4$ and pointwise cubic reaction. Adaptive BDF integration used relative and absolute tolerances $2\times10^{-10}$ and $2\times10^{-12}$. An even cubic-spline reconstruction supplied values on the union of 2,049 equally spaced radii in $[0,5]$ and 1,501 in $[0,0.15]$, giving 3,549 distinct radii. The 289 output times combine 193 equally spaced times from zero to $0.95T_\ast$ with $t=T_\ast(1-\tau)$ for 97 logarithmically spaced $\tau\in[0.001,0.05]$, where $T_\ast=0.020406$. The initial slice and outer boundary were set to their prescribed values. Trajectory and final-time errors use trapezoidal $dr\,dt$ and $dr$ weights, respectively; they measure radial-profile error rather than the five-dimensional volume norm with weight $r^4$.

A saved refinement doubled the finite-volume mesh to 16,384 shells and tightened both BDF tolerances by a factor of ten. Its trajectory differed from the reference by $6.81\times10^{-5}\%$ in the same relative $L^2(dr\,dt)$ norm. At that finer resolution, an independent nodal finite-difference solve differed from the finite-volume field by $2.26\times10^{-6}\%$ over the trajectory and $2.02\times10^{-5}\%$ at the final time.

\paragraph{NLS.}

The NLS reference was computed by Fourier--Strang splitting on 4,096 spatial nodes with $\Delta t\le\pi/81920$, using Fourier differentiation without filtering or dealiasing, with shortened steps to reach the saved output times. Separately doubling its spatial grid and halving its maximum time step changed the reference by at most $6.93\times10^{-5}\%$ and $9.46\times10^{-6}\%$, respectively, while the prescribed $C+B+P+E$ pooled error remained $0.0250\%$ with its response construction held fixed.

\paragraph{Allen--Cahn: coupled-mode fields.}
The main reference uses Fourier-collocation ETDRK4 on $x_i=y_i=-1+2i/384$, with maximum step $1/512$. The 262 output times and error weights are specified in Section~\ref{methods:ac-protocol}. The method uses 32-point contour evaluation of its exponential coefficients and no explicit dealiasing. A separately saved $512^2$ trajectory with maximum step $1/1024$ has twelve output times, including the four displayed frames; reference comparisons use exactly shared times without temporal interpolation.

Restricting the main readout to the nested $192^2$ spatial grid changes either trajectory error by less than $2.2\times10^{-8}$ percentage points. On the ten exactly shared times and a common $512^2$ grid, replacing the finer reference with quintic or Fourier interpolation of the $384^2$ reference changes the pooled errors by less than $4.2\times10^{-9}$ percentage points.

Pooling all twelve finer-grid times instead gives $0.802189\%$ and $0.914589\%$; these values also change the temporal quadrature and cannot be treated as reference error alone. Equal-snapshot pooling on the full 262-time grid gives $0.790567\%$ and $0.907191\%$. 

A separate Fourier--Strang reference differs from the finer ETDRK4 field by $3.09\times10^{-6}\%$ at $t=8$. These are finite numerical sensitivity checks, not rigorous continuum error bounds.

\paragraph{Conditional waiting-time profile.}
The conditional waiting-time study assessed its reduced equation directly, using the normalized ODE defect specified in Section~\ref{methods:waiting}. The returned formula and its endpoint and expansion checks are given in Supplementary Section~\ref{supp:waiting-profile}.
\FloatBarrier

\FloatBarrier

\subsection{Manuscript preparation}
Large language models assisted with drafting and editing the manuscript. The authors are responsible for its scientific content and final text.

\section*{Data Availability}
The reported constructions, fitted states and comparison measurements are available at \url{https://github.com/scaling-group/ASYS-PDE/tree/main/experiments}. The repository also provides scripts for generating the numerical reference data.

\section*{Code Availability}
The code supporting this study is publicly available at \url{https://github.com/scaling-group/ASYS-PDE}.

\section*{Acknowledgements}
This project is partially supported by the National Research Foundation, Singapore, under the NRF fellowship for Liu Yang (Project No. NRF-NRFF17-2025-0006). This work is partially supported by the NUSRI-SZ research grant SZ2025-03.
The funders had no role in study design, data collection and analysis, interpretation of results or manuscript preparation.
We acknowledge NUS IT's Research Computing group for providing high-performance computing resources for this work.

\section*{Author Contributions}
Z.Y. and J.Z. contributed equally to this work. Z.Y. designed the experimental setup and conducted the initial experiments. J.Z. extended the experimental setup, conducted further experiments and expanded the mathematical analysis. L.Y. supervised the research and served as corresponding author. All authors contributed to manuscript preparation and revision.

\section*{Competing interests}
The authors declare no competing interests.

\bibliographystyle{plainnat}
\begingroup
\small
\setlength{\bibsep}{0.5pt}
\bibliography{references}
\endgroup
\ifdefined\asysarticleonly\else
  \clearpage
  \renewcommand{\asysheaderright}{Supplementary Information}
  \renewcommand{\theHfigure}{S\arabic{figure}}
  \renewcommand{\theHtable}{S\arabic{table}}
  \renewcommand{\theHequation}{S\arabic{equation}}
  \appendix
\setcounter{section}{0}
\setcounter{figure}{0}
\setcounter{table}{0}
\setcounter{equation}{0}
\renewcommand{\thefigure}{S\arabic{figure}}
\renewcommand{\thetable}{S\arabic{table}}
\renewcommand{\theequation}{S\arabic{equation}}

\renewcommand{\thesection}{}
\section{Supplementary Information}\label{supp:info}
\setcounter{subsection}{0}
\renewcommand{\thesubsection}{S.\arabic{subsection}}
\begin{table}[htbp]
\centering
\caption{\textbf{Mathematical constructions and their outcomes.} The overview includes the main studies and additional profile, evolution and inference results. Table~\ref{tab:accuracy-comparison} gives the main accuracy and parameter-count comparisons.}
\label{tab:construction-overview}
\begingroup
\small
\setlength{\tabcolsep}{4pt}
\renewcommand{\arraystretch}{1.12}
\begin{tabularx}{\linewidth}{@{}>{\raggedright\arraybackslash}p{.21\linewidth}>{\raggedright\arraybackslash}p{.32\linewidth}Y@{}}
\toprule
Problem and task & Constructed representation & Demonstrated outcome \\
\midrule
gCLM\newline Similarity profile\newline (Section~\ref{sec:gclm}) & Fractional tail coordinate and conditioned profile basis. & The tail gives the limiting field and its approach rate; the rate agrees with the equation's balance within $0.0025\%$. \\
\addlinespace
Fujita\newline Initial-value evolution\newline (Section~\ref{sec:fujita-transient}) & Heat/reaction factorization with a spatially dependent logarithmic response; 14 fitted scalars. & At fixed parameter count and fitting protocol, relative field $L^2$ error falls from $3.99\%$ to $0.0879\%$ over the trajectory, and from $25.43\%$ to $0.702\%$ at the final time. \\
\addlinespace
NLS\newline Periodic wave\newline (Section~\ref{sec:nls}) & Nonlinear interaction predictor and a response to its remaining equation defect. & With prescribed coefficients, the defect response reduces mass drift from $2.52$ to $0.0777$ ppm at $L=3.5$. \\
\addlinespace
Allen--Cahn\newline Moving interfaces\newline (Section~\ref{sec:allen_cahn}) & Analytically integrated coupling between interface harmonics, embedded in a diffuse field. & Initial elongation generates a fourth harmonic with a derived coefficient; the fourteen-parameter field has $0.798\%$ trajectory error. \\
\addlinespace
Keller--Segel\newline Concentrating density\newline (Section~\ref{supp:overview-ks}) & Positive Gaussian--Laguerre mixture with fourth-degree ($L_4$) shape corrections. & Shape corrections retain exact total mass, the prescribed second moment and analytic enclosed mass; density $L^2$ error is $0.00579\%$. \\
\addlinespace
Porous medium\newline Conditional waiting profile\newline (Section~\ref{supp:waiting-profile}) & Positive logarithmic expansion continued through order 13, with a Chebyshev remainder. & Preserves the resonant logarithm; sampled normalized ODE defect is $6.60\times10^{-17}$. \\
\addlinespace
Fujita\newline Stationary profile\newline (Section~\ref{supp:fujita-stationary}) & Positive compactified profile connecting a constrained regular core to an algebraic tail. & Explicit origin amplitude, curvature and tail amplitude; profile $L^2$ error is $2.70\times10^{-9}\%$. \\
\addlinespace
Graveleau\newline Pressure-profile tail\newline (Section~\ref{supp:graveleau-tail-repair}) & Fractional coordinate chosen from the similarity equation's far-field balance. & Removes the constant tail offset that coefficient fitting cannot eliminate and admits the required fractional correction. \\
\addlinespace
Hamilton--Jacobi\newline Coefficient inference\newline (Section~\ref{supp:hj-inverse}) & Heat evolution plus Fourier modes with exponential time functions. & At fixed field size, the time-basis change reduces future-field $L^2$ error from $3.36\%$ to $1.06\%$. \\
\bottomrule
\end{tabularx}
\par\smallskip
\begin{minipage}{\linewidth}\footnotesize\raggedright
All $L^2$ errors are relative and reported in percent. Each row states the assessed quantity; the Graveleau entry concerns the derived mathematical structure, and the stationary Fujita profile agrees with the reference at its numerical resolution. Task definitions, formulas and measurement domains are given in the linked sections.
\end{minipage}
\endgroup
\end{table}

\FloatBarrier
\begin{table}[htbp]
\centering\small
\caption{Numerical references and sampling for the selected constructions. Methods Section~\ref{supp:references} describes each reference and its refinement checks.}
\label{tab:supp-references}
\begin{tabularx}{\linewidth}{@{}p{.18\linewidth}p{.30\linewidth}X@{}}
\toprule
Case & Reference & Reported comparison \\
\midrule
gCLM & Fixed-point integral profile on 4,097 logarithmic nodes, tolerance $10^{-11}$ & 257 log-spaced radii in $[10^{-3},10^3]$; equal-weight relative profile $L^2$. Four reference banks checked grid and domain sensitivity. \\
Fujita evolution & Radial finite volume on 8,192 shells; adaptive BDF with relative/absolute tolerances $2\times10^{-10}/2\times10^{-12}$ & Physical $dr\,dt$ and final $dr$ norms on $3549\times289$ radial/time samples. Grid construction and refinement checks are given in Methods Section~\ref{methods:fujita-reference}. \\
NLS & Fourier--Strang splitting on 4,096 spatial nodes; $\Delta t\le\pi/81920$ & The pooled comparison in Section~\ref{sec:nls} combines prescribed-response errors across six widths on $1024\times161$ space--time grids. Refined predictor/defect controls use the grids specified in Methods Section~\ref{methods:nls-diagnostics}. \\
Allen--Cahn & Fourier collocation with ETDRK4 on $384^2$ nodes, maximum step $1/512$ & All 262 saved times over $[0,8]$, with uniform spatial and trapezoidal temporal weights. A $512^2$ reference provides a separate refinement check. \\
Keller--Segel & Radial finite volume with BDF, 192,000 cells; relative/absolute tolerances $2\times10^{-9}/2\times10^{-11}$ & Density recovered from the saved enclosed mass on 48,001 radial nodes and 77 times; radial and temporal quadrature. \\
Stationary Fujita & Independent boundary-value profile & Relative profile norm in the similarity coordinate, with origin and far-field readouts given in Methods Section~\ref{methods:fujita-stationary}. \\
\bottomrule
\end{tabularx}
\end{table}

\FloatBarrier
\Needspace{0.7\textheight}
\subsection{Agent workspace example}
\label{supp:workspace-example}

Each ASYS worker receives a candidate program together with a task description,
a strategy record and mathematical references. The task sets the equation,
constraints and evaluation criteria; the guidance offers useful ideas and
possible directions. Each worker develops the mathematical form and, where permitted, its fitting procedure, while numerical optimization determines the trainable coefficients; it may also revise the guidance to pass reusable findings to subsequent workers.

The Allen--Cahn example below shows how mathematical knowledge gives this search
a direction. Transition-layer theory motivates a $\tanh$ profile, while curvature
motion and the connected neck suggest describing the field through an evolving
interface. The agent translates these ideas into a differentiable field and refines it using feedback from the equation, illustrating the passage from mathematical knowledge to a working approximation that ASYS seeks to automate.
The excerpts retain the scientific instructions and initial guidance for the
experiment in Section~\ref{sec:allen_cahn}; notation is typeset and operational
details are omitted.

\tcbset{
  asysworkspace/.style={
    breakable,
    colback=codebg,
    colframe=codeframe,
    colbacktitle=codeframe,
    coltitle=black,
    fonttitle=\bfseries\ttfamily,
    boxrule=0.4pt,
    titlerule=0pt,
    arc=1pt,
    left=5pt,
    right=5pt,
    top=5pt,
    bottom=5pt,
    toptitle=1pt,
    bottomtitle=1pt,
    before skip=6pt,
    after skip=10pt
  }
}

\begin{tcolorbox}[asysworkspace, title={Task instructions --- Allen--Cahn (excerpts)}]
\small\raggedright
Find a compact, trainable scalar field $u(x,y,t)$ for
\[
 u_t=\epsilon^2(u_{xx}+u_{yy})+u-u^3,\qquad \epsilon=0.04.
\]
The domain is $(x,y)\in[-1,1)^2$, periodic in both directions, with
$0\leq t\leq8$. The initial condition $u_0$ is a smooth union of two diffuse disks:
\begin{itemize}
\item center $(-0.15,0.02)$, radius $0.22$;
\item center $(0.18,-0.03)$, radius $0.18$.
\end{itemize}
The initial positive phase is already connected. This is not two independent
one-dimensional fronts. The useful target is a readable time-dependent
signed-distance geometry for a peanut that relaxes toward a smoother oval or
rounder interface while preserving the periodic square geometry.

\medskip\noindent[\ldots]\par\medskip

\textbf{Shared representation rules.}
Construct explicit mathematical ans\"atze; do not use neural solution surrogates,
embedded numerical PDE solvers, trajectory tables, or coefficients fitted to
numerical solution samples.
\end{tcolorbox}

\begin{tcolorbox}[asysworkspace, title={Supplied mathematical guidance (excerpts)}]
\small\raggedright
Allen--Cahn diffuse interfaces have a natural $\tanh$ profile across an
$O(\epsilon)$ transition layer. In the sharp-interface limit, curved interfaces
move approximately by mean curvature. The overlap neck is therefore a genuine
2D interaction zone: a candidate that treats the two disks as uncoupled shrinking
circles is missing the main geometry.

The evaluator also checks the initial time direction. Matching the value of the
peanut at $t=0$ is not enough; the one-sided tangent must be consistent with
$\epsilon^2\Delta u_0+u_0-u_0^3$. This makes purely geometric interpolation risky
if it has the right shapes but the wrong early velocity.

\medskip\noindent[\ldots]\par\medskip

\textbf{Directions.}
\begin{itemize}
\item A time-dependent blend from the peanut distance field to an oval or rounded
connected interface, with separate controls for neck smoothing, axis lengths,
and center drift.
\item A curvature-aware early-time correction that targets the PDE tangent at
$t=0$, instead of only improving the initial value loss.
\item Periodic coordinate features or boundary factors that make value and
first-derivative matching natural on $[-1,1)^2$.
\item A compact secondary correction only after the geometry is explicit. Large
Fourier patches should not dominate the submitted structure.
\end{itemize}
\end{tcolorbox}

\FloatBarrier
\subsection{Keller--Segel: construction and assessment}
\label{supp:overview-studies}
\label{supp:overview-ks}

\subsubsection{Reshaping a concentrating density}
\label{sec:keller_segel}

The Keller--Segel model describes chemotaxis, in which cells move toward chemical signals and can aggregate when attraction overcomes diffusive spreading~\citep{keller1970aggregation}; refining an approximate density must therefore also account for the enclosed mass that drives its motion. In the radial equation~\eqref{eq:ks-ivp}, the density $u(r,t)$ diffuses while attracting itself through its enclosed mass $M(r,t)$. As the density concentrates, the formula must capture both its narrowing core and the mass enclosed at each radius, which determines the inward attraction. Section~\ref{methods:ks} specifies the initial Gaussian density, domain, fitting and reference assessment.

From the prescribed Gaussian initial density, ASYS developed a mixture of positive Gaussians with time-dependent weights and widths. The mixture captures concentration by transferring mass into narrower components and gives analytic enclosed mass. The refinement below begins with a fitted thirteen-component approximation from these earlier searches.

To let individual components depart from Gaussian shapes, ASYS multiplied each by a polynomial factor; testing candidate directions against the equation residual led the agent to the fourth-degree Laguerre polynomial $L_4$.

Let $w_j(t)>0$ be the mass fractions, with $\sum_{j=1}^{13}w_j=1$, and $a_j(t)>0$ the squared radial scales. The new coefficients $\beta_j(t)$ control the shape of each component:
\begin{equation}
u(r,t)=9.5\sum_{j=1}^{13}\frac{w_j(t)}{a_j(t)}
 \frac{e^{-q_j}[1+\beta_j(t)L_4(q_j)]^2}{1+\beta_j(t)^2},
\qquad q_j=\frac{r^2}{a_j(t)}.
\label{eq:ks-l4}
\end{equation}
Setting every $\beta_j$ to zero recovers the Gaussian mixture. Squaring preserves nonnegativity, while Laguerre orthogonality gives the normalization $1+\beta_j^2$, keeping each component's mass unchanged. The widths were rescaled to satisfy the equation's known second-moment law~\citep{davila2024ks}.

For example, the fitted fourth component had $\beta_4(T)\approx-0.01165$ at the final time $T=0.86652$: at fixed width, its normalized polynomial factor lowers the central density by $2.33\%$ and raises the value at $q=1$ by $1.45\%$, shifting mass from the center toward the shoulders without changing the component's total mass (Fig.~\ref{fig:ks-construction}d).

The polynomial correction also preserves analytic enclosed mass: integrating each kernel gives a finite combination of polynomials and exponentials. The same expression thus supplies the local density and the nonlocal attraction term at any radius and time, without a separate radial quadrature.

ASYS explored two alternative revisions of the Gaussian mixture. Reweighting the fitting residual by the sizes of the time-derivative, diffusion and attraction terms reduced relative density error from $0.041\%$ to $0.0063\%$. The separate shape revision retained the original residual weights while reshaping the existing components with $L_4$, reaching $0.0058\%$ in the same space--time $L^2$ norm with radial weight $r$ (Fig.~\ref{fig:ks-construction}c). The Laguerre construction attained comparable accuracy while making each component's shape adjustable and retaining analytic enclosed mass.

\begin{figure}[htbp]
\centering
\includegraphics[width=\linewidth]{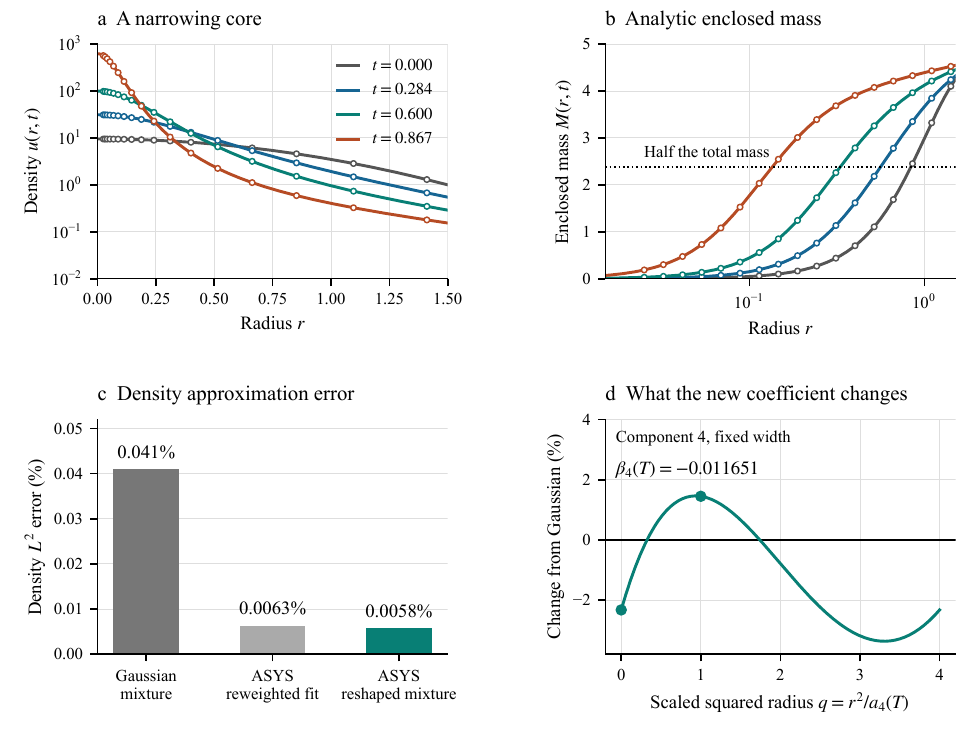}
\caption{\textbf{Concentration in the Gaussian--Laguerre construction.}
\textbf{a,b,}~Density and analytic enclosed mass at four times. Lines show ASYS and circles the numerical reference; colors identify time. The dotted line marks half the radial mass, $M=2.375$.
\textbf{c,}~Relative density error for the Gaussian mixture and two ASYS revisions: reweighting its fitting residual or reshaping its components with $L_4$.
\textbf{d,}~The fitted fourth-component correction moves density from its center toward its shoulders. The curve gives the percentage change from its Gaussian at the same width and final time.}
\label{fig:ks-construction}
\end{figure}

The analytic mass also makes concentration directly measurable: the half-mass radius, defined by $M(r_{1/2},t)=2.375$, contracted from $0.833$ to $0.134$, closely following the numerical reference.

\FloatBarrier
\subsubsection{Formulas and numerical comparisons}

For the density in Eq.~\eqref{eq:ks-l4}, the fourth-degree Laguerre polynomial is
\[
L_4(q)=1-4q+3q^2-\frac23q^3+\frac1{24}q^4.
\]
Using radial mass without $2\pi$, the whole-space identities are
\[
\int_0^\infty r u(r,t)\,dr=4.75,\qquad
\int_0^\infty r^3u(r,t)\,dr=4.75-3.5625t
\]
for the prescribed initial density~\citep{davila2024ks}. A normalized component has mean squared radius $a_j\mu_j$, where
\[
\mu_j(t)=\frac{1+9\beta_j(t)^2}{1+\beta_j(t)^2}.
\]
The second-moment identity therefore requires $\sum_jw_j a_j\mu_j=1-0.75t$. Positive normalized weights and preliminary widths $\widetilde a_j$ are mapped to
\[
a_j(t)=\frac{\widetilde a_j(t)(1-0.75t)}
 {\sum_\ell w_\ell(t)\widetilde a_\ell(t)\mu_\ell(t)}.
\]
The partial-mass integrals are
\[
J_n(q)=\int_0^q e^{-s}s^n\,ds
=n!\left(1-e^{-q}\sum_{k=0}^{n}\frac{q^k}{k!}\right),
\qquad n=0,\ldots,8.
\]
If $[1+\beta_jL_4(q)]^2=\sum_{n=0}^8c_{jn}q^n$, the enclosed mass is
\[
M(r,t)=4.75\sum_j\frac{w_j(t)}{1+\beta_j(t)^2}
 \sum_{n=0}^8c_{jn}(t)J_n(r^2/a_j(t)).
\]
The Gaussian mixture already had analytic enclosed mass; the polynomial-tail integrals extend that capability to the deformed kernels.

The component weights and widths use twenty-four degree-12 Chebyshev curves: two control group masses, twelve control widths and ten control within-group weight logits. Bounded polynomial corrections adjust the inherited curves, while exponentials and normalized weights preserve positivity. The shape revision retained this temporal representation and added thirteen fitted shape coordinates,
\[
\beta_j(t)=(t/T)^3\,0.02\tanh(\theta_j/0.02),\qquad T=0.86652.
\]
The cubic onset preserves the Gaussian precursor's initial density and first two time derivatives; the tanh bounds each deformation. For the fourth component, the fitted value was $\theta_4=-0.0133263$, giving $\beta_4(T)=-0.0116510$.

The field has 325 fitted coefficients and retains 312 previously fitted values from its Gaussian precursor.

The comparison in Fig.~\ref{fig:ks-construction}c uses the fitted Gaussian mixture and two revisions, following the protocol in Section~\ref{methods:ks}. Continuing to fit the unchanged Gaussian mixture left its density error at $0.0409544\%$.

\begin{center}
\small
\begin{tabular}{lr}
\toprule
Construction & Density $L^2$ (\%)\\
\midrule
Gaussian mixture (Fig.~\ref{fig:ks-construction}c) & $0.040954$\\
ASYS reweighted fit & $0.006340$\\
ASYS reshaped mixture ($L_4$) & $0.005789$\\
\bottomrule
\end{tabular}
\end{center}

Refining the numerical reference from 96,000 to 192,000 radial cells changed its final density by $0.00126923\%$ in relative $L^2$, exceeding the $L_4$ field's $0.000562\%$ endpoint discrepancy. Both revised constructions substantially improve on the original Gaussian mixture, but the available reference checks do not resolve their smaller difference in accuracy.

Whereas the Gaussian constructions preserve radial monotonicity, the $L_4$ field has small outward rises in its tail (below $1.8\times10^{-9}$ of its central density), departing from the monotonicity of the regular Gaussian trajectory even though total mass remains exact.

\FloatBarrier
\subsection{NLS: correcting the nonlinear interaction of a wave and its radiation}
\label{sec:forward}
\label{sec:nls}

Whereas the main studies fit trainable coefficients alongside structural revisions, this study uses the governing equation to construct corrections with prescribed coefficients. The focusing nonlinear Schr\"odinger (NLS) equation describes nonlinear wave propagation, including pulses in optical fibers, where dispersion competes with intensity-dependent focusing~\citep{hasegawa1973transmission}. We studied how the boundary changes the wave in the periodic problem
\begin{equation}
i\psi_t+\tfrac12\psi_{xx}+|\psi|^2\psi=0,
\qquad x\in[-L,L),\quad t\in[0,\pi/2],\quad \psi(x,0)=2\operatorname{sech}x.
\label{eq:nls}
\end{equation}
On the real line, this initial wave has a known breather solution: its amplitude repeatedly concentrates and spreads~\citep{satsuma1974}. In a periodic box, boundary-driven radiation interacts with the breathing core. We asked ASYS to construct a finite-box correction to the supplied breather that could be used across box widths.

Using the supplied boundary-source equation, ASYS developed a finite radiation field $B$ around the breather-based core $C$. It then constructed a correction $P$ for their nonlinear interaction and a further correction $E$ for the remaining mismatch with the equation. The resulting field is
\[
\widehat\psi=
\underbrace{C}_{\text{core}}+
\underbrace{B}_{\text{radiation}}+
\underbrace{P}_{\text{interaction predictor}}+
\underbrace{E}_{\text{residual correction}}.
\]
The sum $Z=C+B$ reproduces the initial wave. Its missing interaction comes from the nonlinear term: writing $N(z)=|z|^2z$, adding radiation changes $N(C)$ to $N(Z)$ and leaves the forcing $-[N(Z)-N(C)]$ to be accounted for.

A linear response gives the wave generated by a forcing under the linear Schr\"odinger dynamics, starting from zero, and that correction in turn interacts with the wave to generate further forcing. For a trial correction $d$, let $\mathcal F(d)$ denote the linear response to $-[N(Z+d)-N(C)]$, represented by a finite Fourier sum. A self-consistent correction in this response space would satisfy $d=\mathcal F(d)$.

ASYS constructed the predictor $P=d_1+d_2+d_3+d_4$ through a formal expansion~\citep{nayfeh2000perturbation} in the strength of the core--radiation mismatch forcing. The first term is the correction driven by that mismatch; its further nonlinear interactions generate the higher orders, with repeated linearized feedback included at each order. Evaluating the first four orders at the full forcing strength gives $P$, without fitting additional amplitudes. The hierarchy is thus a perturbative solution of the finite response equation $d=\mathcal F(d)$, evaluated by quadrature. Section~\ref{supp:nls-formula} gives the finite response operators and full recursion; Methods Section~\ref{methods:nls-construction} specifies their numerical evaluation.

Across $L\in\{3.5,4,4.5,5,6,7\}$, adding $P$ reduced relative space--time $L^2$ field error by $67.9$--$95.9\%$, with the largest gain in the narrowest box (Fig.~\ref{fig:nls}).

\Needspace{6\baselineskip}
The predictor's remaining interaction error supplies the next correction. The response demanded by the projected equation differs from $P$ by
\begin{equation}
\Delta=\mathcal F(P)-P.
\label{eq:defect_delta}
\end{equation}
ASYS formed $E$ by summing $\Delta$ and its successive linearized responses around the improved wave $Z+P$~\citep{stetter1978}. Adding $E$ improved conservation of mass, $M(t)=\int_{-L}^L|\psi(x,t)|^2\,dx$, without a mass penalty: at $L=3.5$, the largest sampled departure from the initial mass fell from $2.52$ to $0.0777$ parts per million while field error remained unchanged at the reported precision (Section~\ref{supp:nls-defect-weak}).

To assess what these corrections contribute without amplitude fitting, we evaluated $\widehat\psi$ with its prescribed coefficients. This gave a pooled relative $L^2$ error of $0.0250\%$ across all six widths, compared with $0.530\%$ for $C+B$. The equation supplied both the missing interaction that improved the field and the subsequent correction that improved conservation.

\begin{figure}[htbp]
\centering
\includegraphics[width=\linewidth]{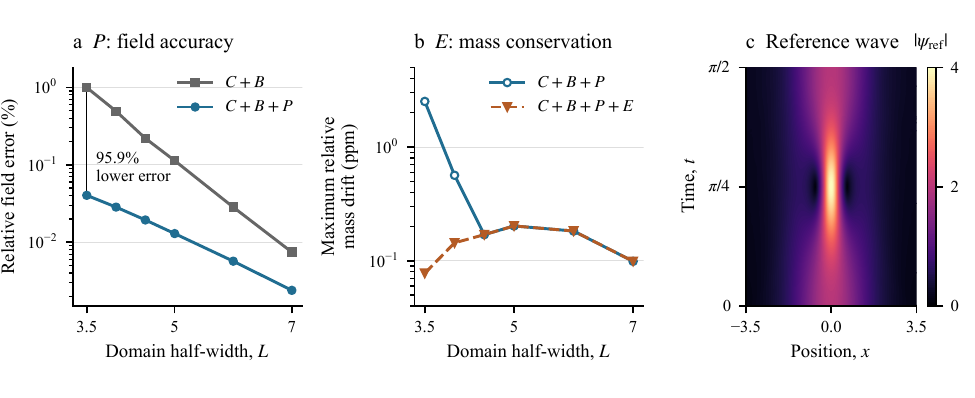}
\caption{\textbf{Constructing the missing nonlinear response.}
\textbf{a,} Adding the ASYS interaction predictor $P$ to the core and radiation $C+B$ reduced field error.
\textbf{b,} The further correction $E$ improved mass conservation in the narrowest boxes. Coefficients were fixed in both comparisons.
\textbf{c,} The numerical reference at $L=3.5$ shows the wave's breathing cycle.}
\label{fig:nls}
\end{figure}

\FloatBarrier
\subsection{Further explicit profile constructions}
\label{supp:profile-extensions}

\subsubsection{A logarithmic completion of a conditional waiting profile}
\label{supp:waiting-profile}
\label{supp:deduction}

In porous-medium flow, the edge of a fluid-filled region can remain stationary before it begins to move~\citep{lacey1982waiting,aronson1983waiting}. Explicit local profiles expose scaling and regularity near this waiting front.

The supplied waiting-profile equation~\eqref{eq:supp_profile_ode} fixes the leading origin derivatives and an order-eight logarithmic resonance; the construction extends this local expansion into a positive explicit approximation on $0\leq x\leq0.01$. Section~\ref{methods:waiting} specifies the supplied expansion, fitting and equation-defect assessment.

ASYS used $u=x/0.01$, $Q=\log(6F/x^2)$ and $D=u\partial_u$. The transformed equation is
\begin{equation}
 e^Q[6+7DQ+D^2Q+2(DQ)^2]-6-14DQ=0.
\label{eq:autonomous_Q}
\end{equation}
A recurrence in powers of $u$ and $\log u$ transports the regular, free and logarithmic coefficients through profile order 13. Let $P_5$ collect the supplied low-order logarithmic expansion, and let $A_5$, $B_5$ and $L_5$ be the derived polynomials multiplying the regular terms, the free coefficient $c$, and $\log u$, respectively. With $T_j$ denoting the degree-$j$ Chebyshev polynomial, the returned expression is
\begin{equation}
 \begin{aligned}
 F(x)&=\frac{x^2}{6}\exp\!\left\{P_5(u)+u^6[A_5(u)+cB_5(u)+L_5(u)\log u]+u^{12}H(u)\right\},\\
 H(u)&=\sum_{j=0}^{16}h_jT_j(2u-1).
 \end{aligned}
\label{eq:full_derived_ansatz}
\end{equation}
The starter log polynomial is
$P_5(u)=0.06u+0.0117u^2+0.004932u^3+0.003639735u^4+0.00543606552u^5$.
Table~\ref{tab:waiting-transport} specifies the three transport polynomials in ascending powers. The eighteen fitted values, $c$ and $h_0,\ldots,h_{16}$, are in Table~\ref{tab:waiting-coefficients}.

The normalized ODE-defect RMS decreased from $1.0654129024\times10^{-4}$ for the supplied polynomial to $6.6016626023\times10^{-17}$; the maximum decreased from $9.2329491756\times10^{-4}$ to $1.7261777980\times10^{-16}$. These are sampled equation defects at cancellation level, not profile-error estimates.

The resulting positive formula preserves the given jets and retains the unsmoothed logarithm at every $x>0$: the leading logarithm leaves the expression seven times continuously differentiable at the origin while its eighth derivative diverges, a distinction written into the formula rather than inferred from finite derivative probes. Exact substitution of the intended rational transport confirms the order-13 cancellation and identifies an order-14 logarithmic-squared term that a smooth $H$ cannot remove. The expression is thus suitable for further asymptotic analysis.

The printed coefficients also give a short monotonicity certificate. Write $P_5=\sum_{j=1}^5p_ju^j$, $d_j=[u^j](A_5+cB_5)$ and $\ell_j=[u^j]L_5$. On $0<u\leq1$, the bounds $|T_j(2u-1)|\leq1$, $|\partial_uT_j(2u-1)|\leq2j^2$ and $nu^n|\log u|\leq e^{-1}$ give
\begin{equation}
\begin{aligned}
 |DQ|\leq{}&\sum_{j=1}^5j|p_j|+\sum_{j=0}^5(6+j)|d_j|\\
 &+(1+e^{-1})\sum_{j=0}^5|\ell_j|+\sum_{j=0}^{16}(12+2j^2)|h_j|<0.454.
\end{aligned}
\label{eq:waiting-monotonicity}
\end{equation}
Thus $F'=(x/6)e^Q(2+DQ)>0$ throughout $0<x\leq0.01$.

\begin{table}[htbp]
\centering\small
\caption{Coefficients of $A_5(u)$, $B_5(u)$ and $L_5(u)$ in the waiting-profile expression, ordered by power $u^j$. These are rounded evaluations of the derived transport coefficients.}
\label{tab:waiting-transport}
\begin{tabular}{rrrr}
\toprule
$j$ & $[u^j]A_5$ & $[u^j]B_5$ & $[u^j]L_5$\\
\midrule
0 & $0$ & $1$ & $-0.00745119648$\\
1 & $-0.010359032135451429$ & $-1.16$ & $0.0086433879168$\\
2 & $0.007684979561788275$ & $0.5511$ & $-0.004106354380128$\\
3 & $-0.002673629947009332$ & $-0.156456$ & $0.00116578439647488$\\
4 & $0.0005525534591464768$ & $0.02235321$ & $-0.0001665581596687008$\\
5 & $-0.000004544849563129317$ & $-0.0175827132$ & $0.00013101225070468952$\\
\bottomrule
\end{tabular}
\end{table}

\begin{table}[htbp]
\centering\small
\caption{Selected waiting-profile coefficients at seventeen significant digits. The scalar $c$ is the free resonant coefficient, distinct from the supplied cubic normalization.}
\label{tab:waiting-coefficients}
\begin{tabular}{lrlr}
\toprule
Coefficient & Value & Coefficient & Value\\
\midrule
$c$ & $-0.021040229650687338$ & $h_{8}$ & $7.3075640917671832\times10^{-8}$\\
$h_{0}$ & $-0.00027280940595691722$ & $h_{9}$ & $-1.0393489618847704\times10^{-8}$\\
$h_{1}$ & $-0.00038804537774768779$ & $h_{10}$ & $-1.1523051840533665\times10^{-10}$\\
$h_{2}$ & $0.00024785634657622011$ & $h_{11}$ & $7.0399464389274407\times10^{-10}$\\
$h_{3}$ & $-0.000082196793794406654$ & $h_{12}$ & $-2.7917938250516926\times10^{-10}$\\
$h_{4}$ & $0.000022095290814842474$ & $h_{13}$ & $6.9273388978046363\times10^{-11}$\\
$h_{5}$ & $-0.0000057718950907469936$ & $h_{14}$ & $-1.1841436370064704\times10^{-11}$\\
$h_{6}$ & $0.0000014901340727612616$ & $h_{15}$ & $1.3055016438080799\times10^{-12}$\\
$h_{7}$ & $-3.5728716964670435\times10^{-7}$ & $h_{16}$ & $-7.1280157812199504\times10^{-14}$\\
\bottomrule
\end{tabular}
\end{table}

\subsubsection{Fujita: a regular core joined to an algebraic tail}
\label{supp:fujita-stationary}

Reaction--diffusion theory studies how local growth competes with spatial spreading, including when solutions become unbounded in finite time~\citep{quittner2019superlinear}. Explicit similarity profiles expose the growing core and far-field decay in one expression.

The stationary similarity profile $\Phi(\eta)$ of the radial cubic-reaction equation must connect its nonlinear core to a slowly decaying far field. Its equation~\eqref{eq:fujita-profile} relates the unknown origin amplitude to the curvature and requires an algebraic tail. Section~\ref{methods:fujita-stationary} specifies the problem, coefficient fitting and independent reference construction.

The selected construction compactifies the half-line and separates the origin curvature from the remaining shape:
\begin{equation}
 \begin{gathered}
 \Phi(\eta)=\frac{A\exp[hz+z^2P_{36}(2z-1)]}{\sqrt{1+k\eta^2}},\qquad z=\frac{k\eta^2}{1+k\eta^2},\\
 k=\frac34,\qquad A^2=\frac12+e^a,\qquad h=\frac12-\frac{e^a}{10k}.
 \end{gathered}
\label{eq:fujita-expression}
\end{equation}
Here $P_{36}(s)=\sum_{j=0}^{36}c_jT_j(s)$. The amplitude parameter $a$ and 37 Chebyshev coefficients give 38 fitted scalars. Every finite parameter choice is positive and even; the factor $z^2$ prevents the polynomial correction from changing the required quadratic origin term. The algebraic prefactor fixes $1/\eta$ decay while the exponential corrects the core and intermediate radii.

Starting from the zero-shape initialization---which uses no fitted profile samples and has $22.8085\%$ initial relative error against the positive-branch reference---coefficient fitting adjusted the interior shape while positivity, the origin relation and algebraic decay held identically across the family.

Against this reference (origin value $6.63471855955$), the selected profile achieved a relative $L^2$ error of $2.6954920925\times10^{-9}\%$ and a maximum absolute error of $2.6568001271\times10^{-11}$; because independent reference methods differed by about $2.52\times10^{-9}\%$ in relative $L^2$, this establishes agreement at the reference's numerical resolution. The frozen source had peak sampled interior ODE residual $4.94\times10^{-12}$; the separate endpoint defect is $350\Phi'(350)+\Phi(350)=-4.35304\times10^{-8}$. Indeed, the ODE tail expansion $\Phi=C/\eta+B/\eta^3+O(\eta^{-5})$ gives $B=2C-C^3$ and $\eta\Phi'+\Phi=-2B/\eta^3+O(\eta^{-5})$, explaining why the leading-tail condition is not exactly satisfied at the finite endpoint.

The coefficients in Table~\ref{tab:fujita-coefficients} expose the profile's main quantities directly:
\begin{equation}
 \begin{gathered}
 \Phi(0)=A=6.63471855955,\qquad \Phi''(0)=-Ae^a/5=-57.7479140847,\\
 \Phi(\eta)\sim C/\eta,\qquad C=\frac{A}{\sqrt k}\exp\!\left(h+\sum_{j=0}^{36}c_j\right)=1.05689553114.
 \end{gathered}
\label{eq:fujita-readouts}
\end{equation}
The approximation therefore supplies both the full stationary shape and analytic core/tail readouts in one expression.

\begin{table}[htbp]
\centering\small
\caption{Selected stationary Fujita coefficients at seventeen significant digits. $k=3/4$ is fixed; $A$ and $h$ follow from $a$ through Eq.~\eqref{eq:fujita-expression}.}
\label{tab:fujita-coefficients}
\begin{tabular}{lrlr}
\toprule
Coefficient & Value & Coefficient & Value\\
\midrule
$a$ & $3.7732088921045812$ & $c_{18}$ & $5.6940882223857279\times10^{-7}$\\
$c_{0}$ & $6.6629233799636163$ & $c_{19}$ & $-2.3756904462103371\times10^{-7}$\\
$c_{1}$ & $-4.5012672833770111$ & $c_{20}$ & $9.5515758444242022\times10^{-8}$\\
$c_{2}$ & $1.5854292806329766$ & $c_{21}$ & $-4.0528179442091812\times10^{-8}$\\
$c_{3}$ & $-0.58321290364586997$ & $c_{22}$ & $1.6042302988790991\times10^{-8}$\\
$c_{4}$ & $0.21821634218094949$ & $c_{23}$ & $-7.0073322052214630\times10^{-9}$\\
$c_{5}$ & $-0.083443483280091463$ & $c_{24}$ & $2.6794362493460203\times10^{-9}$\\
$c_{6}$ & $0.032184090289294372$ & $c_{25}$ & $-1.2351163969522750\times10^{-9}$\\
$c_{7}$ & $-0.012590964745771224$ & $c_{26}$ & $4.3931439029794753\times10^{-10}$\\
$c_{8}$ & $0.0049470421791099652$ & $c_{27}$ & $-2.2416851566144713\times10^{-10}$\\
$c_{9}$ & $-0.0019653559046664535$ & $c_{28}$ & $6.8990115143562463\times10^{-11}$\\
$c_{10}$ & $0.00078179121244574784$ & $c_{29}$ & $-4.2266407176965456\times10^{-11}$\\
$c_{11}$ & $-0.00031414745803181601$ & $c_{30}$ & $1.0013685242988459\times10^{-11}$\\
$c_{12}$ & $0.00012603153178380390$ & $c_{31}$ & $-8.1152237317081682\times10^{-12}$\\
$c_{13}$ & $-0.000051113294099900815$ & $c_{32}$ & $1.3895449403675189\times10^{-12}$\\
$c_{14}$ & $0.000020621788131153491$ & $c_{33}$ & $-1.4066744163981761\times10^{-12}$\\
$c_{15}$ & $-0.0000084336810129699260$ & $c_{34}$ & $2.7738754840176729\times10^{-13}$\\
$c_{16}$ & $0.0000034124802586882915$ & $c_{35}$ & $-1.4177867707104018\times10^{-13}$\\
$c_{17}$ & $-0.0000014079623424171520$ & $c_{36}$ & $7.8187830806573488\times10^{-14}$\\
\bottomrule
\end{tabular}
\end{table}

\FloatBarrier
\subsection{Candidate formulas}
\subsubsection{gCLM: the missing fractional sector}
\label{supp:tail-sector}

The supplied raw envelope is
\begin{equation}
 h_0(x)=x(1+x^2)^{-(1+\alpha)/2}
 \exp\!\left[c\left((1+x^2)^{-1}-1\right)\right],\qquad \alpha=1/\beta.
\label{supp:eq:starter-raw}
\end{equation}
Its large-radius expansion is
\begin{equation}
 h_0(x)=e^{-c}x^{-\alpha}\left[1+
 \left(c-\frac{1+\alpha}{2}\right)x^{-2}+O(x^{-4})\right].
\label{supp:eq:starter-expansion}
\end{equation}
Only the powers $x^{-\alpha-2j}$ occur: for $1<\alpha<2$, the nonlinear profile balance requires an $x^{-2\alpha}$ term ahead of the first available correction, which varying $c$ or the radial scale cannot supply.

\subsubsection{gCLM: tail coefficients and physical rates}
\label{supp:tail-applicability}

Assume an odd integrable profile with a bounded piecewise-smooth core, $f(y)=O(y)$ at zero, and
\begin{equation}
 f(y)=Cy^{-\alpha}+Dy^{-2\alpha}+r(y),\qquad
 r^{(j)}(y)=o(y^{-2\alpha-j}),\quad j=0,1,2,\quad 1<\alpha<2.
\label{supp:eq:f-tail-hyp}
\end{equation}
The odd-profile Hilbert kernel and its zero-constant primitive give
\begin{equation}
 g(y)=\kappa_\alpha Cy^{-\alpha}+o(y^{-\alpha}),\qquad
 V(y)=-\frac{\kappa_\alpha C}{\alpha-1}y^{1-\alpha}+o(y^{1-\alpha}),
 \qquad \kappa_\alpha=\cot(\pi\alpha/2).
\label{supp:eq:g-V-leading}
\end{equation}
The finite core contributes $O(y^{-2})$ to $g$, after its leading fractional term. Substitution in the profile equation first gives $\alpha\beta=1$ and then
\begin{equation}
 \frac{D}{C^2}=K(\alpha):=\lambda\cot\!\left(\frac{\pi\alpha}{2}\right)
 \left[\frac{a\alpha}{\alpha-1}-1\right].
\label{supp:eq:K-def}
\end{equation}
For $\omega(x,t)=-\lambda(T-t)^{-1}f(x(T-t)^{-\beta})$, at fixed $x>0$,
\begin{equation}
 \lim_{t\to T^-}\omega_t=\lambda D x^{-2\alpha},\qquad
 \lim_{t\to T^-}(u_x\omega-a u\omega_x)=\lambda C^2K(\alpha)x^{-2\alpha}.
\label{supp:eq:field-limits}
\end{equation}
Consequently,
\begin{equation}
 \lim_{t\to T^-}(\omega_t+a u\omega_x-u_x\omega)
 =\lambda[D-C^2K(\alpha)]x^{-2\alpha}.
\label{supp:eq:residual-limit}
\end{equation}
The ratio $\rho=D/[C^2K(\alpha)]$ measures agreement of the two limiting rates. This is an analysis of the constructed similarity field, conditional on the stated tail remainders. The selected ASYS program has a smooth core, while the spline-profile program from Run 5 has a compact $C^4$ core; their explicit exterior formulas give $r^{(j)}=O(y^{-\alpha-2-j})$, satisfying the required bounds.

\subsubsection{gCLM: the selected ASYS program}
\label{supp:h3-construction}

Put $x=Ay$, $r=\sqrt{1+x^2}$, $v=r^{-\alpha}$ and $p=2\beta$. Then
\begin{equation}
 h(x)=\frac{x}{r}v\exp[(1-v)^2Q(v)],\qquad f(y)=h(Ay)/A.
\label{supp:eq:h3-def}
\end{equation}
The normalization scale is $A=(2/\pi)\int_0^\infty h(x)/x\,dx$, which enforces $(-Hf)(0)=1$ while $f'(0)=1$ follows from the form of $h$.
At the selected noninteger exponent, the complete correction can be written with collected coefficients as
\begin{equation}
 Q(v)=\sum_{j=0}^7a_j L_j(v)+\sum_{k=0}^3w_k F(q_k,v)
 +[e_{\rm even}+e_{\rm skew}(2v-1)]R(v),
\label{supp:eq:Q-def}
\end{equation}
where $q=(p,p+1,p+2,2p)$, $L_j(v)=\sqrt{2j+1}P_j(2v-1)$ and $R(v)=v^2(1-v)^2/[(v-1/2)^2+1/64]$. The polynomial part uses shifted Legendre polynomials~\citep{dlmf18}. With $\Pi_7$ the $L^2(0,1)$ projection onto degree-seven polynomials,
\begin{equation}
 F(q,v)=\frac{v^q-\Pi_7[v^q](v)}{T(q)},\qquad
 T(q)=\prod_{j=0}^7\frac{q-j}{q+j+1}.
\label{supp:eq:F-proj}
\end{equation}
For direct evaluation,
$\Pi_7[v^q]=\sum_{j=0}^7\frac{2j+1}{q+1}
[\prod_{m=0}^{j-1}(q-m)/(q+m+2)]P_j(2v-1)$.
At an integer $q=j$, the continued value is
$[v^j\log v-\Pi_7(v^j\log v)]/T'(j)$.
This projection removes the part already represented by the polynomial core; the continued limit retains a direction when the powers meet~\citep{borwein1994muntz}.

The coefficients in Table~\ref{tab:gclm-h3-state} collect the mode combinations and coefficient transformation used during fitting (Methods Section~\ref{methods:gclm-coordinates}); they define the selected expression directly. The scale $A$ is determined by normalization.

\begin{table}[htbp]
\centering\small
\caption{Collected coefficients for the selected ASYS expression in Eq.~\eqref{supp:eq:Q-def}. The fitted state has fourteen shape coordinates and one exponent.}
\label{tab:gclm-h3-state}
\begin{tabular}{lr}
\toprule
Coefficient & Value\\
\midrule
$a_0$ & $-0.065562537329050401$ \\
$a_1$ & $0.013303918399269414$ \\
$a_2$ & $-0.0033535579321773985$ \\
$a_3$ & $0.0010663057708611762$ \\
$a_4$ & $-0.00040449848958477674$ \\
$a_5$ & $0.00017693909104070230$ \\
$a_6$ & $-0.000087046299822465832$ \\
$a_7$ & $0.000046993358045409982$ \\
$w_0$ & $-0.000068757035617479612$ \\
$w_1$ & $-0.0000029013082197706332$ \\
$w_2$ & $2.0438417573283607\times10^{-7}$ \\
$w_3$ & $0.0000027133392761763322$ \\
$e_{\mathrm{even}}$ & $4.1367013429493400\times10^{-10}$ \\
$e_{\mathrm{skew}}$ & $-6.5412749367198254\times10^{-9}$ \\
$\beta$ & $0.68451362120671342$ \\
$A$ & $0.76370265908279578$ \\
\bottomrule
\end{tabular}

\end{table}

The factor $(1-v)^2$ is $O(x^4)$ near zero, preserving $h'(0)=1$ and fixing the raw cubic coefficient. At infinity,
$Q(v)(1-v)^2=Q(0)+[Q'(0)-2Q(0)]v+o(v)$, which gives the two tail coefficients in Eq.~\eqref{eq:tail_ratio_analytic}. Both rational modes vanish quadratically at $v=0$ and hence do not directly alter these endpoint values.

\subsubsection{gCLM: the spline-profile exterior (Run 5)}
\label{supp:h5-tail}

This program uses a compact $C^4$ piecewise-quintic core in $\log R$, ending at $41/8$. Outside it, set $x=Ay$, $R=\sqrt{1+x^2}$, $s=R^{-2}$, $U=1-s$, $z=R^{-\alpha}$, $\rho_0=1/8$, $q=z/(\rho_0+z)$ and $w=2(1+\rho_0)q-1$. The expression is
\begin{equation}
 h(x)=\frac{x}{R}R^{-\alpha}\exp\!\left[
 \frac{(\alpha-1)Us}{2}+U^2(b+es)
 +U^3q\left(\sum_{j=0}^3d_jT_j(w)+d_4s\right)\right].
\label{supp:eq:h5-exterior}
\end{equation}
Table~\ref{tab:h5-tail-params} gives the fitted exterior coefficients.
The tail coefficients are $C=e^bA^{-\alpha-1}$ and
$D=C(d_0-d_1+d_2-d_3)A^{-\alpha}/\rho_0$, giving
$C\simeq1.75379333$ and $D\simeq1.13289418$. Its small sampled field error therefore coexisted with a visible deficit in the secondary tail balance.

\begin{table}[htbp]
\centering\small
\caption{Fitted exterior coefficients of the spline-profile program (Run 5).}
\label{tab:h5-tail-params}
\begin{tabular}{lr}
\toprule
Coefficient & Value\\
\midrule
$\beta$ & $0.68451357360607716$\\
$A$ & $0.84712794685052795$\\
$b$ & $0.15351047131762838$\\
$e$ & $0.081330190187120843$\\
$d_0$ & $0.065745660753748367$\\
$d_1$ & $0.0066774659147825627$\\
$d_2$ & $0.0043742139756283143$\\
$d_3$ & $0.000075536825602929490$\\
$d_4$ & $0.0029325761345243059$\\
\bottomrule
\end{tabular}
\end{table}

\subsubsection{gCLM: construction evidence and fitting coordinates}
\label{supp:lineage-progression}

The construction in Table~\ref{tab:gclm-history} developed through changes to both the function family and its coefficient coordinates. Adding mixed-tail terms worsened the initial fractional form; the next construction fixed the redundant raw radius and introduced Jacobi coordinates, though the recorded full shape Gram remained badly conditioned. Projection removed the overlap between fractional powers and the polynomial span, while continued limits preserved a fractional direction when powers coincided. Later enrichment added an even fourth-order tail and repeated the projection after extending the polynomial core. A pair of rational terms supplied further shape freedom. These records describe one search history, without implying a common path across the five searches.

Away from power collisions, the constructions at iterations 3 and 4 span the same generic space,
\begin{equation}
 (1-v)^2\left[\mathcal P_6(v)+\operatorname{span}\{v^p,v^{p+1}\}\right].
\label{supp:eq:generic-space}
\end{equation}
Their change concerns coordinates and their continuation as the exponent moves. The later fixed-$\beta$ comparison isolates the Gram map in the final space (Section~\ref{supp:tail-capable-controls}); the native history combines several changes. No intermediate field errors are inferred from the residual sequence.

\FloatBarrier

\subsubsection{Transient Fujita: local reaction response}\label{supp:fujita-transient}

ASYS separated heat spreading from reaction-driven amplification by writing $u=v/\sqrt q$. Here $v$ is an explicit approximation to the heat evolution of the initial field, with an analytic boundary correction. With $c=\cos(\pi r/10)$ and fitted time scale $T$, the denominator is
\begin{equation}
 \begin{aligned}
 q(r,t)&=c^2 C(t)+(1-c^2)W(r,t),\\
 W(r,t)&=\frac{\exp E(r,t)}{1+\gamma F(t/T,c)},\\
 F(s,c)&=\frac{-\log(1-sc^2)-sc^2-(sc^2)^2/2-(sc^2)^3/3}{c^4}.
 \end{aligned}
\end{equation}
The positive core factor $C$ describes amplification at the origin, and $E$ controls the radial width. Their first three initial time derivatives were fixed from the PDE; fitted polynomial corrections begin at fourth order, with heat responses enforcing the wall condition.

The new response $F$ makes reaction advance at a position-dependent rate through $sc^2$. To construct it, ASYS used the denominator equation obtained for a heat numerator satisfying $v_t=\Delta_5v$, where $\Delta_5=\partial_{rr}+4\partial_r/r$:
\begin{equation}
 q_t=\Delta_5q+2\frac{v_r}{v}q_r-\frac32\frac{q_r^2}{q}-2v^2.
\end{equation}
With $a(r)=u(r,0)$, substituting the leading reaction denominator $q_0=1-2a^2t$ gives a gradient contribution $-24t^2a^2a_r^2/(1-2a^2t)$. Its time integral contains $-3a_r^2 L_3(2a^2t)/a^4$, where $L_3(z)=-\log(1-z)-z-z^2/2$. The cubic term is already supplied by the fixed initial derivatives. Removing it and using the fitted time scale in this higher-order response yields $F$ above. Its continuous limit $F\sim s^4c^4/4$ at the wall preserves the boundary behavior. The fitted values were $T=0.0204059996$ and $\gamma=0.0118439$.

\paragraph{Complete fitted expression.}
\label{supp:fujita-complete-formula}
The following formulas specify the selected fourteen-scalar construction on
$0\le r\le5$ and $0\le t\le0.020385594<T$.
Set $k=\pi/10$, $x=kr$, $c=\cos x$, $s=t/T$ and $\rho=r^2/25$.
The field and its positive reaction denominator are
\begin{equation}
 u(r,t)=\frac{v(r,t)}{\sqrt{q(r,t)}},\qquad
 q=c^2C(t)+(1-c^2)\frac{e^{E(r,t)}}{1+\gamma F(s,c)},\qquad
 F(s,c)=\sum_{m=4}^{\infty}\frac{s^m c^{2m-4}}{m}.
 \label{eq:fujita-complete-field}
\end{equation}
The series is the regular form of Eq.~\eqref{eq:fujita-local-response};
in particular, $F(s,0)=0$ and $F=s^4c^4/4+O(s^5)$.

\paragraph{Radial and wall responses.}
Three even analytic radial functions remove the apparent origin quotients:
\begin{equation}
 S(x)=\frac{\sin x}{x},\qquad
 J(x)=\frac{S(x)-\cos x}{x^2},\qquad
 K(x)=\frac{J(x)-\cos x/3}{x^2},\qquad
 (S(0),J(0),K(0))=(1,1/3,2/15).
 \label{eq:fujita-radial-functions}
\end{equation}
All occurrences of $S,J,K$ below are evaluated at $x=kr$.
The regular inward wall coordinate and spherical prefactor are
\begin{equation}
 d(r)=\frac{c}{k}\left(1+\frac{c^2}{6}+\frac{3c^4}{40}
 +\frac{5c^6}{112}+\frac{35c^8}{1152}+\frac{63c^{10}}{2816}\right),
 \qquad A(r)=\sum_{\ell=0}^{7}(1-\rho)^\ell.
 \label{eq:fujita-wall-coordinate}
\end{equation}
These are regular extensions of $5-r$ and $(5/r)^2$ near the wall;
$d(5)=0$, $d_r(5)=-1$, and both are even at the origin.

For $t>0$, put $z=d/(2\sqrt t)$ and define the finite heat-response recurrence
\begin{equation}
 \begin{aligned}
 \mathcal J_1(z)&=\operatorname{erfc}z,&
 \mathcal J_{3/2}(z)&=2\left(\frac{e^{-z^2}}{\sqrt\pi}
                         -z\operatorname{erfc}z\right),\\
 \mathcal J_{\nu+1}(z)&=
 \frac{\mathcal J_\nu(z)-z\mathcal J_{\nu+1/2}(z)}{\nu},
 &&\nu=1,\tfrac32,\ldots,\tfrac{13}{2},\\
 H_1&=\mathcal J_2(z),\qquad H_2=2\mathcal J_3(z),&
 I_{5/2}&=t^{3/2}\mathcal J_{5/2}(z),\\
 N_m&=-m!\sqrt t\,\mathcal J_{m+3/2}(z),&&m=2,\ldots,6.
 \end{aligned}
 \label{eq:fujita-heat-responses}
\end{equation}
Here $I_\nu=t^{\nu-1}\mathcal J_\nu$ is the half-line heat response
with Laplace transform $e^{-d\sqrt p}/p^\nu$.
At $d=0$, $\mathcal J_\nu(0)=1/\Gamma(\nu)$, so $H_1=H_2=1$.
The functions $t^mN_m$ have inward derivative $t^m$ at the wall.
At fixed $r<5$, these wall responses and all their time derivatives vanish
as $t\downarrow0$; the initial field is defined by $u(r,0)=5c$.

\paragraph{Heat numerator.}
With $g_1=40k^2/\pi$, $g_2=-160k^4/\pi^3$ and $V_0=k^2-2/25$,
\begin{equation}
 \begin{aligned}
 v(r,t)=e^{-k^2t}\Bigg\{&5c-20k^2tS+20k^4t^2J\\
 &+A(r)\left[g_1tH_1+g_2t^2H_2
 +V_0g_1t^2(H_1-H_2)
 -\frac{4g_1}{125}\left(\frac{d^2I_{5/2}}4
                         +\frac{dt^2H_2}{8}\right)\right]\Bigg\}.
 \end{aligned}
 \label{eq:fujita-complete-numerator}
\end{equation}
The first line is the exact heat evolution of $5\cos(kr)$ on the full radial
space: $(\Delta_5+k^2)c=-4k^2S$, $(\Delta_5+k^2)S=-2k^2J$ and
$(\Delta_5+k^2)J=0$. The finite wall correction enforces $v(5,t)=0$.
The resulting bounded-domain numerator approximates the heat evolution while retaining the initial interior time derivatives of the full-space heat field.

\paragraph{Core and width factors.}
Let $a_0,a_1,a_2$ be the three core coefficients and $b_{ij}$,
$i,j=0,1,2$, the nine width coefficients. The core factor is
\begin{equation}
 \begin{aligned}
 C(t)&=(1-s)\exp\left[(1-50T)s+\kappa_2s^2+\kappa_3s^3
                         +s^4\sum_{j=0}^2a_js^j\right],\\
 \kappa_2&=\frac{1+(1000k^2-2500)T^2}{2},\qquad
 \kappa_3=\frac13+
 \frac{(-14000k^4+150000k^2-250000)T^3}{6}.
 \end{aligned}
 \label{eq:fujita-complete-core}
\end{equation}
It has $C(0)=1$, $C'(0)=-50$, $C''(0)=1000k^2$ and
$C'''(0)=-14000k^4$.
Define the fixed radial derivatives
\begin{equation}
 \begin{aligned}
 B_2(r)&=k^2\left(\frac{800cJ}{S^2}-300\right),\\
 B_3(r)&=k^4\left[-\frac{2400cK+19200cJ+7200x^2J^2}{S^2}
                    +3000+\frac{7200J}{S}\right]-30000k^2c^2,\\
 \sigma_3&=k^5\left[\frac{12000}{\pi/2}
                         +\frac{2400}{(\pi/2)^3}\right].
 \end{aligned}
 \label{eq:fujita-width-jets}
\end{equation}
Then the complete width exponent is
\begin{equation}
 \begin{aligned}
 E(r,t)={}&\frac{t^2}{2}B_2(r)+\frac{t^3}{6}B_3(r)
                  +s^4\sum_{i,j=0}^2b_{ij}\rho^i s^j\\
 &-\frac{4\pi^2}{5}t^2N_2+\frac{\sigma_3}{6}t^3N_3
       +\frac25s^4\sum_{j=0}^2(b_{1j}+2b_{2j})s^jN_{j+4}.
 \end{aligned}
 \label{eq:fujita-complete-width}
\end{equation}
The second line cancels the wall slopes of the first, since
$B_2'(5)=-8\pi^2/5$ and $B_3'(5)=\sigma_3$.
Thus $q_r(5,t)=0$ for $t>0$.
Because the fitted shape terms and $F$ begin at fourth order in time and the wall responses are flat initially at every fixed interior radius, the first three initial interior time derivatives prescribed by the PDE are retained.

\paragraph{Fitted physical coefficients.}
The selected coefficients appearing directly in these formulas are
\begin{equation}
 \begin{gathered}
 T=0.020405999578311518,\qquad
 \gamma=0.011843902040415752,\\
 (a_0,a_1,a_2)=10^{-4}
 (1.681220486649258,\ 1.999823043478597,\ 2.845556756446939),\\
 (b_{ij})_{i,j=0}^{2}=10^{-4}
 \begin{pmatrix}
 -0.2540198591897803 & 3.129386111045238 & 3.131464042952323\\
 -5.298560439074358 & -5.161234640063227 & -7.873003917126902\\
 5.507034497237934 & 5.434588429101822 & 2.625284496418639
 \end{pmatrix}.
 \end{gathered}
 \label{eq:fujita-physical-coefficients}
\end{equation}
Rows of $(b_{ij})$ index radial powers and columns index time powers.

\FloatBarrier

\subsubsection{NLS: discrete response construction}\label{supp:nls-formula}
The zero-initial-value response is
\begin{equation}
\mathcal G_L[S](t)=-i\int_0^t e^{\frac i2(t-s)\partial_{xx}}S(s)\,ds.
\end{equation}
The response sources are projected onto cosine modes $0,\ldots,24$ by $\Pi_{24}$. Methods Section~\ref{methods:nls-construction} specifies their quadrature, interpolation and numerical representation.

With $Z=C+B$ and $N(z)=|z|^2z$, the response map in the main text is
\[
\mathcal F(d)=\mathcal G_L[-\Pi_{24}(N(Z+d)-N(C))].
\]
Its linearized feedback and finite response are
\begin{equation}
A_Zv=\mathcal G_L[-\Pi_{24}(2|Z|^2v+Z^2\bar v)],
\qquad
\mathcal R_Z(v)=\sum_{k=0}^{40}A_Z^kv.
\label{eq:dressing}
\end{equation}
The defect construction uses
\[
\Delta=\mathcal F(P)-P,\qquad E=\mathcal R_{Z+P}(\Delta).
\]

The predictor contains four successive responses to the nonlinear mismatch of the background $Z=C+B$. With $N(z)=|z|^2z$, write the quadratic and cubic interactions as
\[
Q_Z(f,g)=2Z\operatorname{Re}(f\bar g)+\bar Zfg,
\qquad
T(f,g,h)=\frac{fg\bar h+fh\bar g+gh\bar f}{3}.
\]
Here $f,g,h$ are correction fields and bars denote complex conjugation. The source hierarchy is
\begin{equation}
\begin{aligned}
s_1&=-[N(C+B)-N(C)],&s_2&=-Q_Z(d_1,d_1),\\
s_3&=-2Q_Z(d_1,d_2)-T(d_1,d_1,d_1),\\
s_4&=-2Q_Z(d_1,d_3)-Q_Z(d_2,d_2)-3T(d_1,d_1,d_2),\\
d_j&=\mathcal R_Z\bigl(\mathcal G_L[\Pi_{24}s_j]\bigr),&
P&=\sum_{j=1}^4d_j .
\end{aligned}
\label{eq:nls-hierarchy}
\end{equation}
Multiplying the initial source $s_1$ by a bookkeeping parameter $\eta$, expanding $d=\sum_j\eta^jd_j$, and setting $\eta=1$ gives these four orders. The expansion counts successive responses to the full mismatch forcing.

For $u=t/T_{\max}$ and $T_{\max}=\pi/2$, the first response divided by $u^2$ has a finite analytic endpoint,
\begin{equation}
\lim_{u\to0}\frac{\mathcal G_L[S](uT_{\max})}{u^2}
=-\frac{iT_{\max}^2}{2}\partial_tS(0).
\end{equation}
For $S=-[N(C+B)-N(C)]$, this becomes $T_{\max}^2\Pi_{24}[C(0)^2S_{\rm rad}(0)]/2$, with $B_t(0)=-iS_{\rm rad}(0)$. Higher nonlinear sources and positive feedback powers have zero scaled endpoint. The leading response in the complete defect cancels the corresponding term in $P$, so $E/u^2$ also has zero endpoint.

Let $\widehat\Phi_{knm}$ denote the stored Chebyshev coefficients of the eight orthonormal directions and $M$ their fixed coordinate transformation, constructed as described in Methods Section~\ref{methods:nls-construction}. With cosine multiplicities $\mu_0=1$ and $\mu_{n>0}=2$, the field is
\begin{equation}
\psi=C+B+\sum_{n=0}^{24}\mu_n\cos(n\pi x/L)F_n(t),\qquad
F_n(t)=u^2\sum_{k=0}^{320}T_k(2u-1)(\widehat\Phi M\alpha)_{kn}.
\label{supp:eq:nls_modal_eval}
\end{equation}
For the reported field, $\alpha$ is fixed to represent $P+E$; the predictor comparison fixes it to represent $P$. Here $T_k$ are the Chebyshev polynomials.

\subsubsection{Allen--Cahn: coupled interface harmonics}\label{supp:ac-mode-coupling}

The main construction represents a nearly circular interface by Eq.~\eqref{eq:ac-polar-interface}. For inward normal speed $c\kappa$, the exact polar curvature law at fixed angle is
\begin{equation}
\partial_t r=-c\frac{r^2+2r_\vartheta^2-r r_{\vartheta\vartheta}}
{r(r^2+r_\vartheta^2)}.
\label{eq:ac-polar-curvature}
\end{equation}
Its enclosed area decreases at rate $2\pi c$. The area normalization in the ansatz therefore separates $R(t)^2=R_0^2-2ct$ from the angular shape amplitudes. Expanding the law to quadratic order couples products of modes into their sum and difference harmonics.

The second mode decays as $e_2=a s$ at linear order, with $s=R(t)^2/R_0^2$. Isolating its quadratic forcing of the fourth mode gives
\begin{equation}
\dot e_2=-\frac{2c}{R^2}e_2,\qquad
\dot e_4=-\frac{14c}{R^2}e_4+\frac{11c}{2R^2}e_2^2.
\end{equation}
Integrating with $e_4(0)=b$ yields Eq.~\eqref{eq:ac-fourth-mode}. The coefficient $11/20$ is fixed by this integration. Other mode interactions are omitted from this isolated example, but retained through mode six in the quadratic construction.

More generally, write $p_m=(m^2-2)/2$ and $a_m=e_m(0)$. The fourteen-parameter construction uses
\begin{equation}
 e_m(s)=a_ms^{p_m}+\sum_{(n,k)\to m}
 \frac{C_{mnk}a_na_k}{2(p_m-p_n-p_k)}\left(s^{p_n+p_k}-s^{p_m}\right),
 \qquad m=2,\ldots,6.
\end{equation}
Each unordered pair $2\le n\le k\le6$ contributes to retained sum and difference modes. For $n<k$, the coefficients are $C_{n+k,n,k}=n^2+k^2+nk-1$ and $C_{k-n,n,k}=n^2+k^2-nk-1$; for $n=k$, $C_{2n,n,n}=(3n^2-1)/2$. Terms whose target lies outside modes two through six are omitted. The denominators are nonzero for these interactions. Only the linear amplitudes enter the forcing, giving a finite quadratic expansion with analytically integrated time dependence.

\paragraph{From the polar curve to the diffuse field.}
The moving curve is embedded in a field with width $w=\sqrt2\epsilon$. In coordinates $(X,Y)$ rotated by the fitted angle $\theta_0$ about $(c_x,c_y)$, set $\rho=(X^2+Y^2+\kappa^2)^{1/2}$, $\kappa=0.015$, and
\[
 H_m+iS_m=\left(\frac{X+iY}{\rho}\right)^m,\qquad
 B=\frac{\sqrt{R^2+\kappa^2}}{\sqrt{1+\frac12\sum_{m=2}^6e_m^2}}
 \left(1+\sum_{m=2}^6e_mH_m\right).
\]
Let $B_\vartheta$ and $B_{\vartheta\vartheta}$ denote angular derivatives at fixed $\rho$. With $o=B-\rho$ and $\nu=\sqrt{1+(B_\vartheta/B)^2}$, the approximate normal distance is
\begin{equation}
 d=\frac{o}{\nu}+\frac{\chi o^2}{1+(o/0.12)^4},\qquad
 \chi=\frac{B_\vartheta^2(BB_{\vartheta\vartheta}-B^2-2B_\vartheta^2)}{2\nu^5B^5}.
 \label{supp:eq:ac-normal-distance}
\end{equation}
The quadratic term follows from the local distance condition $|\nabla d|^2=1$ and is damped away from the interface. The medial regularization removes the polar singularity. Exact area normalization concerns the ideal polar curve in Eq.~\eqref{eq:ac-polar-interface}; the complete diffuse field also contains the following terms.

For each supplied disk, let $r_j^2=|(x,y)-c_j|^2$ and $\rho_{j,\delta}=\sqrt{r_j^2+\delta^2e^{-r_j^2/\delta^2}}$. The initial distance-like phase and field are
\[
 \ell_j=R_j-\rho_{j,\delta_0},\qquad
 q_0=\frac w2\log\left(e^{2\ell_1/w}+e^{2\ell_2/w}+e^{2(\ell_1+\ell_2)/w}\right),\qquad
 u_0=\tanh(q_0/w),\quad \delta_0=0.002.
\]
Because this small core smoothing differs from the prescribed radial floor, its initial field error is measured rather than assumed zero. The two disk centers and radii are those in Section~\ref{supp:allen_cahn}.

The initial peanut and moving polar phase are joined by
\begin{equation}
\begin{aligned}
 q(x,y,t)&=(1-b_t)(q_0+R-R_0)+b_t d+q_n,\\
 b_t&=1-e^{-k_b t},\\
 q_n&=A_n(1-e^{-k_n t})^2e^{-0.15t}
       \exp\left[-(\ell/0.13)^2-(n/0.23)^2\right].
\end{aligned}
\label{supp:eq:ac-blended-phase}
\end{equation}
The coordinates along and across the neck are
\[
 \ell=\frac{0.33(x-0.039)-0.05(y+0.009)}{\sqrt{0.33^2+0.05^2}},\qquad
 n=\frac{0.05(x-0.039)+0.33(y+0.009)}{\sqrt{0.33^2+0.05^2}}.
\]
This localized term adjusts the neck while vanishing in both value and first time derivative at $t=0$.

An additive transient supplies the initial PDE velocity. Its spatial derivatives use $\delta(t)^2=\delta_0^2+4\epsilon^2t$. Define the three weights $\omega=\operatorname{softmax}[2(\ell_1,\ell_2,\ell_1+\ell_2)/w]$, the vectors $G_1=-\nabla\rho_{1,\delta(t)}$, $G_2=-\nabla\rho_{2,\delta(t)}$, $G_3=G_1+G_2$, and
\[
 G=\sum_{j=1}^3\omega_jG_j,\qquad
 L=\sum_{j=1}^3\omega_j\nabla\!\cdot G_j+
 \frac2w\left(\sum_{j=1}^3\omega_j|G_j|^2-|G|^2\right).
\]
The required velocity and the initial velocity of the blended phase give
\begin{equation}
\begin{aligned}
 J(x,y,t)&=(1-u_0^2)\left[\frac{\epsilon^2L}{w}+u_0(1-|G|^2)\right],\\
 V_q(x,y)&=\frac{1-u_0^2}{w}\left[-\frac c{R_0}+k_b(d(x,y,0)-q_0)\right],\\
 \widetilde u&=\tanh(q/w)+te^{-k_\ell t}(J-V_q).
\end{aligned}
\label{supp:eq:ac-initial-layer}
\end{equation}
At $t=0$, $J$ is the Allen--Cahn velocity of the smoothed union before outer closure, while its growing smoothing length regularizes derivatives within the decaying transient.

Finally, for $z_q=\operatorname{clip}[(|q|-0.6)/0.4,0,1]$, let $W(q)=1-z_q^3(10-15z_q+6z_q^2)$. The returned field is
\[
 u(x,y,t)=-1+W(x)W(y)[1+\widetilde u(x,y,t)].
\]
This closure makes values and the first two spatial derivatives match at opposite boundaries. The displayed interfaces are zero contours of this complete field. Table~\ref{tab:ac-curvature-coefficients} lists the fitted constants of both fields.

\begin{table}[htbp]
\centering\small
\caption{Fitted physical constants in the two coupled-mode fields. The five $a_m$ are initial polar amplitudes. The twelve-parameter extension has no independent $a_5$ or $a_6$. Values are rounded for display.}
\label{tab:ac-curvature-coefficients}
\begin{tabular}{lrr}
\toprule
Constant & Quadratic construction & Selected cubic extension\\
\midrule
$R_0$ & $0.279659907$ & $0.280675398$\\
$a_2$ & $0.303722205$ & $0.312116045$\\
$a_3$ & $0.051702899$ & $0.065560151$\\
$a_4$ & $0.017843257$ & $0.019740090$\\
$a_5$ & $0.036411614$ & $0$\\
$a_6$ & $0.012567262$ & $0$\\
$c_x$ & $-0.022573204$ & $-0.022675420$\\
$c_y$ & $-0.000570287$ & $0.000341406$\\
$\theta_0$ & $-0.149890467$ & $-0.146447002$\\
$c$ & $0.00152276736$ & $0.00155867936$\\
$k_b$ & $0.334201339$ & $0.407834936$\\
$k_\ell$ & $2.004330677$ & $2.004330680$\\
$A_n$ & $0.012264453$ & $0.009262527$\\
$k_n$ & $1.258173814$ & $1.499999986$\\
\bottomrule
\end{tabular}
\end{table}

\paragraph{Twelve-parameter extension.}
This variant generates the fifth and sixth harmonics from the first three and retains selected cubic feedback. With $a=a_2$, $d_3=a_3$ and $b=a_4$, it uses
\begin{equation}
\begin{aligned}
 e_2&=s\left[a+\frac{23}{40}a^3(1-s^2)+\left(\frac{11}{14}ab-\frac{121}{280}a^3\right)(1-s^7)\right],\\
 e_3&=d_3s^{7/2},\qquad
 e_4=bs^7+\frac{11}{20}a^2(s^2-s^7),\\
 e_5&=\frac97ad_3(s^{9/2}-s^{23/2}),\\
 e_6&=\frac32a\left(b-\frac{11}{20}a^2\right)(s^8-s^{17})
       +\frac{51}{280}a^3(s^3-s^{17})+\frac{13}{20}d_3^2(s^7-s^{17}).
\end{aligned}
\end{equation}
The diffuse-field assembly is unchanged. This is a selected cubic extension, rather than a complete third-order expansion.

\paragraph{Curvature checks.}
Under the test in Methods Section~\ref{methods:ac-protocol}, the fourteen-scalar quadratic program's projected mode-velocity discrepancy was $3.43783\times10^{-6}$, $4.35226\times10^{-7}$ and $5.47970\times10^{-8}$ at deformation amplitudes $\eta=0.04$, $0.02$ and $0.01$. The successive reductions by factors of $7.90$ and $7.94$ are consistent with a cubic remainder (Fig.~\ref{fig:ac-mode-coupling}f).

These checks concern small shape deformations under curvature motion. The finite-width field, including the neck and initial transient, is assessed separately in Section~\ref{supp:ac-coupled-metrics}.

\paragraph{Generated shape change.}
Independent evolutions from pure elongation developed the fourth harmonic predicted by Eq.~\eqref{eq:ac-fourth-mode} (Table~\ref{tab:ac-generated-harmonic}). Doubling $a$ increased the final PDE amplitude by factors of $3.994$ and $3.996$ at $\epsilon=0.02$ and $0.04$, respectively. Halving the interface thickness reduced the amplitude discrepancy from approximately $6\%$ to $2\%$. Direct evolution of the polar curvature equation agreed with the quadratic prediction to within $0.23\%$.

\begin{table}[htbp]
\centering\small
\caption{Fourth harmonic generated from pure elongation, measured at $s=0.7$. All initial fourth harmonics are zero. The prediction uses $c=\epsilon^2$ for the PDE and the prescribed physical time. The discrepancy is $100|e_4-e_4^{\rm prediction}|/|e_4^{\rm prediction}|$.}
\label{tab:ac-generated-harmonic}
\begin{tabular}{@{}lccrrr@{}}
\toprule
Evolution & $\epsilon$ & $a$ & Predicted $10^3e_4$ & Measured $10^3e_4$ & Discrepancy (\%) \\
\midrule
Polar curvature & --- & 0.04 & 0.358728 & 0.358529 & 0.0554 \\
Polar curvature & --- & 0.08 & 1.434913 & 1.431718 & 0.223 \\
Allen--Cahn PDE & 0.04 & 0.04 & 0.358728 & 0.336934 & 6.08 \\
Allen--Cahn PDE & 0.04 & 0.08 & 1.434913 & 1.346222 & 6.18 \\
Allen--Cahn PDE & 0.02 & 0.04 & 0.358728 & 0.352206 & 1.82 \\
Allen--Cahn PDE & 0.02 & 0.08 & 1.434913 & 1.406731 & 1.96 \\
\bottomrule
\end{tabular}
\end{table}

For the $a=0.08$, $\epsilon=0.02$ PDE run, spatial and temporal refinement changed $e_4$ by at most $2.67\times10^{-10}$ and $1.40\times10^{-13}$ over the trajectory. Doubling the angular extraction grid changed it by $4.11\times10^{-14}$ at the checked times. The circular controls had fourth-mode amplitudes below $2.92\times10^{-10}$, compared with the smallest nonzero predicted endpoint amplitude of $3.59\times10^{-4}$. The complete PDE and polar-curvature test consumed $21.5$ CPU-minutes across nine single-core jobs.

\FloatBarrier
\subsection{Comparison results}

\subsubsection{Single-session coding controls}
\label{supp:single-session-controls}

Table~\ref{tab:single-session-controls} reports all six single-session outputs and the original reported ASYS constructions. All three gCLM fits completed 20,000 updates and passed the fixed equation evaluator's qualification checks.

The three gCLM programs combine a Chebyshev polynomial core with fractional-tail corrections. The first two span the same nine-parameter family with differently scaled coefficient coordinates; the third adds a further fractional term. Although all three admit the secondary tail, none constructs the projected fractional and rational basis or the Gram-coordinate transformation developed in later iterations of the selected ASYS search, beyond the tail correction already present in its first revised candidate.

All three Fujita programs use spatially varying logarithms and preserve the first initial interior PDE time derivative. None constructs the fourth-order response in Eq.~\eqref{eq:fujita-local-response} while preserving the first three initial derivatives, as the reported ASYS formula does.

\begin{table}[htbp]
\centering\small
\caption{Single-session coding controls. $n_\theta$ counts fitted scalars, including the gCLM exponent. The primary error is the gCLM profile error on 257 logarithmic radii or the Fujita space--time field error; the final-time column applies to Fujita. All errors are relative $L^2$ percentages. ASYS rows use the original reported constructions and coefficients. The gCLM ASYS row reports the selected construction; the median across five ASYS searches was $3.90\times10^{-5}\%$. Methods, Section~\ref{method:single-session}, specifies the generation and fitting procedures.}
\label{tab:single-session-controls}
\begin{tabular}{llrrr}
\toprule
Task & Construction & $n_\theta$ & Primary error (\%) & Final-time error (\%) \\
\midrule
gCLM & Single session 1 & 9 & $7.70\times10^{-4}$ & --- \\
& Single session 2 & 9 & $7.83\times10^{-4}$ & --- \\
& Single session 3 & 10 & $1.89\times10^{-3}$ & --- \\
& ASYS & 15 & $2.47\times10^{-8}$ & --- \\
\midrule
Fujita & Single session 1 & 31 & 2.10 & 7.29 \\
& Single session 2 & 16 & 0.222 & 0.406 \\
& Single session 3 & 35 & 2.09 & 5.51 \\
& ASYS & 14 & 0.0879 & 0.702 \\
\bottomrule
\end{tabular}
\end{table}

\subsubsection{gCLM: tail-capable comparisons and coordinate controls}
\label{supp:tail-capable-controls}

Table~\ref{tab:gclm-tail-capable} gives the four equation-selected SIGS/FEX outcomes underlying the medians in Table~\ref{tab:accuracy-comparison}. Methods, Section~\ref{method:gclm-comparisons}, specifies the supplied coordinate, search allocations and fitting procedure.

\begin{table}[htbp]
\centering\small
\caption{SIGS and FEX adaptations supplied with the fractional coordinate. Each row is one equation-selected search outcome after a 20,000-update final fit, including the fitted exponent $\beta$. Field error is unweighted relative $L^2$ in percent on 257 logarithmic radii in $[10^{-3},10^3]$.}
\label{tab:gclm-tail-capable}
\begin{tabular}{lrr}
\toprule
Adaptation & Residual MSE & Field error (\%)\\
\midrule
SIGS, seed 0 & $2.16\times10^{-10}$ & $5.67\times10^{-3}$\\
SIGS, seed 1 & $1.31\times10^{-9}$ & $1.82\times10^{-2}$\\
FEX-derived, seed 0 & $6.52\times10^{-10}$ & $7.64\times10^{-3}$\\
FEX-derived, seed 1 & $3.04\times10^{-9}$ & $9.38\times10^{-3}$\\
\bottomrule
\end{tabular}
\end{table}

Because the adapted expressions admit the secondary fractional tail, its absence alone cannot explain the remaining field-error gap; this comparison concerns the achieved accuracy of these adaptations, whose expression spaces and search budgets differ from ASYS.

\paragraph{Matched conditioning in both families.}
The four fits in Table~\ref{tab:gclm-matched-conditioning} compare identity and Gram coordinates for the polynomial and ASYS families under the common protocol in Methods, Section~\ref{method:gclm-conditioning}. All four completed 20,000 updates; fitting took 1,607--2,577 seconds.

\begin{table}[htbp]
\centering\small
\caption{Matched four-arm conditioning comparison. Profile error is relative $L^2$ in percent on 257 logarithmic radii in $[10^{-3},10^3]$. Signed tail-balance error is $100[D/(C^2K)-1]$, where $K$ is the PDE relation in Eq.~\eqref{supp:eq:K-def}. All four fits use fixed $\beta$, fourteen shape coefficients and 20,000 updates.}
\label{tab:gclm-matched-conditioning}
\begin{tabular}{llrrr}
\toprule
Family & Coordinates & Residual MSE & Profile error (\%) & Tail error (\%)\\
\midrule
Polynomial & Identity & $6.89\times10^{-11}$ & $7.18\times10^{-4}$ & $-11.7548$\\
Polynomial & Gram & $1.45\times10^{-7}$ & $2.35\times10^{-1}$ & $-14.9825$\\
ASYS & Identity & $1.12\times10^{-12}$ & $4.81\times10^{-4}$ & $+1.49248$\\
ASYS & Gram & $8.72\times10^{-20}$ & $2.23\times10^{-8}$ & $+0.00249738$\\
\bottomrule
\end{tabular}
\end{table}

The ASYS Gram fit has about 32,200 times lower profile error than the better polynomial fit, which uses identity coordinates (Table~\ref{tab:gclm-matched-conditioning}). Gram coordinates improve ASYS's error by a factor of about 21,600 but worsen the polynomial's by a factor of about 327. At the fixed exponent, the transformation reduces the basis Gram condition number from $2.23\times10^8$ to $1.07$ for the polynomial and from $1.60\times10^8$ to $95.8$ for ASYS. A nearly orthogonal basis alone therefore does not ensure a well-conditioned nonlinear fitting problem; these results concern optimization under the specified Adam schedule rather than the best attainable approximation in either family. Repeating the frozen readout on all four reference banks changes profile errors by at most $3.91\times10^{-11}$ percentage points. The secondary-tail discrepancy remains distinct from the sampled profile error.

\subsubsection{Transient Fujita: response and coefficient effects}
\label{supp:fujita-frozen-responses}

Table~\ref{tab:fujita-frozen-responses} evaluates both responses at the same three coefficient vectors. At the local-fit coefficients, the local response improves trajectory accuracy by a factor of $3.32$ and final-time accuracy by $1.50$. At the uniform-fit coefficients, the two responses yield nearly identical fields because fitting reduces the response strength to $\gamma=3.67\times10^{-6}$; in each pair, the central traces coincide because the response enters $q$ through $1-c^2$, which vanishes at the origin.

\begin{table}[htbp]
\centering\small
\caption{Fujita responses at fixed coefficients. Initial denotes the common initialization of the developed fourteen-parameter formula; local fit and uniform fit denote the two separately fitted coefficient vectors. Equation objectives use the same normalized residual and weighting on training or independent points. Field errors are relative $L^2$ percentages. No parameters are fitted in this diagnostic; Methods, Section~\ref{method:fujita-comparison}, specifies the evaluation.}
\label{tab:fujita-frozen-responses}
\begin{tabular}{llrrrr}
\toprule
Coefficients & Response & Training loss & Independent loss & Trajectory (\%) & Final (\%)\\
\midrule
Initial & Local & $1.03\times10^{-7}$ & $6.83\times10^{-8}$ & $0.07975$ & $0.6764$\\
Initial & Uniform & $2.63\times10^{-2}$ & $2.25\times10^{-2}$ & $0.2969$ & $1.0530$\\
Local fit & Local & $4.29\times10^{-8}$ & $3.18\times10^{-8}$ & $0.08793$ & $0.7021$\\
Local fit & Uniform & $2.63\times10^{-2}$ & $2.25\times10^{-2}$ & $0.2917$ & $1.0561$\\
Uniform fit & Local & $1.29\times10^{-6}$ & $1.22\times10^{-6}$ & $3.9907$ & $25.4334$\\
Uniform fit & Uniform & $1.34\times10^{-6}$ & $1.19\times10^{-6}$ & $3.9907$ & $25.4334$\\
\bottomrule
\end{tabular}
\end{table}

Fitting lowers the independent residual objective for both forms, by a factor of about $2.15$ for the local response and about $19{,}000$ for the uniform response, while increasing their field errors. The residual--field discrepancy therefore persists away from the fitting points. The factor of $45$ between the separately fitted fields measures the outcome of the common equation-based fitting procedure; it does not determine either family's best attainable field approximation.

\subsubsection{Transient Fujita: neural sampling comparison}
\label{supp:fujita-neural-control}

The neural correction and paired sampling protocols are specified in Methods, Section~\ref{method:fujita-neural}. For the 720-point fits, the residual objective on independent points was 719--2,523 times its training value.

Denser sampling reduced the independent-to-training residual ratio to 2.84--6.18 and lowered the median trajectory error from $6.33\%$ to $5.30\%$ (Table~\ref{tab:fujita-neural-fits}), though the fitted fields still overestimated the final core amplitude (Fig.~\ref{fig:fujita-neural-controls}). Table~\ref{tab:accuracy-comparison} reports the denser setting. This neural comparison assesses a different representation and fitting procedure; the local-versus-uniform comparison changes the analytical response under a common fitting procedure.

\begin{table}[htbp]
\centering\small
\caption{Fujita neural sampling comparison. All errors are relative $L^2$ percentages for the returned states. Each pair shares its initial model state and fitting procedure. The 720-point fits stopped after 2,407--2,481 seconds; the 5,760-point fits reached the 3,585-second training cutoff.}
\label{tab:fujita-neural-fits}
\begin{tabular}{lrrrr}
\toprule
 & \multicolumn{2}{c}{720 points} & \multicolumn{2}{c}{5,760 points}\\
\cmidrule(lr){2-3}\cmidrule(lr){4-5}
Repeat & Trajectory & Final time & Trajectory & Final time\\
\midrule
1 & 7.903 & 146.94 & 6.008 & 84.01\\
2 & 6.094 & 86.88 & 5.300 & 66.56\\
3 & 6.331 & 92.91 & 5.013 & 63.48\\
\bottomrule
\end{tabular}
\end{table}

\begin{figure}[htbp]
\centering
\includegraphics[width=\linewidth]{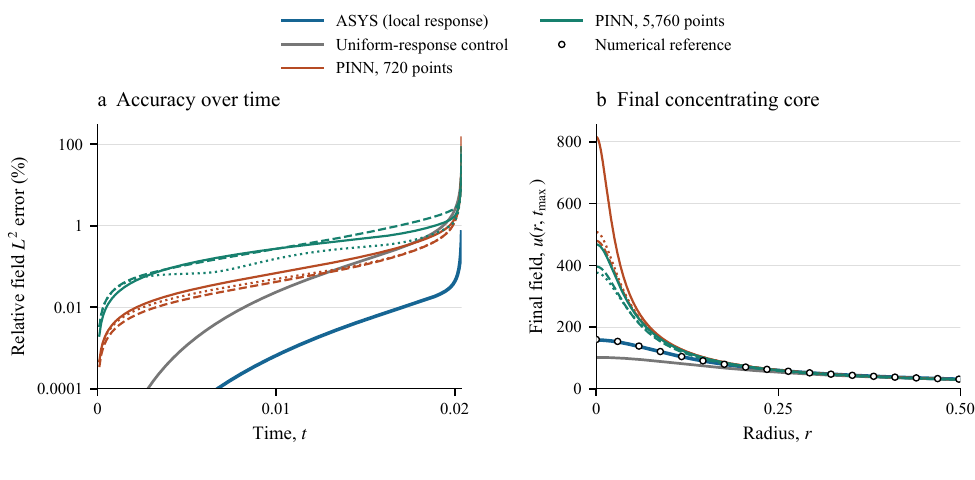}
\caption{\textbf{Fujita neural corrections at two sampling densities.} Both settings start from the supplied concentrating ansatz and use the same one-hour fitting allowance. Solid, dashed and dotted curves distinguish the three paired initializations; all curves use the returned states in Table~\ref{tab:fujita-neural-fits}. The local and uniform analytical responses retain their separate 300-second comparison. \textbf{a,} Relative radial-profile error across the full domain over time; the common initial field has zero error and is omitted from the logarithmic plot. \textbf{b,} Final profiles in the concentrating core. Numerical-reference samples are shown as open circles.}
\label{fig:fujita-neural-controls}
\end{figure}

\FloatBarrier
\subsubsection{Allen--Cahn: a physics-informed neural baseline}
\label{supp:ac-current-neural-controls}

The neural field and fitting protocol are specified in Methods, Section~\ref{method:ac-neural}. All zero contours were present at every evaluation time.

All three neural fits completed within their allowance, taking about 1,186 seconds each. Their median trajectory error was $0.122\%$, compared with $0.798\%$ for the reported ASYS field. Table~\ref{tab:ac-current-neural-controls} reports each neural fit; Fig.~\ref{fig:ac-neural-controls} compares their field and interface errors with the original ASYS construction.

\begin{table}[htbp]
\centering\small
\caption{Allen--Cahn PINN baseline. Each fit uses 12,801 parameters, the same equation loss and L-BFGS settings, and a 1,200-second allowance. Trajectory, final-time and reference-band errors are relative $L^2$ percentages. The last column is $10^3$ times the time-RMS interface distance.}
\label{tab:ac-current-neural-controls}
\begin{tabular}{lrrrr}
\toprule
Repeat & Trajectory & Final time & Band & $10^3d_{\rm rms}$\\
\midrule
1 & 0.1224 & 0.1222 & 0.6569 & 0.4931\\
2 & 0.1148 & 0.1117 & 0.6130 & 0.4487\\
3 & 0.1384 & 0.1818 & 0.7562 & 0.5526\\
\bottomrule
\end{tabular}
\end{table}

\begin{figure}[htbp]
\centering
\includegraphics[width=\linewidth]{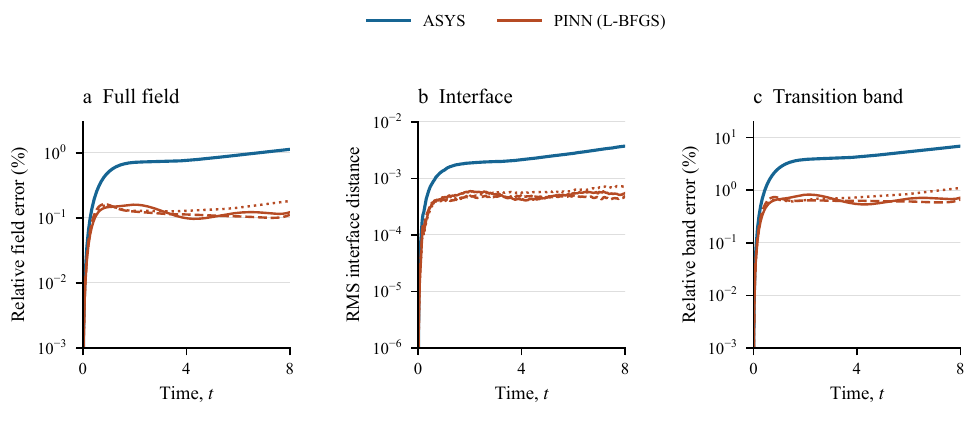}
\caption{\textbf{The original ASYS curvature construction and the PINN baseline.} The reported ASYS field and three PINN/L-BFGS fits are evaluated on the same numerical reference. Blue shows ASYS; solid, dashed and dotted orange curves show the three neural fits. \textbf{a,} Relative field error over the full periodic square. \textbf{b,} Bidirectional RMS distance between sampled zero contours. \textbf{c,} Relative field error within the reference transition band $|u_{\rm ref}|\le0.9$.}
\label{fig:ac-neural-controls}
\end{figure}

\FloatBarrier
\subsection{Detailed result metrics}
\subsubsection{gCLM: field errors and tail coefficients}

Runs 1--5 denote five independent searches from the same starting ansatz, with one program retained by equation score from each search. Run 3 supplied the selected ASYS program discussed in the main text. Figure~\ref{fig:supp-gclm-search} shows their equation-residual histories; Table~\ref{tab:gclm-all} gives the retained profiles' errors against the same fine reference bank. Methods Section~\ref{supp:eval} defines the logarithmic-radius error $E_{\log}$, physical-radius error $E_y$, peak-normalized maximum error $E_\infty$ and exponent error $|\Delta\beta|$.

\begin{figure}[htbp]
\centering
\includegraphics[width=\linewidth]{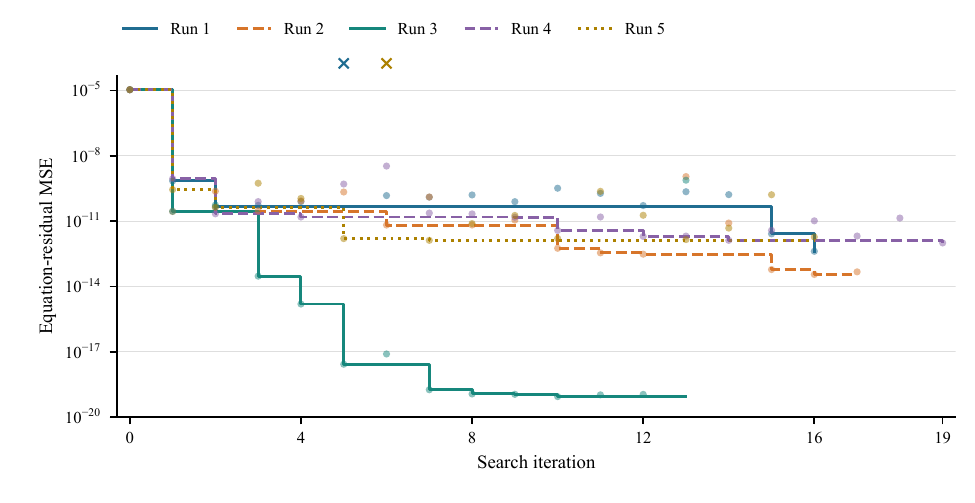}
\caption{\textbf{Progress across five independent gCLM searches.} Lines show the lowest equation-residual MSE reached so far; dots show individual programs. Crosses above the axes mark failed evaluations, which have no residual value. Iteration 0 is the shared starting program, and each curve ends at its search's last completed iteration. The curves show equation residuals used during search, rather than errors against the reference profile.}
\label{fig:supp-gclm-search}
\end{figure}

\begin{table}[htbp]
\centering\small
\caption{Five equation-selected profiles at $a=1/4$, evaluated on radii $10^{-3}$--$10^3$ against the 4,097-node fine bank. The median $E_{\log}$ was $3.90\times10^{-5}\%$. Relative field errors are reported in percent.}
\label{tab:gclm-all}
\begin{tabular}{lrrrr}
\toprule
Search & $E_{\log}$ (\%) & $E_y$ (\%) & $E_\infty$ (\%) & $|\Delta\beta|$\\
\midrule
Run 1 & $1.53\times10^{-4}$ & $1.86\times10^{-4}$ & $1.76\times10^{-4}$ & $1.61\times10^{-7}$\\
Run 2 & $1.51\times10^{-6}$ & $3.35\times10^{-6}$ & $1.62\times10^{-6}$ & $6.93\times10^{-10}$\\
Run 3 & $2.47\times10^{-8}$ & $2.72\times10^{-8}$ & $2.60\times10^{-8}$ & $4.41\times10^{-11}$\\
Run 4 & $3.90\times10^{-5}$ & $5.30\times10^{-5}$ & $4.34\times10^{-5}$ & $3.32\times10^{-8}$\\
Run 5 & $5.35\times10^{-5}$ & $6.64\times10^{-5}$ & $5.62\times10^{-5}$ & $4.76\times10^{-8}$\\
\bottomrule
\end{tabular}
\end{table}

The resulting forms differ: Run 1 uses a M\"obius--Chebyshev core and rational/confluent tails; Run 2 uses Sobolev--Jacobi coordinates and a movable pole; Run 3 uses projected Mellin and rational modes; Run 4 uses Chebyshev and confluent mixed-tail terms; Run 5 uses a $C^4$ local quintic core. Their fitted/buffer scalar counts are respectively $13/10$, $19/149$, $15/495$, $11/29$ and $16/2021$. Stored basis maps are distinct from fitted coefficients.

For a raw exterior $h(x)=(x/r)r^{-\alpha}\exp[b+\ell_\alpha r^{-\alpha}+\cdots]$, normalization gives
\begin{equation}
 C=e^b A^{-\alpha-1},\qquad D=e^b\ell_\alpha A^{-2\alpha-1},\qquad
 D/C^2=Ae^{-b}\ell_\alpha.
\label{supp:eq:cd-extract}
\end{equation}
These are algebraic readouts of the saved formulas. Table~\ref{tab:gclm-tail} compares them with the balance in Eq.~\eqref{supp:eq:K-def}. Although Run 5 achieved a small profile error, its secondary tail coefficient fell about $10.6\%$ below the required value, whereas Run 1 had a larger profile error but a smaller tail deficit; the analytical tail readout therefore provides a complementary test beyond sampled error.

\begin{table}[htbp]
\centering\small
\caption{Tail diagnostics for the five saved profiles. $\rho=D/[C^2K(\alpha)]$. The differentiated exterior assumptions were checked analytically for Run 3 and Run 5; the physical rate interpretation for other rows remains conditional on those assumptions.}
\label{tab:gclm-tail}
\begin{tabular}{lrrrrr}
\toprule
Program & $\beta$ & $C$ & $D$ & $\rho-1$ & $D/C^2-K$\\
\midrule
Run 1 & $0.684514$ & $1.75379$ & $1.25399$ & $-1.09\times10^{-2}$ & $-0.00449$\\
Run 2 & $0.684514$ & $1.75378$ & $1.27627$ & $+6.70\times10^{-3}$ & $+0.00276$\\
Run 3 & $0.684514$ & $1.75378$ & $1.26780$ & $+2.49\times10^{-5}$ & $+1.03\times10^{-5}$\\
Run 4 & $0.684514$ & $1.75377$ & $1.28716$ & $+1.53\times10^{-2}$ & $+0.00631$\\
Run 5 & $0.684514$ & $1.75379$ & $1.13289$ & $-1.06\times10^{-1}$ & $-0.0439$\\
\bottomrule
\end{tabular}
\end{table}

\subsubsection{Transient Fujita: field accuracy and fitting cost}\label{supp:fujita-metrics}\label{supp:fujita-comparison}

The local-response fit stopped after $96.3$ seconds and 859 loss evaluations under the no-improvement rule; the uniform-response fit used $300.1$ seconds and 2,582 evaluations. Relative radial-profile $L^2$ error was $0.0879\%$ over the trajectory and $0.702\%$ at its final time, compared with $3.99\%$ and $25.43\%$ for the uniform response. The norms are defined in Methods Section~\ref{methods:fujita-reference}.

\subsubsection{NLS: field, mass and residual diagnostics}\label{supp:nls-defect-weak}
The field discrepancy $\mathcal E$ and sampled mass drift $\Delta M$ are defined in Methods Section~\ref{methods:nls-diagnostics}. Adding $P$ to $C+B$ gave a $67.9$--$95.9\%$ field-error reduction. The pooled error across the six widths is $0.529892\%$ for $C+B$ and $0.0249688\%$ for the prescribed $C+B+P+E$ reported in Section~\ref{sec:nls}. Table~\ref{tab:nls-construction-controls} compares the predictor and defect correction on the finer readout grids, with all amplitudes prescribed.

\begin{table}[htbp]
\centering\small
\caption{Prescribed NLS predictor and defect correction. The field discrepancy of $C+B+P+E$ agrees with that of $C+B+P$ at the displayed precision. Mass drift is in parts per million. Both fields use the common readout grids specified in Methods Section~\ref{methods:nls-diagnostics}.}
\label{tab:nls-construction-controls}
\begin{tabular}{rrrr}
\toprule
$L$ & $\mathcal E_P$ (\%) & $\Delta M_P$ (ppm) & $\Delta M_{P+E}$ (ppm)\\
\midrule
3.5 & 0.04032 & 2.520 & 0.07771\\
4 & 0.02840 & 0.5649 & 0.1438\\
4.5 & 0.01938 & 0.1699 & 0.1699\\
5 & 0.01293 & 0.2026 & 0.2026\\
6 & 0.005689 & 0.1827 & 0.1827\\
7 & 0.002391 & 0.09902 & 0.09902\\
\bottomrule
\end{tabular}
\end{table}

Adding $E$ lowered mass drift by $96.9\%$ at $L=3.5$ and $74.5\%$ at $L=4$ while leaving wider boxes essentially unchanged, without a mass-conservation penalty.

\Needspace{6\baselineskip}
At $L=3.5$, doubling the mass quadrature from 2,048 to 4,096 spatial points at the same 321 saved times changed the $P+E$ drift from $0.07473$ to $0.07771$ ppm. A separate check on 4,096 spatial points with nested uniform grids of 321, 641 and 1,281 times preserved both narrow-box improvements: the last temporal doubling changed the drift by at most $0.211\%$.

For $P$, the standard-layout weak diagnostic in Eq.~\eqref{eq:nls-weak-diagnostic} ranged from $1.11\times10^{-8}$ at $L=3.5$ to $2.72\times10^{-12}$ at $L=7$. The shifted values were about four times smaller. Sensitivity to the finite quadrature and to the breather's boundary derivative jump limits the use of these diagnostic values as continuum residual estimates.

\subsubsection{Allen--Cahn: coupled-mode fields}\label{supp:ac-coupled-metrics}
Methods Section~\ref{methods:ac-protocol} gives the fitting procedure and error norm for the two compact fields in Section~\ref{supp:ac-mode-coupling}. Their weighted error energies are $0.00193691480$ and $0.00255235607$ for the quadratic construction and selected cubic extension, respectively, with common reference energy $30.4055615$.

\begin{table}[htbp]
\centering\small
\caption{Accuracy of the complete coupled-mode Allen--Cahn fields with frozen coefficients. Trajectory and final-field errors are relative $L^2$ percentages. The equation residual is the separate fixed search-evaluator MSE.}
\label{tab:ac-coupled-accuracy}
\begin{tabular}{lrrrr}
\toprule
Construction & $n_\theta$ & Trajectory (\%) & $t=8$ (\%) & Equation MSE\\
\midrule
Quadratic modes & 14 & $0.798139$ & $1.13278$ & $4.76630\times10^{-5}$\\
Selected cubic extension & 12 & $0.916208$ & $1.17654$ & $4.73542\times10^{-5}$\\
\bottomrule
\end{tabular}
\end{table}

The extension has a slightly lower equation residual but a higher field error (Table~\ref{tab:ac-coupled-accuracy}). Both fixed initial-value MSEs are $2.62506\times10^{-14}$, both periodic-boundary MSEs are zero, and their initial-secant MSEs are $1.11128\times10^{-7}$ and $1.17878\times10^{-7}$. The zero-contour overlays and spatial error maps (Fig.~\ref{fig:ac-coupled-errors}) show where discrepancies occur within the full-domain norm.

\begin{figure}[htbp]
\centering
\includegraphics[width=\linewidth]{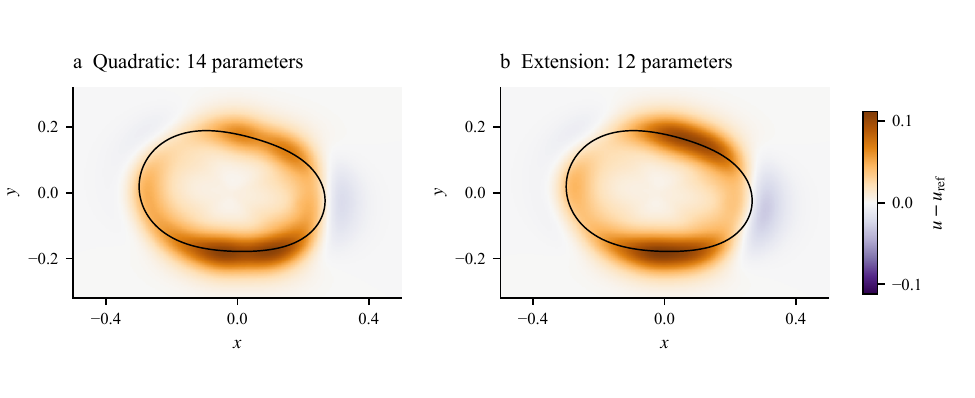}
\caption{\textbf{Where the coupled-mode fields differ from the reference.} Field errors at $t=8$ for the fourteen-parameter quadratic construction and twelve-parameter extension, on a common color scale. Black contours mark the reference interface. The display is cropped around the evolving phase; the reported norms use the full periodic square.}
\label{fig:ac-coupled-errors}
\end{figure}

\FloatBarrier
\subsection{Further profile and inference constructions}
\label{supp:additional-constructions}

\subsubsection{Repairing the far field of a Graveleau pressure profile}
\label{supp:graveleau-tail-repair}

For the Graveleau shrinking-hole profile defined in Section~\ref{methods:graveleau}, the similarity equation~\eqref{eq:graveleau-profile} determines which corrections can accompany the leading pressure $\Psi\sim C\eta^p$. Here $\eta$ is radius divided by the hole radius, $\beta$ is the contraction exponent and $p=2-1/\beta$. ASYS identified a structural obstruction in its earlier family,
$\Psi=(\eta^p-1)\exp[xP(x)]/p$, $x=1-\eta^{-1}$, where $P$ is a fitted polynomial. Its far field retains a constant offset $-\exp(P(1))/p$---leaving a nonzero limiting equation defect for every finite choice of coefficients---whereas the equation requires a first correction proportional to $\eta^{2p-2}$.

The revised family uses the fractional coordinate $x=1-\eta^{p-2}$:
\begin{equation}
 \begin{aligned}
 \Psi(\eta)&=\beta\eta^p x\exp[xP(x)],\\
 \Psi(\eta)&=C\eta^p+D\eta^{2p-2}+O(\eta^{3p-4}),\\
 C&=\beta e^{P(1)},\qquad D=-C[1+P(1)+P'(1)].
 \end{aligned}
\end{equation}
Retaining the existing nine-coefficient shape polynomial within the 24-scalar pressure construction, this revised family preserves the front value and slope, removes the constant offset and admits the $\eta^{2p-2}$ correction so that its coefficients can express the leading far-field balance $D=-2C^2p^2/(2-p)$. The result is a repair of the approximation family: the tail balance becomes expressible through its coefficients.

\subsubsection{An exponential time basis for Hamilton--Jacobi inference}
\label{supp:hj-inverse}

For the Hamilton--Jacobi inverse task in Section~\ref{methods:hj}, ASYS fitted a field and the four unknown coefficients $(d_x,d_y,b_x,b_y)$ in Eq.~\eqref{eq:hj-inverse}. Its construction combines exact heat evolution with a Fourier correction. Quadratic interactions of diffusing Fourier modes produce differences of exponentials in time, motivating the returned representation
\begin{equation}
 \begin{aligned}
 u(x,y,t)&=e^{tL_d}u_0(x,y)+\sum_{k,j}a_{kj}\phi_k(x,y)I_j(w(t/H)),\\
 L_d&=d_x\partial_{xx}+d_y\partial_{yy},\qquad
 w(s)=\frac{1-e^{-s/2}}{1-e^{-1/2}},\\
 I_j(w)&=\int_0^w P_j(2z-1)\,dz,\qquad j=0,\ldots,5.
 \end{aligned}
\end{equation}
Here $\phi_k$ are 1,257 real Fourier functions with integer wavevectors $|k|\leq20$, $P_j$ are Legendre polynomials and $H=34/98$ is the observation horizon. The six time functions span $1-e^{-ms/2}$, $m=1,\ldots,6$, with $s=t/H$. Their zero initial value preserves $u_0$ exactly. There are 7,542 field amplitudes, fitted jointly with the four physical coefficients.

Replacing polynomial physical-time functions with this exponential time coordinate, at fixed spatial basis, field size and fitting settings, reduced future-field relative $L^2$ error from $3.36\%$ to $1.06\%$; final-time error fell from $8.27\%$ to $2.37\%$. Mean relative coefficient error was essentially unchanged ($0.784\%$ versus $0.782\%$ for the polynomial-time fit); the gain was in continuation of the field beyond the observations.

\fi
\end{document}